%% file: main.tex
\documentclass[runningheads]{llncs}
\usepackage{graphicx}

\usepackage{tikz}
\usepackage{comment}
\usepackage{amsmath,amssymb} 
\usepackage{color}
\usepackage[accsupp]{axessibility}  

\input{latex/header}

\begin{document}
\pagestyle{headings}
\mainmatter
\def\ECCVSubNumber{2804}  

\title{\name: Rethinking the Evaluation of 3D Face Reconstruction} 

\titlerunning{\name: Rethinking the Evaluation of 3D Face Reconstruction}
%
\author{Zenghao Chai\inst{1}$^{*}$\orcidlink{0000-0003-3709-4947} \and
Haoxian Zhang\inst{2}$^{*}$\orcidlink{0000-0001-7078-868X} \and
Jing Ren\inst{2}\orcidlink{0000-0003-3114-3517} \and Di Kang\inst{2}\orcidlink{0000-0002-8996-0897} \and Zhengzhuo Xu\inst{1}\orcidlink{2222--3333-4444-5555} \and Xuefei Zhe\inst{2}\orcidlink{0000-0002-5005-7166} \and Chun Yuan\inst{1,3}$^\dagger$\orcidlink{0000-0002-3590-6676} \and Linchao Bao\inst{2}$^\dagger$\orcidlink{0000-0001-9543-3754} } 
\authorrunning{Z. Chai et al.}
%
\institute{Shenzhen International Graduate School, Tsinghua University, China \and
Tencent AI Lab, China, $^3$ Peng Cheng National Laboratory, China \\
\footnotetext{$^*$ Equal Contributions.}
\footnotetext{$^\dagger$ Corresponding authors: \email{yuanc@sz.tsinghua.edu.cn; linchaobao@gmail.com}.}}


\maketitle
{
\setlength{\textfloatsep}{6pt}

\setlength{\abovecaptionskip}{0pt}
\setlength{\belowcaptionskip}{0pt}

\begin{abstract}
The evaluation of 3D face reconstruction results typically relies on a rigid shape alignment between the estimated 3D model and the ground-truth scan. We observe that aligning two shapes with different reference points can largely affect the evaluation results. This poses difficulties for precisely diagnosing and improving a 3D face reconstruction method.
In this paper, we propose a novel evaluation approach with a new benchmark {\name}, consists of 100 globally aligned face scans with accurate facial keypoints, high-quality region masks, and topology-consistent meshes.
Our approach performs region-wise shape alignment and leads to more accurate, bidirectional correspondences during computing the shape errors.
The fine-grained, region-wise evaluation results provide us detailed understandings about the performance of state-of-the-art 3D face reconstruction methods. For example, our experiments on single-image based reconstruction methods reveal that DECA performs the best on nose regions, while GANFit performs better on cheek regions.
Besides, a new and high-quality 3DMM basis, {\nextpp}, is further derived using the same procedure as we construct {\name} to align and retopologize several 3D face datasets.
We will release {\name}, {\nextpp}, and our new evaluation pipeline at \href{https://realy3dface.com}{https://realy3dface.com}.

\keywords{3D Face Reconstruction, Evaluation, Benchmark, 3DMM}
\end{abstract}

\section{Introduction}

3D face reconstruction is a hotspot with broad applications in real world including face alignment \cite{3DDFA,3DDFA_v2}, face recognition \cite{3dmm10000,posecvpr18,facerecpami09}, and face animation \cite{annimtog13,annimtog16} among many others. 
How to estimate high fidelity 3D facial mesh \cite{MoFA,deep3d,uncertaintycvpr20} from monocular RGB(-D) images or image collections is a challenging problem in the fields of computer vision, computer graphics and machine learning.

Various methods have been proposed to tackle this problem, among which DNNs, especially CNNs~\cite{CNNRegress,learnfromsyn,MoFA} and GCNs~\cite{GCNcvprw20,GCNcvpr20}, have made great progress due to their great expressiveness.
However, developing new reconstruction methods and evaluating different methods or 3DMM basis are severely constrained by available datasets. 
Existing open-source 3D face datasets~\cite{MICC,facescape,FRGC2,BU4DFE} have some unneglectable flaws. 
For example, the face scans are in different scales and random poses, and the provided keypoints are not accurate or discriminative enough, which makes it extremely hard to align the input shapes to the predicted face for evaluation.
Moreover, due to the lack of ground-truth annotations in the original face scans, standard evaluation pipeline relies on nearest-neighboring correspondences to measure the similarity between the scan and the estimated face shape, which completely ignores substantive characteristics and discards shape geometry of human faces.

\begin{figure*}[!t]
\centering
\begin{overpic}[trim=0cm 30.5cm 28.3cm 0cm,clip,width=1\linewidth,grid=false]{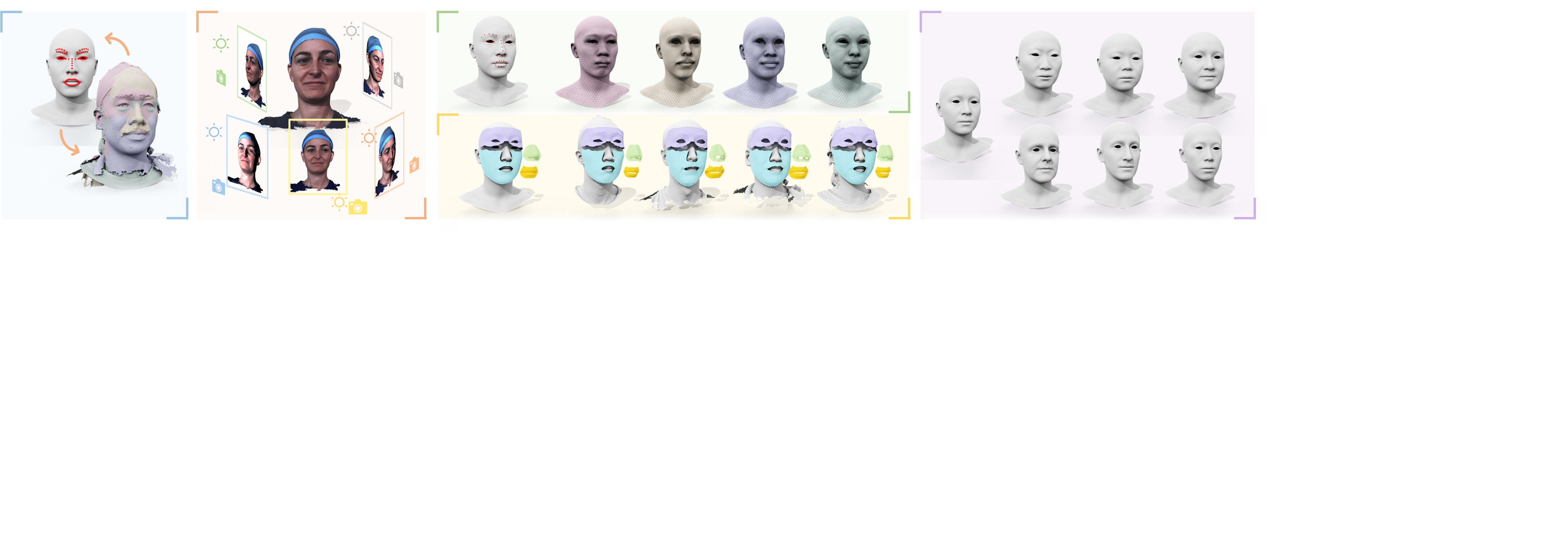}
    \put(0,15.3){\tiny\bfseries (a)}
    \put(15.7,15.3){\tiny\bfseries (b)}
    \put(35,15.3){\tiny\bfseries (c)}
    \put(35,7){\tiny\bfseries (d)}
    \put(73.5,15.3){\tiny\bfseries (e)}
    
    \put(3,1){\tiny\bfseries\itshape aligned scans}
    \put(19,1){\tiny\bfseries\itshape multi-view images}

    \put(35,11){\itshape\bfseries\tiny {\tempshape}}
    \put(41,12){\tiny\bfseries\itshape $\mathcal{K}_L$}
        
    \put(35,2.4){\itshape\bfseries\tiny {\tempshape}}
    \put(35,0.5){\itshape\bfseries\tiny (regions)}

    \put(74.5,13.2){\tiny\bfseries\itshape shape}      
    \put(74.5,12){\tiny\bfseries\itshape mean}
    
    \put(83,15.8){\tiny\bfseries\itshape shape components}       
    \put(81,8){\tiny\bfseries\itshape $\pm 3\sigma$}   
    \put(87.5,8){\tiny\bfseries\itshape $\pm 3\sigma$} 
    \put(94,8){\tiny\bfseries\itshape $\pm 3\sigma$} 
    \put(81.5,1){\tiny\bfseries\itshape $1$st}   
    \put(88,1){\tiny\bfseries\itshape $2$nd} 
    \put(94.5,1){\tiny\bfseries\itshape $3$rd} 
    
\end{overpic}
\caption{
\textbf{{\name}}: a \underline{\textbf{Re}}gion-\underline{\textbf{a}}ware benchmark based on the \underline{\textbf{LY}}HM~\cite{LYHM} dataset.
Our benchmark contains $100$ high-quality face shapes and \emph{each} individual has \textbf{(a)} a rescaled and globally aligned scan, 
\textbf{(b)} $5$ synthesized \emph{multi-view} images with various GT
camera parameters and illuminations, 
\textbf{(c)} a retopologized full-head mesh in {\nextone}~\cite{hifi3dface2021tencentailab} topology with consistent and semantically meaningful $68$ keypoints, 
\textbf{(d)} $4$ consistent region masks defined on both the retopologized mesh and the original scan, 
and \textbf{(e)} {\nextpp} 3DMM: the first three PCs
with the mean shape show the ethnic diversity.
}
\label{fig:autowrap}
\end{figure*}
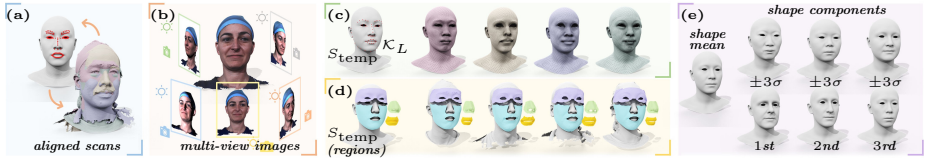

To fill this gap, we propose a new benchmark named {\name} for evaluating 3D face reconstruction methods. {\name} contains 3D face scans of $100$ individuals from the LYHM~\cite{LYHM} dataset, where the face scans are consistently rescaled, globally aligned, and wrapped into topology-consistent meshes.
More importantly, since we have predefined facial keypoints and masks of the retopologized mesh template, the keypoints and masks can be transferred to original face scans. In this case, we get the high-quality facial keypoints and masks of the original raw face scans, which enable us to perform more accurate alignments and fine-grained, region-wise evaluations for estimated 3D face shapes.
See Fig.~\ref{fig:autowrap} for an illustration.
Our benchmark contains individuals from different ethnic, age, and gender groups (see Fig.~\ref{fig:eg_benchmark} for some examples).
Utilizing the retopologizing procedure built for {\name}, we further present a high-quality and powerful 3DMM basis named {\nextpp} by aligning and retopologizing several 3D face datasets.
We conduct extensive experiments to evaluate state-of-the-art 3D face reconstruction methods and 3DMMs, which reveal several interesting observations and potential future research directions.

\myparagraph{Contributions.} To summarize, our main contributions are:
\begin{itemize}[leftmargin=*,nosep,nolistsep]
\item A new 3D face benchmark {\name} that contains prealigned scans with accurate facial keypoints and region masks, retopologized meshes, and rendered high-fidelity multi-view images with camera parameters.
\item A thorough investigation of the flaws in the standard evaluation pipeline for measuring face reconstruction quality. 
\item A novel, informative evaluation approach for 3D face reconstruction, with an elaborated region-wise, bidirectional alignment pipeline.
\item Extensive experiments for benchmarking state-of-the-art 3D face reconstruction methods and 3DMMs.
\item A new full-head 3DMM basis {\nextpp} built from several 3D face datasets with high-quality, consistent mesh topology.
\end{itemize}

\begin{figure*}[!t]
    \centering
    \begin{overpic}[trim=2cm 16.5cm 9.3cm 0cm,clip,width=1\linewidth,grid=false]{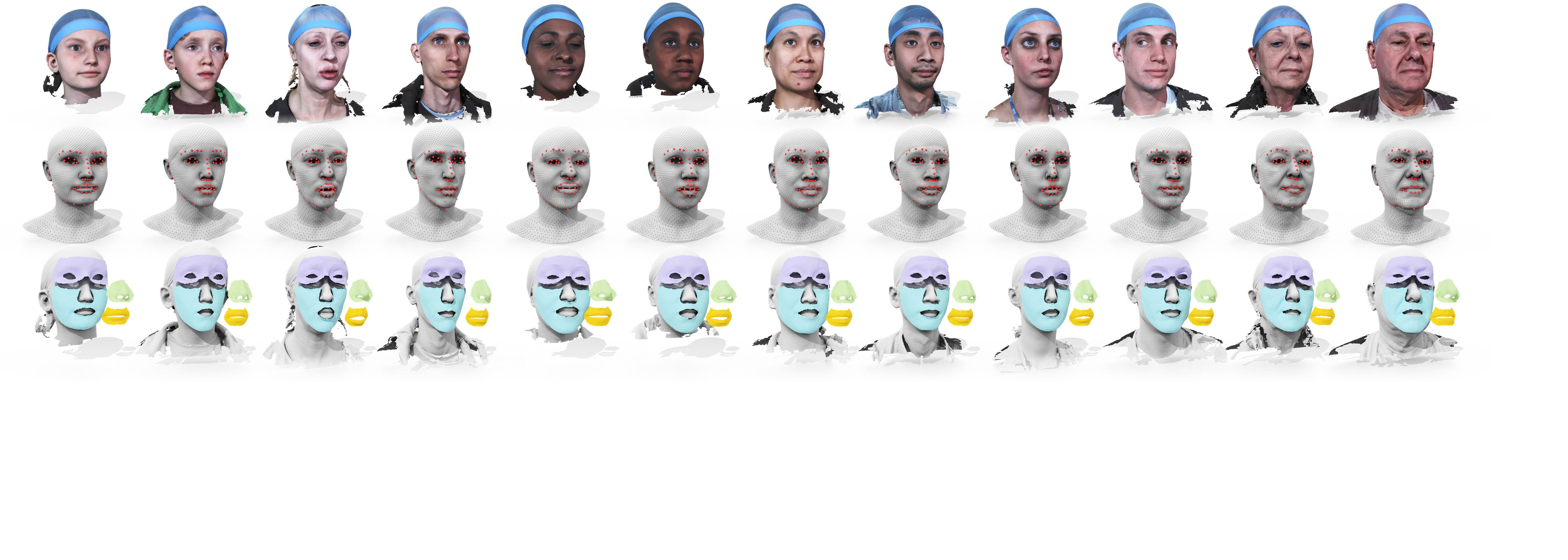}
    \end{overpic}
    \caption{\textbf{Examples from {\name}}. \emph{Top}: aligned high-resolution scans with textures. \emph{Middle}: retopologized meshes in {\nextone} topology with semantically consistent keypoints (red points). \emph{Bottom}: high quality face region masks of each scan.}\label{fig:eg_benchmark}
\end{figure*}

\section{Related work}
Face reconstruction has drawn great attention in the past decades in both computer vision and computer graphics communities~\cite{3DDFA,conditionestiiccv21,MGCNet,meingameaaai20,LAP,avatarmecvpr20,wu2019mvf}.
Below we review the topics that are most closely related to our work, and a full in-depth review can be found in~\cite{3dmmfuturetog20,facereview18,facereview14}.

\myparagraph{3D Face Database.}
High quality 3D scan datasets greatly promote the development in the field of 3D face reconstruction. 
Massive face databases~\cite{3DDFA,celeba,FFHQ} have made it possible to train models for face reconstruction in a self-supervised manner.
However, the 3D face scans in the existing datasets~\cite{FRGC2,MICC,LYHM,BU4DFE} are in different scales and random poses,
and only a small set of inaccurate keypoints are provided for alignment. 
Another type of databases~\cite{facescape,FaceWarehouse,BU4DFE} contains retopologized meshes that are registered from high fidelity scans where all the meshes share the same topology. This type of databases is essential to construct 3D Morphable Models (3DMMs)~\cite{3DMM,3dmm10000,facescape,bfm09}, statistical models of facial shape and texture, which can be used for face regression and editing. 

\myparagraph{Single-View 3D Face Reconstruction.}
3D face reconstruction from a single-view image has received glaring attention over the past decades, though estimating 3D information from a single 2D image is challenging and severely ill-posed.
With the help of 3D morphable face models~\cite{3DMM,flametopo,3dmmfuturetog20,FaceWarehouse,combine3dmmcvpr19,albedommcvpr20}, the reconstruction problem is simplified into a tractable parametric regression. 
A straightforward solution to estimate 3DMM coefficients is based on analysis-by-synthesis~\cite{BlanzV03,Face2Face,YamaguchiSNZ0OM18,HuSWNSFSSCL17}, where the optimization objective usually consists of facial landmark alignment, photo consistency and statistical regularizers. These optimization-based approaches are computationally expensive and sensitive to initialization. 
Recently, many deep learning based models~\cite{CNNRegress,MoFA,unsupcvpr2018,learnfromsyn} are proposed to predict the 3DMM coefficients in a supervised or self-supervised way.
This type of methods is robust for face reconstruction but has limited expressive power. 
To address this issue, GANFit~\cite{ganfit,gecer2021fast} proposes to parameterize the texture maps using the latent code of a Texture GAN for face regression. 
Some nonlinear 3DMMs~\cite{Nonlinear3DMM,tran2018on,tran2019towards} are proposed for stronger expressiveness of face geometry. 
Some other work~\cite{DECA,yajingtip,beyond3dmmeccv20,conditionestiiccv21,Recognizable} utilize additional geometry and appearance representation (such as displacement and normal maps) to recover high-frequency details.
Moreover, a recent surge of end-to-end approaches try to reconstruct 3D face shape directly from a depth map or UV position map~\cite{PRNet,DF2Net,img2imgtransiccv17,normavatarcvpr21,pixelavatarcvpr21,probsymcvpr20}. However, these non-3DMM methods are prone to produce unrealistic and malformed faces compared to 3DMM-based methods.

\myparagraph{Evaluation of Face Reconstruction.}
Existing evaluation protocols usually utilize the off-the-shelf datasets \cite{MICC,FaceWarehouse,FRGC2,BU4DFE} to estimate the similarity between the reconstructed shape and the raw scan. 
Specifically, the reconstructed shape and the input scan are aligned using some predefined keypoints~\cite{disencvpr2018,yajingtip,FengBenchmark,beyond3dmmeccv20} or ICP~\cite{ICP,deep3d,ganfit,gecer2021fast,unsupcvpr2018,RingNet}.
Then Root Mean Square Error (RMSE)~\cite{CNNRegress,deep3d,pixelface20,personalizedeccv20} or Normalized Mean Square Error (NMSE)~\cite{beyond3dmmeccv20,poseaccv20,tran2019towards,uncertaintycvpr20} is calculated between the corresponding points on the input scan and the reconstructed shape. 
The correspondences are usually established in two ways, namely finding the nearest neighbor with smallest point-to-point distance \cite{beyond3dmmeccv20,yajingtip} or point-to-plane distance \cite{deep3d,poseaccv20,normavatarcvpr21}.
Some benchmarks \cite{FengBenchmark,RingNet,pixelface20} propose to use predefined keypoints or disk-shaped region masks to measure shape similarity. 
Such measurement contains semantic prior but does not faithfully represent the overall face shape.

\section{Background}

\subsection{Notation \& Preliminaries}
We use \emph{triangle} mesh $\Scale[0.9]{S = \{\,V, F\,\}}$ to represent face models with vertex positions $\Scale[0.9]{V}$ and triangle face list $\Scale[0.9]{F}$.
A \emph{region-of-interest} of the mesh $\Scale[0.9]{S}$ is denoted as $\Scale[0.9]{\mathcal{R}_{S}}$, which can be represented as an indicator function or a list of face IDs.
We denote the \emph{keypoints} on the mesh $\Scale[0.9]{S}$ as $\Scale[0.9]{\mathcal{K}_S}$, which is a list of manually selected or automatically detected vertices that are semantically meaningful on $\Scale[0.9]{S}$.

{
For a specific face shape, we consider three associated meshes: 
(1) $\Scale[0.9]{S_H}$: the \emph{ground-truth} mesh with \emph{high} resolution, which is constructed from multi-view images; 
(2) $\Scale[0.9]{S_L}$: the \emph{ground-truth} retopologized mesh with \emph{low} resolution, which is obtained by wrapping the {\nextone}~\cite{hifi3dface2021tencentailab} mesh topology to the shape $\Scale[0.9]{S_H}$; 
(3) $\Scale[0.9]{S_P}$: the \emph{predicted} face mesh constructed from existing techniques.
Note that different reconstruction methods may choose different mesh topologies (i.e., different size of $\Scale[0.9]{V_P}$ and $\Scale[0.9]{F_P}$). 
For simplicity of notation, we denote the \emph{regions} and \emph{keypoints} defined on shape $\Scale[0.9]{S_H/S_L/S_P}$ as $\Scale[0.9]{\mathcal{R}_{H}/\mathcal{R}_{L}/\mathcal{R}_{P}}$ and $\Scale[0.9]{\mathcal{K}_{H}/\mathcal{K}_{L}/\mathcal{K}_{P}}$ respectively.
}

A \emph{map} between the shape $\Scale[0.9]{S_i}$ and $\Scale[0.9]{S_j}$ is denoted as $\Scale[0.9]{\mymap{i}{j}{}}$, where the subscript represents the map \emph{direction}. For example, $\Scale[0.9]{\mymap{p}{h}{}}$ represents a map from the predicted mesh $\Scale[0.9]{S_P}$ to the high-resolution GT
mesh $\Scale[0.9]{S_H}$. 
We consider two different types of map, the vertex-to-vertex map $\Scale[0.9]{\mymap{i}{j}{vtx}}$ and the vertex-to-point (also called vertex-to-plane) map $\Scale[0.9]{\mymap{i}{j}{pts}}$. 
Specifically, $\Scale[0.9]{\mymap{i}{j}{vtx}}$ maps each vertex in shape $S_i$ to a \emph{vertex} on 
$\Scale[0.9]{S_j}$, and $\Scale[0.9]{\mymap{i}{j}{pts}}$ maps each vertex on shape $\Scale[0.9]{S_i}$ to a \emph{point} in a face of shape $\Scale[0.9]{S_j}$. 
We will use the superscript to disambiguate the two maps when necessary. 

\myparagraph{Normalized Mean Square Error (NMSE)} computes the distance between two surfaces $\Scale[0.9]{S_i}$ and $\Scale[0.9]{S_j}$ based on a given map $\Scale[0.9]{\mymap{i}{j}{}}$ and is denoted as $\Scale[0.9]{e\big(\mymap{i}{j}{}\big)}$:
\begin{equation}\label{eq:nmse}
    \Scale[0.92]{e\big(\mymap{i}{j}{}\big) = \frac{1}{n_v} \sum_{v\in S_i} \big\Vert\, v - \mymap{i}{j}{}(v) \,\big\Vert_F^2}
\end{equation}
where $\Scale[0.9]{n_v}$ is the number of vertices in shape $\Scale[0.9]{S_i}$, and $\Scale[0.9]{\mymap{i}{j}{}(v)}$ gives the \emph{coordinates} of the mapped position (of a vertex/point on shape $\Scale[0.9]{S_j}$) for vertex $\Scale[0.9]{v\in S_i}$.

\myparagraph{Iterative Closest Point (ICP)} can be applied to align two shapes via solving a rigid transformation and a nearest neighbor map iteratively to minimize NMSE:
\begin{equation}\label{eq:icp}
\Scale[0.92]{\min_{\rot, \tran, \mymap{i}{j}{}}\quad \sum_{v\in S_i} \big\Vert\, \rot v + \tran - \mymap{i}{j}{}(v)\,\big\Vert_F^2} ,
\end{equation}
where the 3D rotation matrix $\Scale[0.9]{\rot}$ and the 3D translation vector $\Scale[0.9]{\tran}$ are computed to align $\Scale[0.9]{S_i}$ to $\Scale[0.9]{S_j}$. For simplicity, we denote $\Scale[0.9]{\big[\rot, \tran, \mymap{i}{j}{}, S_i^*\big] = \textbf{ICP}\big(S_i \rightarrow S_j\big)}$, where $S_i^*$ is obtained by transforming $S_i$ via $(\rot, \tran)$. 
For the purpose of efficiency, some previous works~\cite{disencvpr2018,yajingtip,FengBenchmark} only consider a small set of predefined corresponding keypoints to solve for $\Scale[0.9]{\rot}$ and $\Scale[0.9]{\tran}$ instead of using every vertex $\Scale[0.9]{v\in S_i}$.

\subsection{3DMM and Face Reconstruction}
\myparagraph{Formulation} The goal is to reconstruct a 3D face shape $\Scale[0.9]{S_P}$ from a single RGB(-D) image or an image collection. 
To reduce the search space of plausible faces, 3DMM is commonly used for face representation: $\Scale[0.9]{S = \bar{S} + \Phi\alpha}$, where $\Scale[0.9]{\bar{S}}$ is the \emph{mean} shape, and $\Scale[0.9]{\Phi}$ is the \emph{principle components} (PCs) trained on some 3D face scans with neutral expression. 
Then the face reconstruction problem is reduced to a regression problem of solving for the facial parameter $\Scale[0.9]{\alpha}$.
Existing methods either use deep networks to predict the $\alpha${~\cite{MoFA,unsupcvpr2018,yajingtip,deep3d,3DDFA_v2,DECA}} or optimize various energy terms (e.g., most commonly used keypoint loss and photometric loss~\cite{BlanzV03,ganfit,gecer2021fast,inoptimize,Face2Face,YamaguchiSNZ0OM18}) with different regularization terms (e.g., not deviate too far from mean face~\cite{Face2Face,ganfit,gecer2021fast,hifi3dface2021tencentailab,ganoptimize}).

\begin{table}[!t]
\centering
\caption{\textbf{Overview of 3DMMs.} 
}
\label{tab:3dmm_topo}

\input{latex/tables/table_3DMM_cmp.tex}
\end{table}

\myparagraph{3DMM}
The facial basis $\Phi$ determines the expressiveness power of the corresponding 3DMM and affect the reconstruction quality. 
Publicly available 3DMMs include BFM~\cite{3DMM,bfm09}, LSFM~\cite{3dmm10000}, FLAME~\cite{flametopo}, LYHM~\cite{LYHM}, FaceScape (FS)~\cite{facescape}, FaceWareHouse (FWH)~\cite{FaceWarehouse}, and {\nextone}/{\nexttwo}~\cite{hifi3dface2021tencentailab} (see Tab.~\ref{tab:3dmm_topo}).

\myparagraph{Standard Evaluation Pipeline}
Some datasets~\cite{MICC,FRGC2,RingNet,FengBenchmark,beyond3dmmeccv20} also provide ground-truth scans, i.e., $\Scale[0.9]{S_H}$ with keypoints $\Scale[0.9]{\mathcal{K}_H}$, that are associated with the input images, which allow us to evaluate the quality of the reconstructed shape $\Scale[0.9]{S_P}$. Standard evaluation process consists of three steps: 
(1) first rescale and align $\Scale[0.9]{S_P}$ with $\Scale[0.9]{S_H}$ based on some sparse keypoints or applying ICP, 
(2) find the map $\Scale[0.9]{\mymap{p}{h}{}}$~\cite{raytracingiccv21,normavatarcvpr21,Piao_2021_CVPR,Ramon_2021_ICCV,personalizedeccv20} or $\Scale[0.9]{\mymap{h}{p}{}}$~\cite{RingNet,GCNcvprw20,R_2021_CVPR,Yenamandra_2021_CVPR} between the aligned shapes by nearest neighbor searching, 
(3) compute the NMSE of the nn-map $\Scale[0.9]{e\big(\mymap{p}{h}{}\big)}$ or $\Scale[0.9]{e\big(\mymap{h}{p}{}\big)}$.

\section{Motivation}

Multiple methods have been proposed to tackle the face reconstruction problem.
However, to the best of our knowledge, there does not exist an effective evaluation protocol to fairly and reliably compare reconstructed faces from different methods.
We observe the following issues in existing protocols:
{
\InsertBoxR{0}{\parbox{0.3\linewidth}{
\centering\hspace{0pt}
\begin{overpic}[trim=0cm 30cm 22cm 0cm,clip,width=1\linewidth,grid=false]{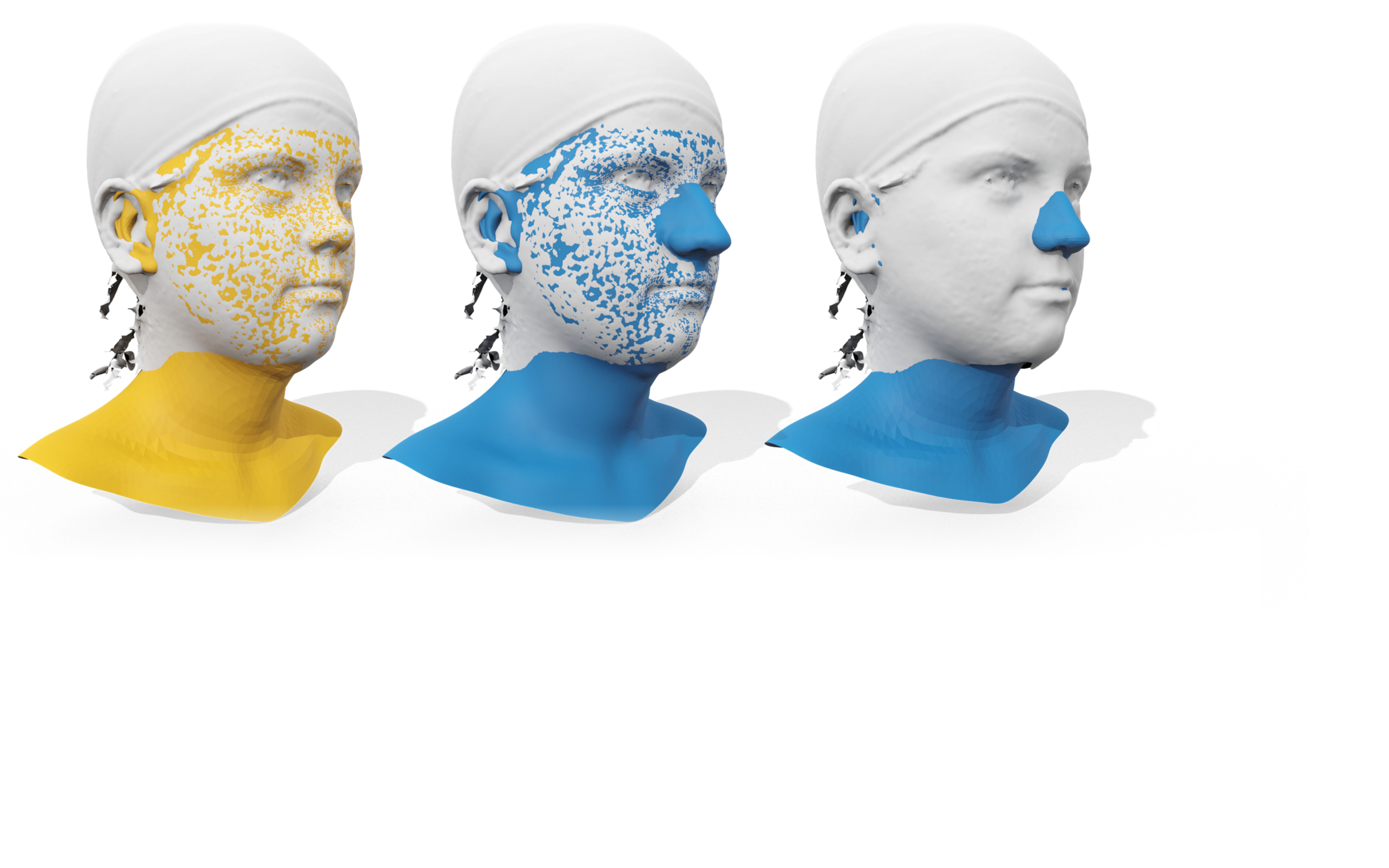}
\put(-1,42){\tiny\bfseries\itshape $S_H$}
\put(-1,15.5){\tiny\bfseries\itshape $S_L$}
\put(29,15.5){\tiny\bfseries\itshape $S_P$}
\put(60,15.5){\tiny\bfseries\itshape $S_P$}
\put(38,48.5){\tiny\bfseries\itshape \underline{G.T.} align}
\put(68,48.5){\tiny\bfseries\itshape \underline{ICP} align}
\end{overpic}
}}[1]
(1) Global alignment is extremely sensitive to the provided keypoints and local changes. 
The NMSE metric based on global alignment often fails to reflect the true shape difference. 
Take the inset figure as an example: we have the ground-truth high-res mesh $\Scale[0.9]{S_H}$ colored in white, and the ground-truth low-res mesh $\Scale[0.9]{S_L}$ colored in yellow. 
We then modify the nose region of $\Scale[0.9]{S_L}$ while keeping the rest part unchanged, which leads to a toy predicted mesh $\Scale[0.9]{S_P}$ colored in blue. Perceptually, we expect $\Scale[0.9]{S_P}$ to be aligned with $\Scale[0.9]{S_H}$ in the same way as $\Scale[0.9]{S_L}$ (as shown in the middle) to compute the shape differences.
However, ICP computes the alignment in a global way and as shown on the right, $\Scale[0.9]{S_P}$ is rotated backwards w.r.t. $\Scale[0.9]{S_H}$, having the complete facial region of $\Scale[0.9]{S_P}$ behind $\Scale[0.9]{S_H}$, which leads to exaggerated errors. In this case, considering region-based alignment can help to avoid global mismatch (see Fig.~\ref{fig:replace_sample}).
}

{
\InsertBoxR{0}{\parbox{0.25\linewidth}{
\centering\hspace{0pt}
    \centering
    \begin{overpic}[trim=1cm 13.5cm 25.4cm 1.6cm,clip,width=1\linewidth,grid=false]{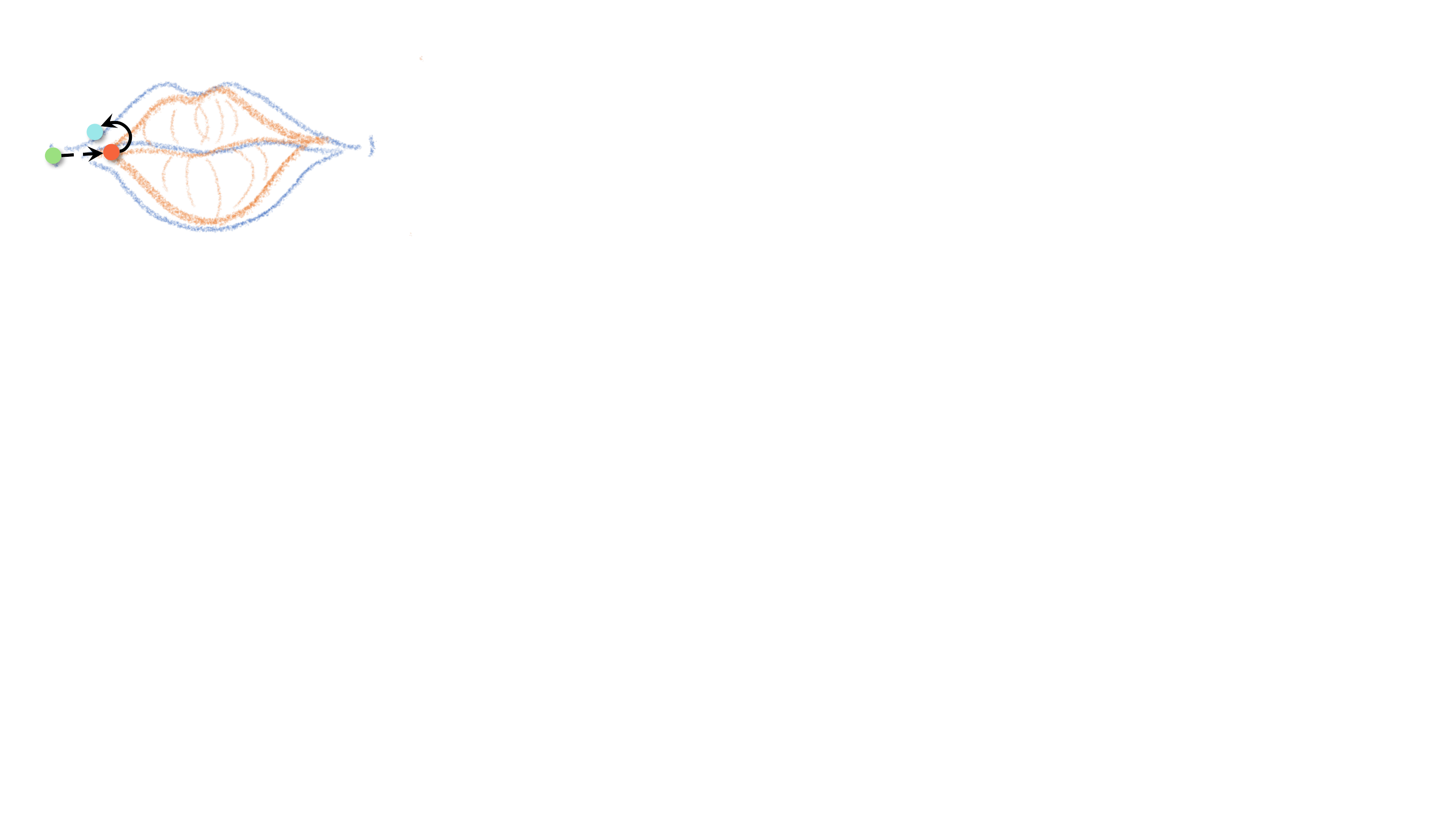}
    \put(20,18){\scriptsize\bfseries$x$}
    \put(10,39){\scriptsize\bfseries $y_1$}
    \put(1,18){\scriptsize\bfseries $y_2$}
    \put(80,40){\scriptsize \textcolor{blue!80}{$S_H$}}
    \put(80,13){\scriptsize \textcolor{orange}{$S_P$}}
    \end{overpic}
}}[1]
(2) Another limitation is that it is hard to establish accurate and meaningful correspondences based on the global \emph{rigid} alignment from a single direction (i.e., from $\Scale[0.9]{S_P}$ to $\Scale[0.9]{S_H}$ only). 
For example, the inset figure shows two aligned lips and we are supposed to measure the NMSE on the nn-map from $\Scale[0.9]{S_P}$ to $\Scale[0.9]{S_H}$. For the red vertex $\Scale[0.9]{x\in S_P}$, its nearest neighbor in $\Scale[0.9]{S_H}$ is the blue vertex $\Scale[0.9]{y_1}$. 
In this case, considering the map in the other direction, i.e., from $\Scale[0.9]{S_H}$ to $\Scale[0.9]{S_P}$, can help to establish the correct correspondence from $\Scale[0.9]{y_2}$ to $\Scale[0.9]{x}$.
}

These observations inspire us to consider \emph{region-based} and \emph{bidirectional} alignment for establishing semantically more meaningful correspondences between the predicted mesh $S_P$ to the ground-truth mesh $\Scale[0.9]{S_H}$.
However, to make this happen, meshes $\Scale[0.9]{S_H}$ with ground-truth region masks $\Scale[0.9]{\mathcal{R}_H}$ are requested, which are not available in any of the open datasets~\cite{FRGC2,MICC,facescape,LYHM,ICT3DRFE,beyond3dmmeccv20}.
To fill the gap, we propose a new benchmark {\name}, which provides ground-truth high-res mesh $\Scale[0.9]{S_H}$ and retopologized mesh $\Scale[0.9]{S_L}$ in {\nextone} topology, with accurate ground-truth keypoints and region masks.
To justify the usefulness of our benchmark, we show two applications: evaluating and comparing (1) the reconstruction quality of faces obtained from $9$ different methods in different choices of topology, (2) the expressiveness power of $8$ different 3DMM basis for face regression.

\myparagraph{Paper Structure}
Sec.~\ref{Sec:construction-data} discusses the details of {\name} benchmark and our 3DMM. 
Sec.~\ref{sec:evaluation} illustrates our novel region-aware bidirectional evaluation pipeline. 
In Sec.~\ref{sec:results} we extensively justify the usefulness of {\name} and the advantages of our evaluation pipeline over the standard one on face reconstruction task, and demonstrate that {\nextpp} is more expressive than existing 3DMMs.

\section{{\name}: A New 3D Face Benchmark} \label{Sec:construction-data}
\myparagraph{Overview}
Our benchmark {\name} contains $100$ individuals, and each individual is modelled by a ground-truth high-resolution mesh $\Scale[0.9]{S_H}$ (the aligned 3D scan from LYHM~\cite{LYHM}) and a retopologized low-resolution mesh $\Scale[0.9]{S_L}$ using {\nextone} \cite{hifi3dface2021tencentailab} topology in neutral expression (see Fig.~\ref{fig:eg_benchmark} for some examples). 
The meshes $\Scale[0.9]{(S_H, S_L)}$ of all individuals are consistently \emph{scaled} and \emph{aligned}. 
We also provide $68$ keypoints and $4$ region masks, which are semantically meaningful, for \emph{both} $\Scale[0.9]{S_H}$ and $\Scale[0.9]{S_L}$ of \emph{each} individual. 
For each individual, five high-quality and realistic multi-view images (including one frontal image) are rendered with well designed lighting condition and ground-truth camera parameters.

We explain the detailed construction procedure of {\name} as follows.

{
\myparagraph{{\nextone} Topology}
We choose this topology for our benchmark for the following reasons:
(1) LYHM has overdense samplings at the boundary of the eyes and mouth.
(2) LSFM does not have edge loops to define the contours of the eyes and mouth.
(3) FLAME has unnatural triangulation which cannot model some realistic muscle movements such as raising the eyebrows.
As a comparison, {\nextone} has better triangulation and balanced samplings to make realistic and nuanced expressions. Besides, {\nextone} also has eyeballs, interior structure of the mouth, and the shoulder region, which all benefit downstream applications such as talking head.
See supplementary for visualized comparisons.
}

\myparagraph{Construction Pipeline} 
We start by collecting $1235$ scans from LYHM and preparing a template shape {\tempshape} in {\nextone} topology (see Fig.~\ref{fig:autowrap}(c)) with predefined $68$ keypoints {\tempkpt} and $4$ region masks {\tempregion} (including nose, mouth, forehead and cheek region). 
Firstly, we re-scale and rigidly align the input scans to the template shape {\tempshape}, leading to our ground-truth high-resolution meshes $\Scale[0.9]{S_H}$ (i.e., aligned scans). 
We then follow~\cite{FengBenchmark,RingNet} to define an evaluation region which is a disk centered on the nose tip.
Secondly, we ``wrap'' (i.e., perform non-rigid registration) {\tempshape} to each $\Scale[0.9]{S_H}$ to get the retopologized $\Scale[0.9]{S_L}$ such that $\Scale[0.9]{S_L}$ have the same topology as {\tempshape} but reflect the shape of $S_H$. Note that we have keypoints $\Scale[0.9]{\mathcal{K}_{L} = \mathcal{K}_{\text{temp}}}$ and regions $\Scale[0.9]{\mathcal{R}_{L} = \mathcal{R}_{\text{temp}}}$ since $\Scale[0.9]{S_L}$ and {\tempshape} share the same {\nextone} topology.
We then transfer $\Scale[0.9]{\mathcal{K}_{L}}$ and $\Scale[0.9]{\mathcal{R}_{L}}$ from $\Scale[0.9]{S_L}$ to $\Scale[0.9]{S_H}$ for each individual.
We also set up a rendering pipeline for synthesizing \emph{multi-view} images for the textured high-resolution mesh $\Scale[0.9]{S_H}$. 
Such a controlled environment enables {\name} to focus on reflecting the reconstruction ability of different methods.
Finally, we filter out samples with wrapping error larger than $0.2$mm and ask an expert artist with $3$ year modeling experience to select $100$ individuals among all the processed scans with the highest model quality, across different genders, ethnicity, and ages, to obtain our {\name} benchmark.

\myparagraph{Challenges \& Solutions} We observe two major challenges during the above construction procedure: (1) the raw scans are in different scales and poses with inaccurate sparse keypoints~\cite{lmk_detect_2012}, which makes it difficult to align them consistently. 
To tackle this problem, we iterate through the following steps until convergence: first, render a frontal face image of $\Scale[0.9]{S_H}$ with texture using the initial/estimated transformation to align $\Scale[0.9]{S_H}$ to $\Scale[0.9]{S_{\text{temp}}}$ (note that the frontal pose needs to be determined from the alignment transformation as the frontal facing pose is unknown for a given scan); second, detect a set of 2D facial keypoints on the rendered image of $\Scale[0.9]{S_H}$ using state-of-the-art landmark detector; third, project the 2D keypoints into 3D using the rendering camera pose; fourth, update the alignment transformation from $\Scale[0.9]{S_H}$ to $\Scale[0.9]{S_{\text{temp}}}$ using the correspondences between the projected 3D keypoints on $\Scale[0.9]{S_H}$ and the known 3D keypoints on $\Scale[0.9]{S_{\text{temp}}}$.

(2) Another challenge is, after we get the retopologized $\Scale[0.9]{S_L}$, how to accurately transfer the region mask from the low-resolution mesh $\Scale[0.9]{S_L}$ (inherited from $\Scale[0.9]{S_{\text{temp}}}$) to the high-resolution mesh $\Scale[0.9]{S_H}$. 
One naive solution would be using nearest neighbor mapping from $\Scale[0.9]{S_L}$ to $\Scale[0.9]{S_H}$ to transfer the region mask. However, since the resolution of $\Scale[0.9]{S_H}$ can be $50\times$ larger than that of $\Scale[0.9]{S_L}$, this naive solution will introduce disconnected and noisy region mask. 
To avoid such flaws, we use the vertex-to-point mapping from both directions to find candidate regions on $\Scale[0.9]{S_H}$. 
As much more correspondences can be established during the mapping from $\Scale[0.9]{S_H}$ to $\Scale[0.9]{S_L}$, higher-quality, smoother region masks can be obtained.
Finally, we filter out noisy regions (e.g., nostril, eyeballs) and return the largest connected region.

{\myparagraph{{\nextpp}}
With the above procedure, we can further construct a 3DMM basis by retopologizing more 3D face models. 
Specifically, based on the $200$ individuals from {\nextone}~\cite{hifi3dface2021tencentailab}, we additionally process and retopologize 3D face models of $846$ individuals from FaceScape dataset~\cite{facescape} into {\nextone} topology. Together with the aforementioned processed models of $1235$ individuals from LYHM~\cite{LYHM}, we collect and then select $1957$ most representative meshes consisting of individuals from various ethnic groups. We then apply PCA~\cite{3DMM} to obtain our new basis with $526$ PCs (with $99.9\%$ cumulative explained variance)}, which we name as {\nextpp}. 
Tab.~\ref{tab:3dmm_topo} shows the comparison of {\nextpp} to other 3DMMs.
Note that previous 3DMMs are more or less ethnics-biased. For example, BFM~\cite{bfm09} is constructed mostly from Europeans, FLAME~\cite{flametopo} is constructed from scans of the US and European, while {\nextone}~\cite{hifi3dface2021tencentailab} and FS~\cite{facescape} is constructed from scans of Asians.
The LSFM and FLAME contains $50:1$ and $12:1$ of Caucasian and Asian respectively.
In contrast, {\nextpp} is constructed from high-quality models across more balanced ethnic groups that ensures $1:1$ between Caucasian and Asian (plus a few subjects from other ethnicities). 
We examine the expressive powers and reconstruction qualities of different 3DMMs in Sec.~\ref{Sec:new3dmm}.

\input{latex/algos/alg_ricp.tex}
\input{latex/algos/alg_bicp.tex}

\section{A Novel Evaluation Pipeline}\label{sec:evaluation}
To evaluate the quality of a reconstructed or predicted face $\Scale[0.9]{S_P}$, the standard pipeline first globally aligns $\Scale[0.9]{S_P}$ with the ground-truth high-resolution mesh $\Scale[0.9]{S_H}$ to find the nearest-neighbor map $\Scale[0.9]{\mymap{p}{h}{}}$ (or $\Scale[0.9]{\mymap{h}{p}{}}$). 
The similarity or reconstruction error is then measured by the NMSE error $\Scale[0.9]{e\big(\mymap{p}{h}{}\big)}$ (or $\Scale[0.9]{e\big(\mymap{h}{p}{}\big)}$). 
We propose a new evaluation pipeline based on a region-aware and bidirectional alignment.

We focus on how to accurately establish correspondences between $\Scale[0.9]{S_P}$ and a particular region $\Scale[0.9]{\mathcal{R}_H}$ in the ground-truth shape $\Scale[0.9]{S_H}$ (denoted as $\Scale[0.9]{S_H@\mathcal{R}_H}$ for short), which consists of two main steps:
(1) {\ricp} (region-aware ICP): we first get the rigidly transformed shape $\Scale[0.9]{S_P^*}$ from $\Scale[0.9]{S_P}$ by aligning $\Scale[0.9]{S_P}$ to $\Scale[0.9]{\mathcal{R}_H}$ such that the corresponding \emph{region} on $\Scale[0.9]{S_P}$ is well aligned to $\Scale[0.9]{\mathcal{R}_H}$ without taking the rest part of the face into consideration (see Algo.~\ref{Alg:myicp}).
(2) {\bicp} (non-rigid and bidirectional ICP): with the above established correspondences between $\Scale[0.9]{S_P^*}$ and $\Scale[0.9]{\mathcal{R}_H}$ as initialization, we further refine the correspondences by applying non-rigid ICP ({\nicp})~\cite{NICP} to deform $\Scale[0.9]{\mathcal{R}_H}$ to fit $\Scale[0.9]{S_P^*}$ (see Algo.~\ref{Alg:bicp}).
This step yields a deformed shape $\Scale[0.9]{\mathcal{R}_H^*}$ from $\Scale[0.9]{\mathcal{R}_H}$, as well as the correspondences between $\Scale[0.9]{\mathcal{R}_H}$ and $\Scale[0.9]{S_P^*}$ induced from $\Scale[0.9]{\mathcal{R}_H^*}$ (using vertex-to-surface projection).
Our region-wise alignment and the two step coarse-to-fine registration effectively guarantee that {\nicp} can converge to a reasonable deformed shape $\Scale[0.9]{\mathcal{R}_H^*}$ (see supplementary for details).
The resulting correspondences are used for computing errors between $\Scale[0.9]{\mathcal{R}_H}$ and $\Scale[0.9]{S_P^*}$.
Note that the second alignment step can be regarded as upsampling $\Scale[0.9]{S_P^*}$ such that it has a similar resolution to the dense ground-truth scan $\Scale[0.9]{S_H@\mathcal{R}_H}$, which makes the evaluation of reconstructed meshes in different resolutions easier.

We observe that the correspondences established between the rigidly transformed shape $\Scale[0.9]{S_{P}^*}$ and deformed region $\Scale[0.9]{\mathcal{R}_H^*}$ are more accurate than the correspondences detected between the original shapes $\Scale[0.9]{S_{P}}$ and $\Scale[0.9]{S_H}$ for evaluating the similarity between the two shapes in region $\Scale[0.9]{\mathcal{R}_H}$.
Fig.~\ref{Fig:corr_eg} shows such an example. We visualize the corresponding points on $\Scale[0.9]{S_H}$ of the mouth keypoints on $\Scale[0.9]{S_P}$ in blue/green/yellow established by {\gicp}/{\ricp}/{\bicp} respectively, where the red points are the ground-truth mouth keypoints on $\Scale[0.9]{S_H}$.

\setlength{\intextsep}{5pt}%
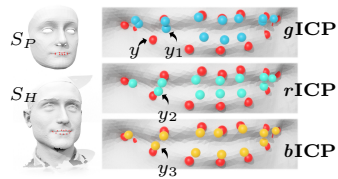
\begin{wrapfigure}{r}{0.35\linewidth}
\centering
\begin{overpic}[trim=1cm 42cm 60cm 0cm,clip,width=1\linewidth,grid=false]{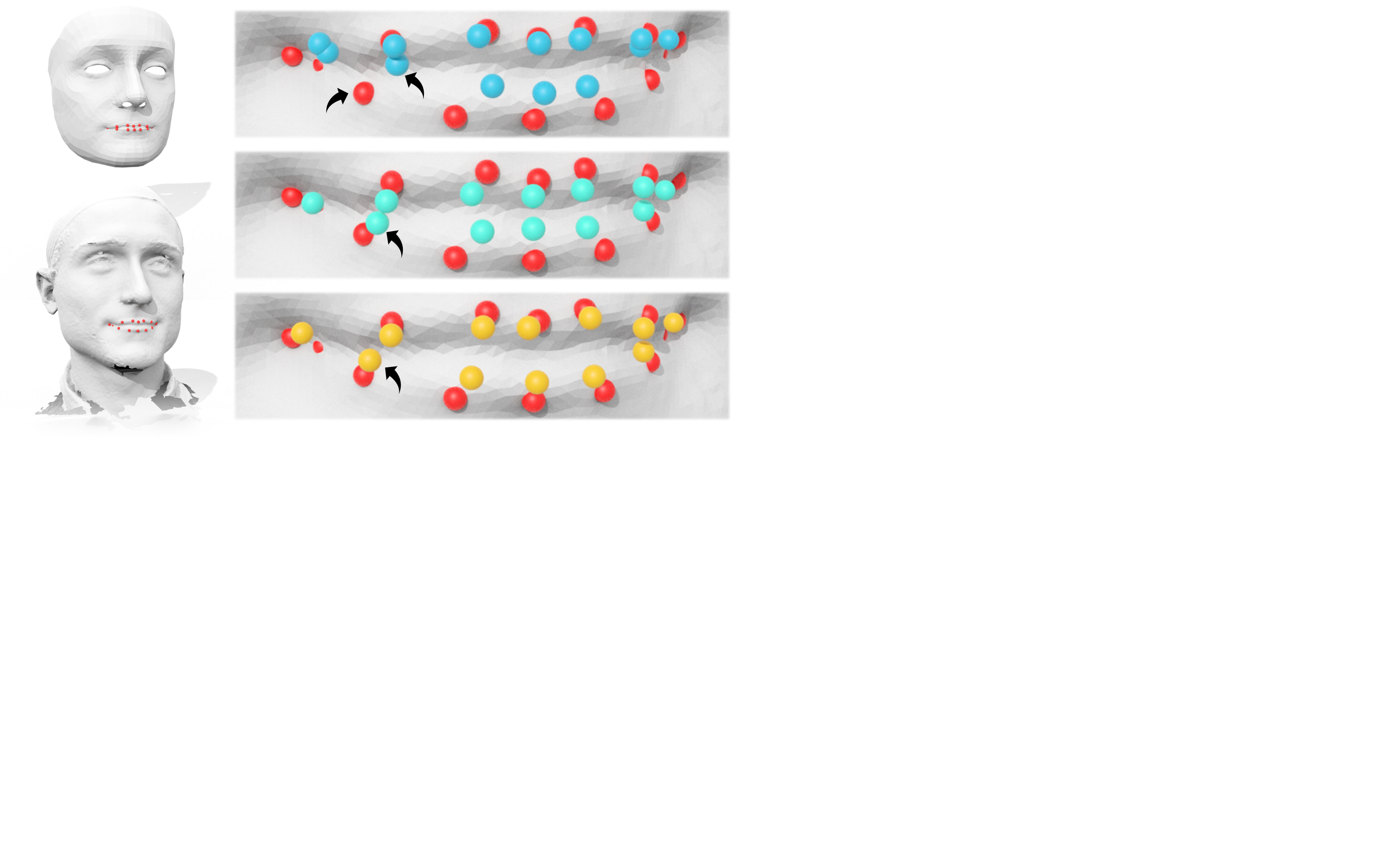}
    \put(-1,45){\scriptsize\bfseries $S_P$}
    \put(-1,28){\scriptsize\bfseries $S_H$}
    \put(36,42){\scriptsize\bfseries $y$}
    \put(47,42){\scriptsize\bfseries $y_1$}
    \put(45,22){\scriptsize\bfseries $y_2$}
    \put(45,5){\scriptsize\bfseries $y_3$}
    \put(84,48){\scriptsize\bfseries \textit{g}ICP}
    \put(84,29){\scriptsize\bfseries \textit{r}ICP}
    \put(84,10){\scriptsize\bfseries \textit{b}ICP}
    \end{overpic}
    \caption{Correspondence established by different ICPs and the GT correspondences (red).}
    \label{Fig:corr_eg}
\end{wrapfigure}
We can see that {\bicp} gives more accurate correspondences since it focus on the mouth region and considers correspondences from both directions. As a result, $\Scale[0.9]{y_3}$ obtained by {\bicp} is closer to the ground-truth $\Scale[0.9]{y}$, while $\Scale[0.9]{y_1}$ obtained by {\gicp} is far from $\Scale[0.9]{y}$. We validate the above analysis with detailed experiments in Sec.~\ref{sec.evalcorresp}.
We use the above established correspondences to measure the NMSE error of the region-wise aligned $\Scale[0.9]{S_{P}^*}$ compared to ground-truth $\Scale[0.9]{S_H}$ on four predefined regions including nose, mouth, forehead, and cheek, respectively (see Fig.~\ref{fig:autowrap}d).
The NMSE error is then transformed back to the physical scale of the raw scan in millimeters (note that each ground-truth $\Scale[0.9]{S_H}$ in our benchmark is rigidly transformed from raw scan, see Sec.~\ref{Sec:construction-data}).
Evaluation on each region individually provides us fine-grained understandings of the qualities of the reconstructed meshes. We present extensive experiments for benchmarking state-of-the-art single-image 3D face reconstruction methods in Sec.~\ref{Sec:eval-all}.

\section{Experiment}\label{sec:results}
In this section we first demonstrate the effectiveness of our {\bicp} which can establish more accurate correspondences than standard ICP for reliably evaluating 3D face reconstructions.
We then compare different face reconstruction methods and 3DMMs using our evaluation protocol to investigate fine-grained shape differences based on regions in a systematically way.

\subsection{Ablation Study: {\bicp} v.s. {\gicp}}\label{sec.evalcorresp}
To demonstrate the advantages of our region-based and bidirectional evaluation protocol over the standard one, we design a controlled experiment illustrated in Fig.~\ref{fig:replace_sample} and Tab.~\ref{tab:replace_exp} and \ref{tab:icp_cmp_exp}. 
Specifically, we carefully construct four predicted shapes $\Scale[0.9]{S_P}$ by modifying the same ground-truth mesh $\Scale[0.9]{S_L}$ such that the ground-truth correspondences between $\Scale[0.9]{S_P}$ and $\Scale[0.9]{S_L}$ are known for reference.

\begin{table}[!t]
\scriptsize
    \input{latex/tables/table_toy_exp}
\end{table}

As illustrated in Fig.~\ref{fig:replace_sample}, we replace the nose/mouth/forehead/cheek region of shape $\Scale[0.9]{S_L}$ with the corresponding region from four different reference shapes $\Scale[0.9]{S_i (i=1,2,3,4)}$ respectively. We then visualize the shape differences computed using {\bicp} (w.r.t. the four specified regions) and {\gicp} (w.r.t. the complete face) in Fig.~\ref{fig:replace_sample} where large (small) errors are colored in red (blue). 

We compare our evaluation pipeline to the standard one in two-fold: (1) we report the alignment error, the distance between the aligned $\Scale[0.9]{S_P}$ and the input $\Scale[0.9]{S_L}$ using ground-truth correspondences in Tab.~\ref{tab:replace_exp}, where the alignment is computed using our region-wise {\ricp} and the global-wise {\gicp}. (2) we report the accuracy of correspondences established via {\gicp}, {\ricp} (our intermediate step), and {\bicp} in Tab.~\ref{tab:icp_cmp_exp} compared to the ground-truth correspondences.

\setlength{\columnsep}{5pt}%
\setlength{\intextsep}{4pt}%
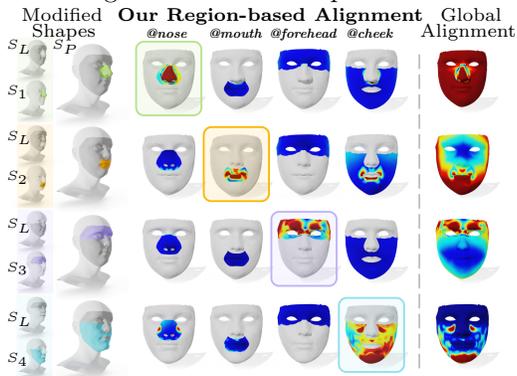
\begin{wrapfigure}{r}{0.55\linewidth} 
\centering
\input{latex/img_camera_ready_lowres/figtex_region_replace}
\caption{We visualize the shape difference between $\Scale[0.9]{S_P}$ and $\Scale[0.9]{S_L}$ using our {\bicp} (shown on separate regions) and {\gicp}, where $\Scale[0.9]{S_P}$ is constructed by replacing $\Scale[0.9]{S_L}$'s nose/mouth/forehead/cheek region with the corresponding region from $\Scale[0.9]{S_1/S_2/S_3/S_4}$.
}\label{fig:replace_sample}
\end{wrapfigure}
We can see that {\gicp} is extremely sensitive to local changes. For example, replacing the nose only can lead to errors in the complete face (first row in Fig.~\ref{fig:replace_sample}) and replacing the mouth can lead to errors even in forehead (second row) according to {\gicp}. As a comparison, our {\bicp} can correctly localizes the shape differences in the modified regions and lead to more accurate alignment than {\gicp} as shown in Tab.~\ref{tab:replace_exp}.
This suggests that region-based alignment can better quantify shape differences especially in the case of local or subtle changes.
Moreover, Tab.~\ref{tab:icp_cmp_exp} shows that both our {\ricp} and {\bicp} help to find more accurate correspondences to evaluate shape differences.

\subsection{Evaluating Face Reconstruction Methods} \label{Sec:eval-all}
We compare recent state-of-the-art face reconstruction methods on our {\name} benchmark using our new evaluation protocol including: (1) linear-3DMM based methods: ExpNet~\cite{CNNRegress,FacePoseNet,ExpNet}, RingNet~\cite{RingNet}, MGCNet~\cite{MGCNet}, Deep3D~\cite{deep3d}, 3DDFA-v2~\cite{3DDFA_v2}, GANFit~\cite{ganfit,gecer2021fast}, DECA-coarse~\cite{DECA}, and (2) non(linear)-3DMM methods: PRNet~\cite{PRNet}, Nonlinear 3DMM (N-3DMM)~\cite{Nonlinear3DMM,tran2018on,tran2019towards}.

We report the statistics (mean and std.) of errors 
over $100$ shapes in {\name} using our {\bicp} (in each separate region and complete face) and using standard {\gicp} (in complete face) in Tab.~\ref{tab:main_result}. {\gicp} suggests that DECA and Deep3D are the best among the tested methods for face reconstruction. However, our {\bicp} suggests that DECA models the nose region in a much better and more accurate way than others, 
but it obtains less satisfactory result in the mouth region.
Fig.~\ref{fig:res_mtd} shows some qualitative results, where the best reconstructed face selected using {\bicp} ({\gicp}) is highlighted in orange (blue\&purple) box. We also conduct user study to ask people to vote for the best (labeled $\star$) and second best (labeled $\dagger$) face. We can see that the faces selected by {\bicp} are indeed visually more similar (validated by our user study) to the G.T. shapes than those selected by {\gicp}.

\begin{figure*}[!t]
\input{latex/img_camera_ready_lowres/figtex_res_mtd}
\caption{\textbf{Comparing different face reconstruction methods}. 
We visualize the reconstruction error of each face using the standard evaluation pipeline (top left) and our novel evaluation pipeline (bottom left, shown in four regions), where large (small) errors are colored in red (blue). 
Note only the cropped region shown in G.T. is counted for evaluation.
The best reconstructed face selected by our measurement (orange boxes) is visually closer to the ground-truth meshes than the ones selected using the standard measurements (blue \& purple boxes). 
See more examples in our supplementary.
}
\label{fig:res_mtd}
\end{figure*}
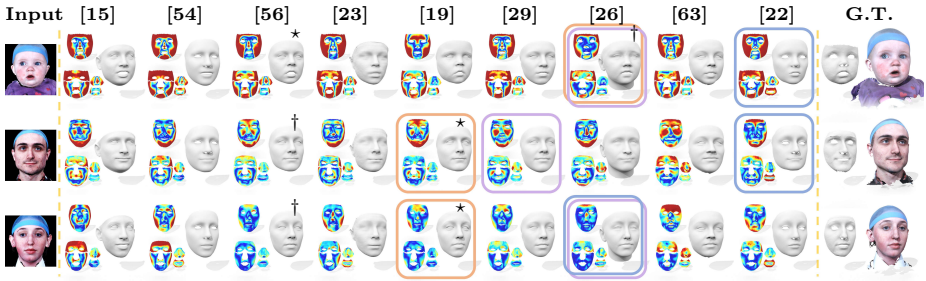

\begin{table*}[!t]
\caption{\textbf{Comparing different face reconstruction methods on {\name}.} We report the statistics of errors measured using our new pipeline ({\bicp}) and the standard one ({\gicp} from both directions). The best (second best) method w.r.t. average error is highlighted in red (blue).}\label{tab:main_result}

\input{latex/tables/table_main_result}
\end{table*}

\begin{figure*}[!t]
\begin{minipage}[t]{0.34\textwidth}
\centering
\input{latex/img_camera_ready_lowres/figtex_eg_merge}
\caption{We use our region-aware metric to find the best matched nose $\Scale[0.9]{\mathcal{R}_N}$, mouth $\Scale[0.9]{\mathcal{R}_M}$, cheek $\Scale[0.9]{\mathcal{R}_C}$, and forehead $\Scale[0.9]{\mathcal{R}_F}$ from existing methods, and merge them into a new face. Our merged face is more similar to the G.T. shape than DECA, the best reconstructed face voted by {\gicp}.} %
\label{fig:eg_merge}
\end{minipage}
\hfill
\begin{minipage}[t]{0.64\textwidth}
\input{latex/img_camera_ready_lowres/figtex_eg_convergence}
\caption{Comparing the fitting errors of different 3DMMs. We visualize an example of the RGB-D fitting errors during the optimization iterations using different 3DMMs and the corresponding reconstructed faces. We observe that {\nexttwo} and {\nextpp} give the most realistic reconstructed faces (especially in the mouth region) and show superior converging rate with smallest converged loss among the tested 3DMMs. As a comparison, FS and LSFM show instability during the fitting process and lead to erroneous reconstruction results.}
\label{fig:convergence}
\end{minipage}
\end{figure*}

\begin{table}[!t]
\caption{\textbf{Comparing different 3DMM basis on {\name}.} The best (second best) methods w.r.t. average error are highlighted in red (blue).} \label{tab:rgbd_result}

\input{latex/tables/table_rgb_fitting_result}
\end{table}

Moreover, another advantage of our {\bicp} evaluation protocol is its region-aware nature, which allows us to compare different methods in some particular region.
Fig.~\ref{fig:eg_merge} shows such an example, where we select the best matched region from different methods according to our region-aware {\bicp} and merge them into a new face. Perceptually, the merged face is clearly better than the face reconstructed using DECA (selected by {\gicp}).

\subsection{Evaluating Different 3DMMs} \label{Sec:new3dmm}

We use our new evaluation approach to compare different 3DMMs on {\name} including: LYHM~\cite{LYHM}, BFM~\cite{bfm09}, FLAME~\cite{flametopo}, LSFM~\cite{3dmm10000}, FaceScape basis (FS)~\cite{facescape}, {\nextone} and {\nexttwo}~\cite{hifi3dface2021tencentailab}. 
In this test, we use different basis and run standard RGB(-D) fitting algorithm~\cite{hifi3dface2021tencentailab} using photo loss (with/without depth loss), ID loss, landmark loss and regularization loss to regress the 3DMM coefficients from the given 2D images provided by {\name}.
For RGB fitting, we use a frontal face image of each individual as input.
For RGB-D fitting, a frontal rendered depth image in addition to the RGB image is used. 
Fig.~\ref{fig:convergence} shows the fitting errors over iterations using different 3DMMs. 
The qualitative and quantitative comparisons are presented in Fig.~\ref{fig:res_basis} and Tab.~\ref{tab:rgbd_result}, respectively.
Note that both LYHM and our 3DMM basis in Tab.~\ref{tab:rgbd_result} use some test shapes in {\name} to construct the 3DMM. They are listed in the table only for reference.

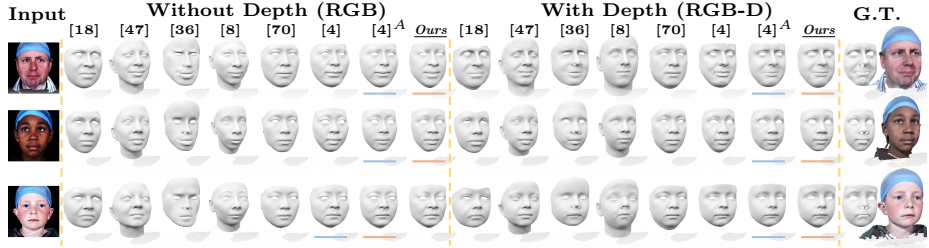
\begin{figure*}[!t]
\input{latex/img_camera_ready_lowres/figtex_res_basis}
\caption{\textbf{Comparing different 3DMMs}. From \emph{Left to right}: LYHM~\cite{LYHM}, BFM~\cite{bfm09}, FLAME~\cite{flametopo}, LSFM~\cite{3dmm10000}, FS~\cite{facescape}, {\nextone}~\cite{hifi3dface2021tencentailab}, {\nexttwo}~\cite{hifi3dface2021tencentailab}, and {\nextpp}. We highlight the best (second best) reconstructed face via red (blue) underline. Only the cropped region shown in G.T. is counted for evaluation. See more examples in supplementary.}\label{fig:res_basis}
\end{figure*}

As shown in Tab.~\ref{tab:rgbd_result}, {\nexttwo}, {\nextone}, and FS achieve similar performance in RGB fitting, while {\nexttwo} and {\nextone} perform much better than the other 3DMMs in RGB-D fitting, indicating that {\nexttwo} and {\nextone} are more expressive and can better fit the geometry especially in RGB-D fitting with extra depth information.
To directly justify the expressive power of our new 3DMM basis {\nextpp}, we compare {\nextpp} to {\nextone}/{\nexttwo} over $3$ shapes that are unused by all three 3DMMs. We fit the 3D ground-truth scans using the 3DMMs. Fig.~\ref{fig:eg_next_basis} shows the error heatmaps. 
The average error with our new basis decreases over $20\%$ compared to {\nexttwo}.

Finally, an important observation from Tabs.~\ref{tab:main_result}-\ref{tab:rgbd_result} is that, while top performing single image reconstruction methods achieve results of $1.657 \sim 2.953$, the best RGB-D fitting records $0.878$, which is far better than them. It implies that there is still much room for improvements in single image reconstruction methods.

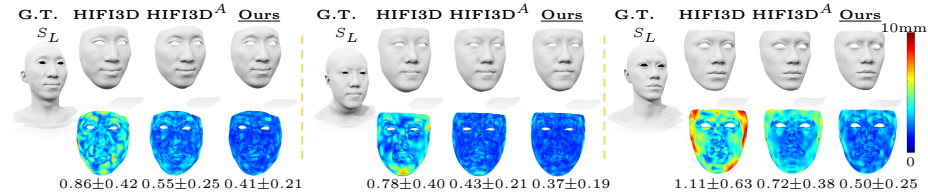
\begin{figure*}[t!]
\centering
\input{latex/img_camera_ready_lowres/figtex_three_next}
\caption{\textbf{Comparing our 3DMM to {\nextone} and {\nexttwo}}.\emph{Top}: Fitted faces $S_P$, \emph{Bottom}: Fitting errors, where large (small) errors are colored in red (blue).}
\label{fig:eg_next_basis}
\end{figure*}

\section{Conclusions}

In this work, we introduce a new benchmark {\name} for 3D face reconstruction that provides
accurate and consistent facial keypoints, region masks on the scans, and consistently retopologized meshes. %
During the construction procedure of the benchmark, we also derive a new powerful 3DMM basis {\nextpp}.
The benchmark allows us to design a novel region-aware and bidirectional evaluation pipeline to measure shape similarity, which is justified to be more reliable than the standard evaluation pipeline based on global alignment. 
Furthermore, we compare and analyse existing single-image face reconstruction methods and state-of-the-art 3DMM basis using our new evaluation approach on {\name}, which is the first to obtain fine-grained region-wise analyses in the 3D face community. Moreover, it would be interesting to research how our benchmark can be used for supervised learning of face reconstruction.

\noindent \textbf{Acknowledgment.}
This work was supported by SZSTC Grant No. JCYJ20190 809172201639 and WDZC20200820200655001, Shenzhen Key Laboratory ZDSY S20210623092001004.

}
\clearpage

\appendix
\title{Supplementary Material for\\ \name: Rethinking the Evaluation of 3D Face Reconstruction} 

\titlerunning{\name: Rethinking the Evaluation of 3D Face Reconstruction}
%
\author{Zenghao Chai\inst{1}$^{*}$\orcidlink{0000-0003-3709-4947} \and
Haoxian Zhang\inst{2}$^{*}$\orcidlink{0000-0001-7078-868X} \and
Jing Ren\inst{2}\orcidlink{0000-0003-3114-3517} \and Di Kang\inst{2}\orcidlink{0000-0002-8996-0897} \and Zhengzhuo Xu\inst{1}\orcidlink{2222--3333-4444-5555} \and Xuefei Zhe\inst{2}\orcidlink{0000-0002-5005-7166} \and Chun Yuan\inst{1,3}$^\dagger$\orcidlink{0000-0002-3590-6676} \and Linchao Bao\inst{2}$^\dagger$\orcidlink{0000-0001-9543-3754} } 
\authorrunning{Z. Chai et al.}
%
\institute{Shenzhen International Graduate School, Tsinghua University, China \and
Tencent AI Lab, China, $^3$ Peng Cheng National Laboratory, China \\
\footnotetext{$^*$ Equal Contributions.}
\footnotetext{$^\dagger$ Corresponding authors: \email{yuanc@sz.tsinghua.edu.cn; linchaobao@gmail.com}.}}

\maketitle


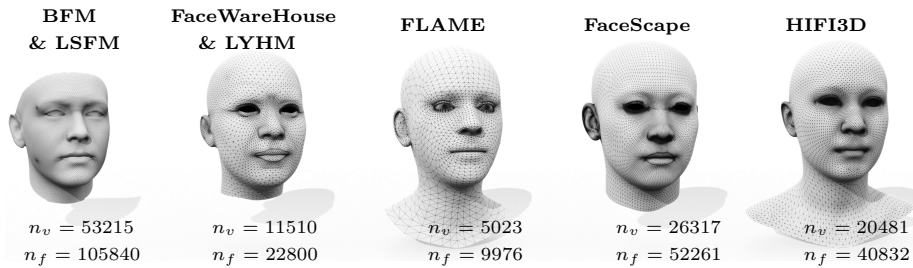
\begin{figure*}[h!]
\centering
\input{latex/img_camera_ready_supp_lowres/figtex_3dmm_topo}
\caption{Topology of different 3DMMs, where $n_v$ and $n_f$ represent the number of vertices and triangles respectively. In our {\name} benchmark, we choose {\nextone} topology since it has better triangulation and balanced samplings with eyeballs and shoulder regions.}\label{fig:3dmm_topo}
\end{figure*}

In this supplementary material, we provide additional technical details, qualitative examples, and discussions
that could not be fitted into the main paper due of lack of space, which is organized as follows: 
We first give full details of how to construct our new benchmark {\name} in Sec.~\ref{append:benchmark}.
Additional experiments and results can be found in Sec.~\ref{append:res}, where we justify the quality of {\name}.
In Sec.~\ref{append:implementation} we discuss the implementation details and choices of parameters. Finally, we discuss the limitation and future work in Sec.~\ref{append:limitaion}.

\section{Details of Constructing {\name} Benchmark}\label{append:benchmark}

\subsection{Preparing the Template Shape {\tempshape}} \label{append:keypointdefine}

We first prepare a template shape {\tempshape} which is crucial for registering and retopologizing high-resolution scans from different datasets. 
We take the mean shape from {\nextone}~\cite{hifi3dface2021tencentailab} as our template shape that contains $20481$ vertices and $40382$ triangles. 
We then ask an \emph{experienced} artist to label semantically meaningful and important keypoints {\tempkpt} and region masks {\tempregion} since they play an important role in our proposed {\bicp} based face similarity evaluation.

\myparagraph{{\nextone} Topology} We choose {\nextone} for the following reasons:
(1) BFM (LSFM) does not have edge loops to define the contours of the eyes and mouth.
(2) FaceWareHouse (LYHM) has overdense samplings around the boundary of the eyes and mouth.
(3) FLAME has unnatural triangulation which cannot model some realistic muscle movements such as raising the eyebrows.
(4) FaceScape does not have eyeballs, interior structure of the mouth, or the shoulder region, which limits the expressiveness of different expressions. 
As a comparison, {\nextone} has better triangulation and balanced samplings to make realistic and nuanced expressions. 
Besides, {\nextone} also has independent eyeballs, interior structure of the mouth, and the shoulder region, which all benefit downstream applications such as talking head generation. 
Please see Fig.~\ref{fig:3dmm_topo} for the topology of each 3DMM mentioned in Tab.~1 in the main paper.

\begin{figure}[!t]
    \centering
    \begin{overpic}[trim=0cm 53cm 42cm 0cm,clip,width=0.8\linewidth,grid=false]{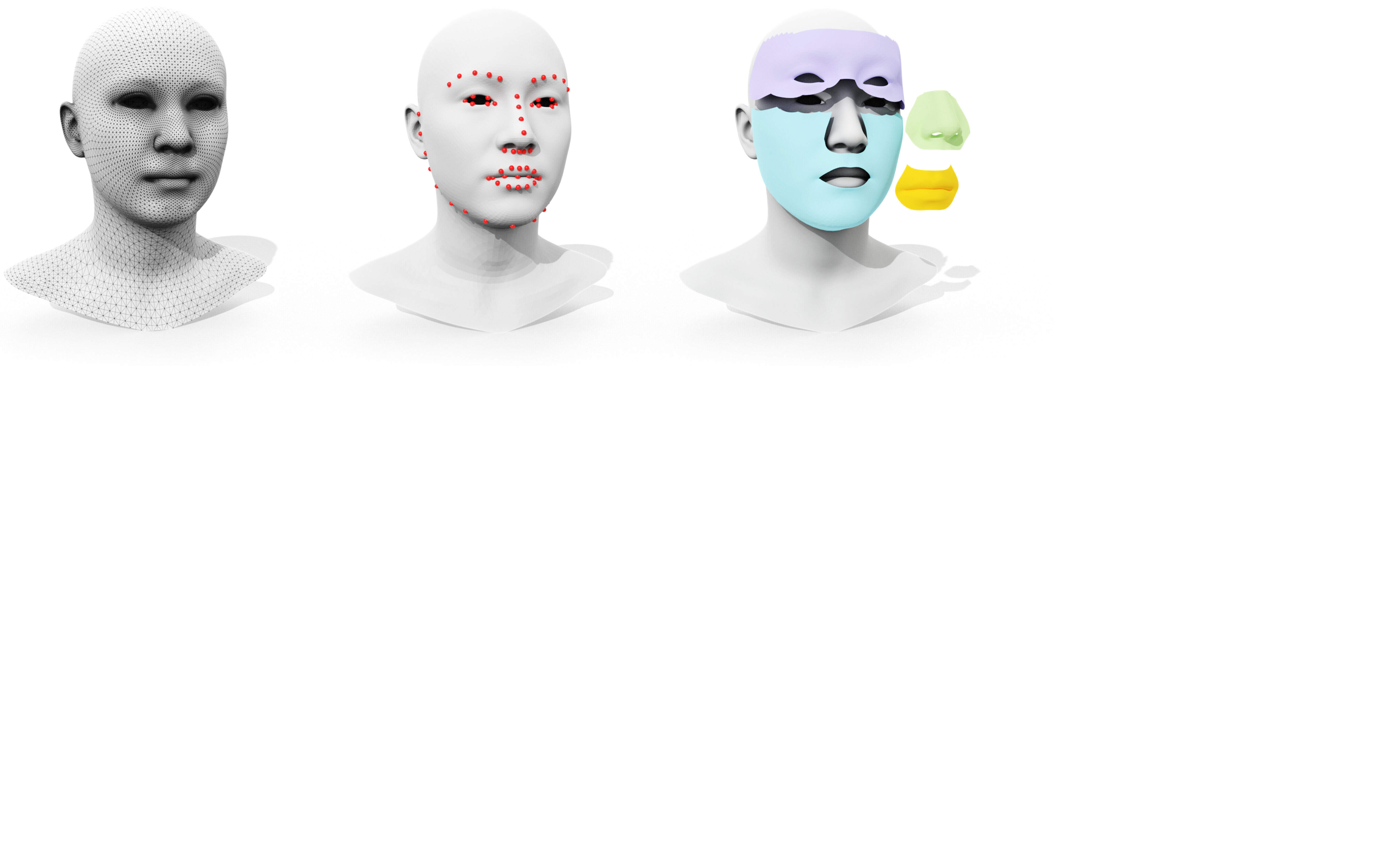}
    \put(0,36){\scriptsize\bfseries Template Shape {\tempshape}}
    \put(39,36){\scriptsize\bfseries Keypoints {\tempkpt}}
    \put(71,36){\scriptsize\bfseries Region Masks {\tempregion}}
    \put(71,29){\scriptsize $\mathcal{R}_{F}$}
    \put(71,17){\scriptsize $\mathcal{R}_{C}$}
    \put(94,28){\scriptsize $\mathcal{R}_{N}$}
    \put(94,11.5){\scriptsize $\mathcal{R}_{M}$}
    \end{overpic}
    \caption{We predefine $68$ keypoints and $4$ region masks on the template shape for constructing {\name}.}
    \label{fig:append:template_shape}
\end{figure}

\myparagraph{Region Masks}
Four region masks are annotated in the {\tempshape}, namely the nose region $\mathcal{R}_N$, the mouth region $\mathcal{R}_M$, the forehead region $\mathcal{R}_F$, and the cheek region $\mathcal{R}_C$. 
Each region mask is defined as a list of vertices and faces with smooth boundary (thanks to the good structural triangulation of {\nextone} topology). 
When constructing the region masks, we particularly \emph{exclude} the ear, eyeball, nostril regions because these regions might not be reconstructed in some reconstruction methods or not considered in some 3DMMs. 
We also include some \emph{overlapping} regions between two adjacent masks to avoid boundary instability during evaluation.
Please see Fig.~\ref{fig:append:template_shape} for an illustration.

\myparagraph{Keypoints}
We prepare three sets of keypoints on {\tempshape} for different use cases:
(1) \textbf{Keypoints for alignment and wrapping}. We ask experts to manually label $118$ keypoints on the facial region of the template shape, including $24$ keypoints on the eyebrow, $48$ on the eyelids, $10$ on the nose and nose bridge, $36$ on the mouth. 
This set of keypoints is used to align and retopologize the input scans (elaborated in Sec.~\ref{append:alignment} and Sec.~\ref{append:wrapping} respectively).
(2) \textbf{Keypoints for evaluation}. 
Existing methods~\cite{bfm09,flametopo,3DDFA_v2,deep3d,3DDFA,MoFA} usually include $68$ semantically meaningful keypoints for evaluation or defining landmark loss for training. To setup a comparable setting, we also prepare $68$ keypoints with the same semantic information as previous work, including $10$ keypoints on the eyebrows, $12$ on the eyelids, $9$ on the nose and nose bridge, $20$ on the mouth, and $17$ on the cheek contour. 
This set of keypoints will be transferred to the ground-truth scans and the retopologized meshes for evaluation (e.g., used in {\gicp}, {\ricp} and {\bicp} in our evaluation pipeline as introduced in Sec. 6 in our main paper). We particularly denote this set of keypoints as {\tempkpt}. 
(3) \textbf{Keypoints for 3DMM fitting}. We follow~\cite{hifi3dface2021tencentailab} to prepare $86$ keypoints including, $18$ on the eyebrows, $16$ on the eyelids, $15$ on the nose and nose bridge, $20$ on the mouth, and $17$ on the cheek contour. 
This set of keypoints will be transferred to our newly introduced basis {\nextpp} for 3DMM fitting.

\subsection{Aligning Scans to {\tempshape}}\label{append:alignment}
To construct our benchmark {\name} and 3DMM basis {\nextpp}, we need to collect and register large set of scans from different datasets~\cite{facescape,LYHM,hifi3dface2021tencentailab}.
However these scans are in random pose and scales. 
For example, the surface area of scans in LYHM~\cite{LYHM} dataset ranges from $82,913$ mm$^2$ to $340,916$ mm$^2$, while the scans in FaceScape~\cite{facescape} may have opposite orientations.

\begin{figure}[!t]
    \centering
    \begin{overpic}[trim=3cm 45cm 86cm 0cm,clip,width=0.6\linewidth,grid=false]{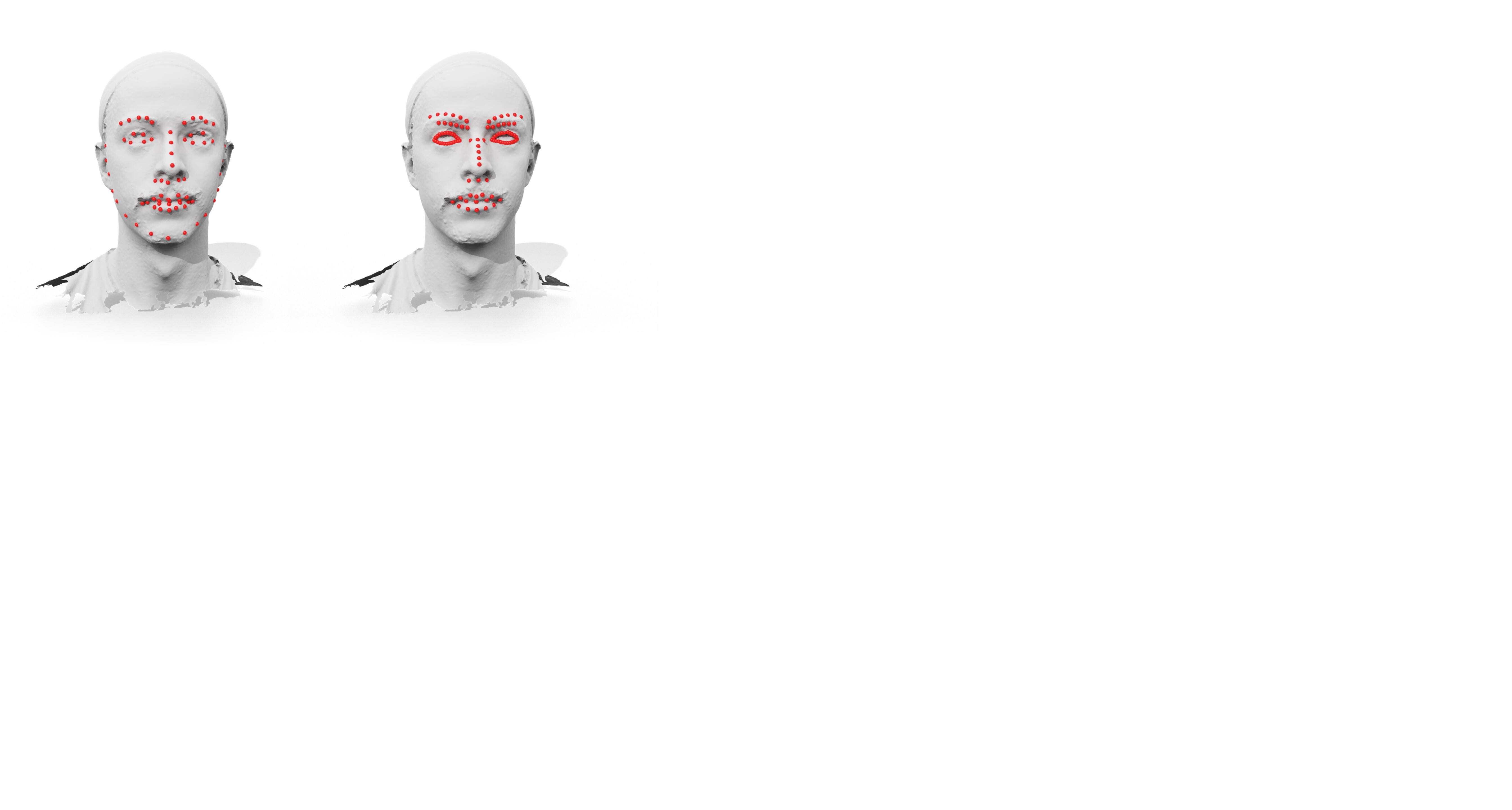}
    \end{overpic}
    \caption{\emph{Left}: inaccurate keypoints (e.g., in the nose) provided in LYHM dataset, which are extracted using the mixture-of-trees algorithm~\cite{lmk_detect_2012}. \emph{Right}: our high-quality keypoints obtained from a state-of-the-art landmark detector for global alignment \& registration.}
    \label{fig:append:scan_lmk}
\end{figure}

Therefore, our first step is to rescale and align the input scans to the template shape {\tempshape}.
Specifically, for a given scan $S_H$, we first rescale and align it to {\tempshape} using the provided keypoints from the source dataset. However, those provided keypoints are not accurate enough, as shown on the left of Fig.~\ref{fig:append:scan_lmk}, which leads to unsatisfactory alignment. 
To tackle this problem, we iterate through the following steps until convergence: 
(1) render a frontal face image of $\Scale[0.9]{S_H}$ with texture ($3$k+ resolution) using the initial/estimated transformation to align $\Scale[0.9]{S_H}$ to $\Scale[0.9]{S_{\text{temp}}}$ (note that the frontal pose needs to be determined from the alignment transformation as the frontal facing pose is unknown for a given scan); 
(2) detect $256$ 2D facial keypoints on the rendered image of $\Scale[0.9]{S_H}$ using a state-of-the-art landmark detector\footnote{we only keep $118$ keypoints in the facial region, which are in correspondences with the 118 keypoints defined on {\tempshape}, i.e., the first set of keypoints we discussed in Sec.~\ref{append:keypointdefine}.}; 
(3) project the 2D keypoints into 3D using the rendering camera pose; 
(4) update the alignment transformation from $\Scale[0.9]{S_H}$ to $\Scale[0.9]{S_{\text{temp}}}$ using the correspondences between the projected 3D keypoints on $\Scale[0.9]{S_H}$ and the known 3D keypoints on $\Scale[0.9]{S_{\text{temp}}}$. 
Note that we solve for a scale factor, rotation matrix, and translation vector for the shape transformation.

\subsection{Synthesizing Multi-view Images}
\begin{figure}[!t]
    \centering
    \begin{overpic}[trim=0cm 5cm 22cm 0cm,clip,width=1\linewidth,grid=false]{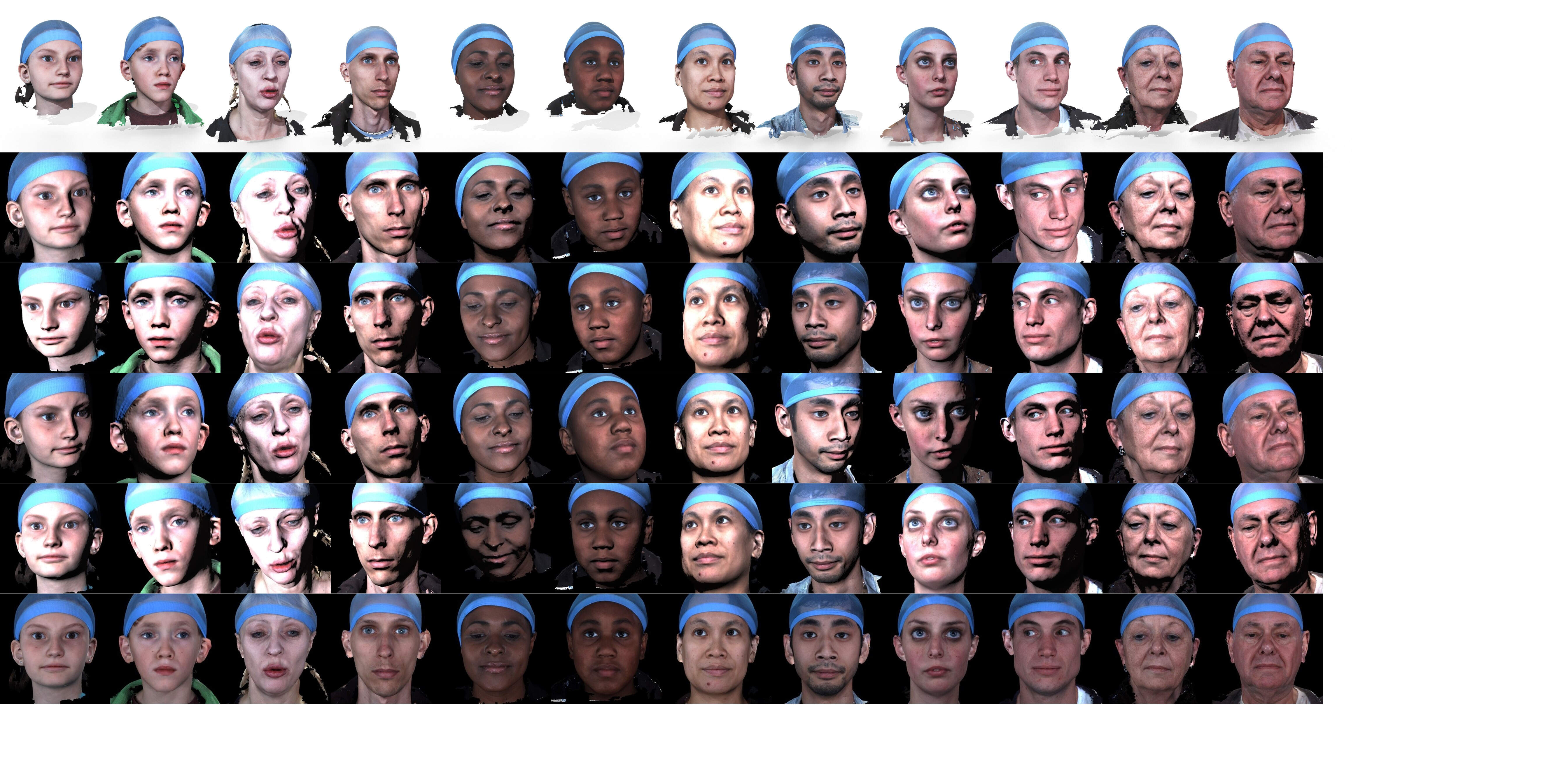}
    \end{overpic}
    \caption{\textbf{Examples of our synthesized images in {\name} benchmark.} 
    \emph{First row}: aligned high-resolution scans with textures. 
    \emph{Second-forth rows}: multi-view images of each scan. 
    \emph{Fifth row}: frontal images of each scan.}
    \label{fig:append:multiview}
\end{figure}

Since all the high-resolution scans have been aligned to the template shape {\tempshape}, which has known position and orientation, we can now synthesize multi-view images for each scan in a controlled setting. 
Specifically, We render the input scan with its corresponding texture on black background through a \emph{perspective camera}.
We fix the intrinsic parameters ($\Scale[0.9]{f_x}=2500,\Scale[0.9]{f_y}=2500,\Scale[0.9]{c_x}=512,\Scale[0.9]{c_y}=512$) of the camera and change the extrinsic parameters and lighting conditions to get a set of multi-view images in $1024\times 1024$ resolution, including a frontal image and $4$ images in \emph{random} poses (with angles less than $20$ degree). 
We also generate the ground-truth depth map for each image and record the ground-truth camera parameters. 
Our generated multi-view RGB-D image collection makes our benchmark suitable for evaluating face reconstruction methods under various input settings (i.e., single/multi-view RGB(-D) images).
Fig.~\ref{fig:append:multiview} shows some examples of the generated images in our {\name} benchmark.

\subsection{Retopologizing the Aligned Scans}\label{append:wrapping}
For each scan $\Scale[0.9]{S_H}$, we wrap the template shape $\Scale[0.9]{S_{\text{temp}}}$ to obtain $\Scale[0.9]{S_L}$, a ground-truth mesh in relatively low resolution with consistent topology across different individuals. Recall that $\Scale[0.9]{S_{\text{temp}}}$ adopts the {\nextone} topology that contains $20481$ vertices and $40832$ triangles together with 3 sets of predefined keypoints and facial region masks.
We follow~\cite{FaceWarehouse} to retopologize the input scan in neutral expression via a two-step approach.
(1) The facial region of $\Scale[0.9]{S_{\text{temp}}}$ is deformed to fit the facial region of $\Scale[0.9]{S_H}$ using non-rigid ICP technique~\cite{NICP}. The total energy on mesh deformation includes a smoothness term and a landmark loss term, where the predefined keypoints on $\Scale[0.9]{S_{\text{temp}}}$ (the first set of keypoints introduced in Sec.~\ref{append:keypointdefine}) are forced to be as close as possible to the automatically detected keypoints on $\Scale[0.9]{S_H}$ (introduced in Sec.~\ref{append:alignment}).
(2) We postprocess the deformed $\Scale[0.9]{S_{\text{temp}}}$ to remove the spikes, which come from fitting to the noisy regions in $\Scale[0.9]{S_H}$. We use Laplacian-based editing operations to fix this issue and obtain our high-quality mesh $\Scale[0.9]{S_L}$. See Fig.~2 in the main paper for some examples of the registered scans.

\subsection{Transferring Keypoints and Region Masks }
With the help of the retopologized meshes $\Scale[0.9]{S_L}$, we can now transfer the keypoints and region masks defined on $\Scale[0.9]{S_{\text{temp}}}$ to the high-resolution scans $\Scale[0.9]{S_H}$. First of all, we can easily transfer the keypoints/region masks from $\Scale[0.9]{S_{\text{temp}}}$ to $\Scale[0.9]{S_L}$ (by vertex index) since they share the same mesh topology. Then the keypoints/region masks are transferred from $\Scale[0.9]{S_L}$ to $\Scale[0.9]{S_H}$ as follows:
(1) Firstly, we traverse each point in a region $\Scale[0.9]{\mathcal{R}_L}$ on $\Scale[0.9]{S_L}$, and find its closest plane in $\Scale[0.9]{S_H}$. We then collect these mapped triangles and their one-ring neighbors as the candidate corresponding region $\Scale[0.9]{\mathcal{R}_H}$ on $\Scale[0.9]{S_H}$.
However, due to the significant difference in resolution between $\Scale[0.9]{S_L}$ and $\Scale[0.9]{S_H}$, the candidate region $\Scale[0.9]{\mathcal{R}_H}$ only contains limited and isolated vertices and triangles on $\Scale[0.9]{S_H}$. 
(2) Secondly, we improve $\Scale[0.9]{\mathcal{R}_H}$ by searching from the other direction, i.e., from $\Scale[0.9]{S_H}$ to $\Scale[0.9]{\mathcal{R}_L}$. We find the vertices in $\Scale[0.9]{S_H}$ such that their nearest neighbor in $\Scale[0.9]{S_L}$ (in vertex-to-plane distance) lie in the region $\Scale[0.9]{\mathcal{R}_L}$. We then include these vertices into $\Scale[0.9]{\mathcal{R}_H}$. 
In this way, we get a more complete region $\Scale[0.9]{\mathcal{R}_H}$ on $\Scale[0.9]{S_H}$.
Note that this step can be greatly accelerated by only considering a bounding box calculated based on step one's results instead of considering all the vertices in $\Scale[0.9]{S_H}$ for searching.
(3) Thirdly, we filter out the vertices lie in eyeball, nostril, or mouth cavity from $\Scale[0.9]{\mathcal{R}_H}$ since these regions might be wrongly included into $\Scale[0.9]{\mathcal{R}_H}$ due to nearest neighbor searching. 
To achieve this, we construct pseudo faces in these cavity regions on the template shape and find the vertices in $\Scale[0.9]{S_H}$ that have nearest neighbor lying in these pseudo faces. 
These vertices will be excluded from $\Scale[0.9]{\mathcal{R}_H}$.
(4) We then crop a region centered at the nose tip for each scan for evaluation. Specifically, the region has a radius of $0.7 \times (d_\text{outer\_eye} + d_\text{nose})$, where $d_\text{outer\_eye}$ is the outer-eye-distance and $d_\text{nose}$ is the distance between nose bridge and nose lower cartilage, respectively.
(5) Finally, we find the maximum connected region of $\Scale[0.9]{\mathcal{R}_H}$ and take it as the final region mask on $\Scale[0.9]{S_H}$.

\begin{figure*}[!t]
    \centering
    \begin{overpic}[trim=0cm 59cm 0cm 0cm,clip,width=1\linewidth,grid=false]{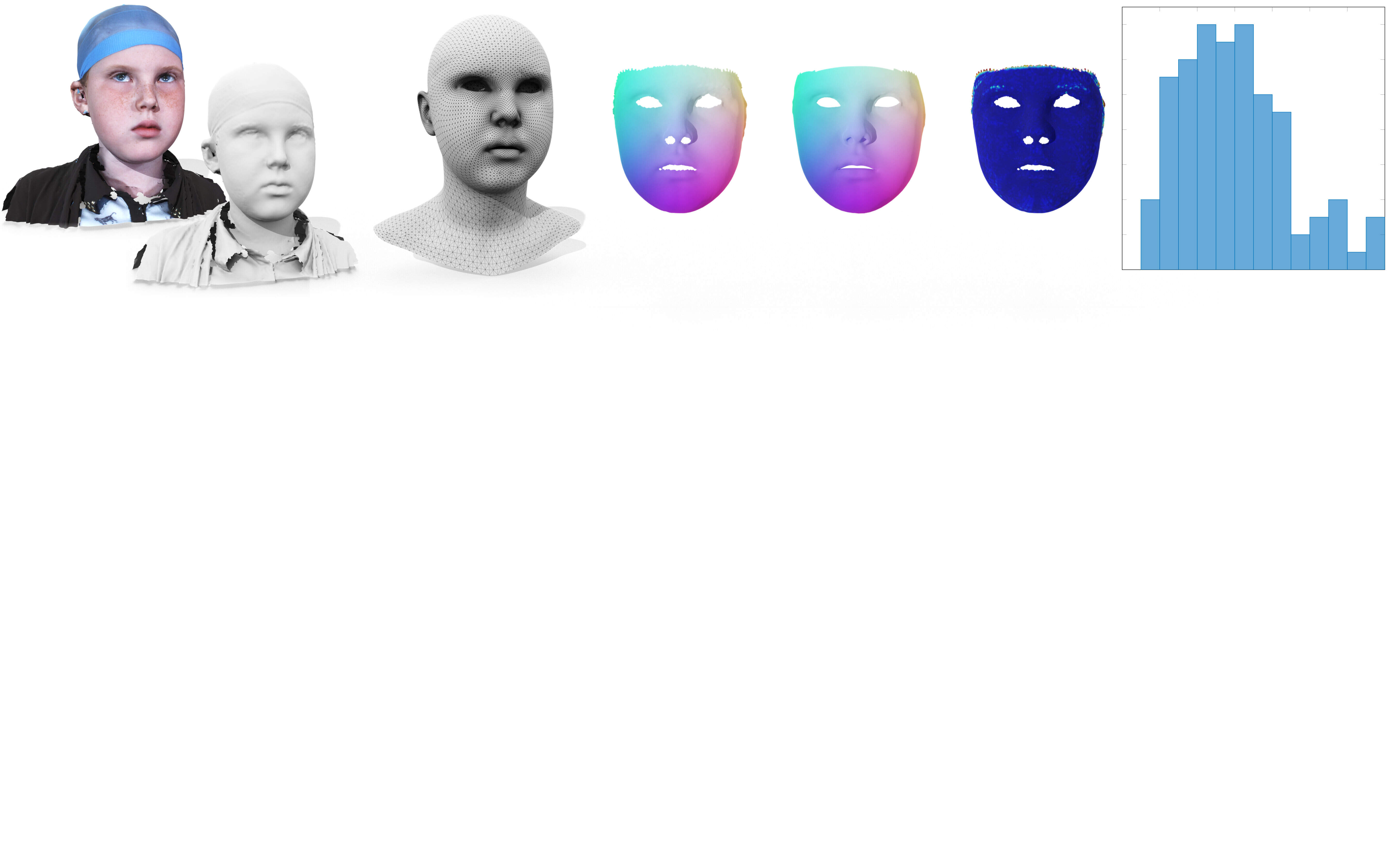}
    \put(4,22){\tiny Aligned Scan $S_H$}
    \put(26,22){\tiny Retopologized Mesh $S_L$}
    \put(49,22){\tiny Map/Error between $S_H$ and $S_L$}
    \put(82,22){\tiny Error of {\name}}
    
    \put(44,18){\tiny Corres. on $S_H$}
    \put(57,18){\tiny Corres. on $S_L$}
    \put(70,18){\tiny Error on $S_H$}
    \put(47,4){\scriptsize *vertices in correspondences}
    \put(48,2){\scriptsize are assigned the same color}
    
    \put(100,15){\scriptsize \rotatebox{270}{\# Shapes}}
    \put(83,0){\scriptsize Average Error}
    \put(80,0.5){\tiny 0}
    \put(98,0.5){\tiny 0.11mm}
    \put(79,20){\tiny 15}
    \end{overpic}
    \caption{Illustration of {\name} quality. \emph{Left}: samples of $S_H$, $S_L$, correspondence \& error maps on $S_H$ and $S_L$. \emph{Right}: error distribution of $100$ individuals in {\name}, $x$-axis represents the error range between $0\sim 0.11$mm, $y$-axis represents the shape numbers.}\label{fig:append:dt_err}
\end{figure*}

\section{Additional Experiment and Discussion}\label{append:res}

\subsection{Quality of {\name}}
To demonstrate the reliability of the retopologized meshes $\Scale[0.9]{S_L}$, we evaluate the similarity between $\Scale[0.9]{S_L}$ and the corresponding scan $\Scale[0.9]{S_H}$ as illustrated in Fig.~\ref{fig:append:dt_err}. 
Note that $\Scale[0.9]{S_L}$ and $\Scale[0.9]{S_H}$ are aligned and we can compute a map $\Scale[0.9]{\mymap{h}{l}{pts}}$ from $\Scale[0.9]{S_H}$ to $\Scale[0.9]{S_L}$ via nearest neighboring searing in Euclidean space. 
In the middle of Fig.~\ref{fig:append:dt_err} we visualize the map via color transfer. 
We then evaluate the shape similarity by NMSE, i.e., $\Scale[0.9]{e\big(\mymap{h}{l}{pts}\big)}$. 
We also visualize the per-vertex error on $\Scale[0.9]{S_H}$, which shows extremely small errors in the facial region. 
We then evaluate the shape similarity between $\Scale[0.9]{S_H}$ and $\Scale[0.9]{S_L}$ on the $100$ individuals in {\name} and report the average errors in the histogram in Fig.~\ref{fig:append:dt_err}.
Specifically, the NMSE in the facial region ranges from $0.047\sim0.108$ mm. The average error over all vertices in the facial region across $100$ individuals is $0.070$ mm. 
This suggests that our retopologized mesh $\Scale[0.9]{S_L}$ are in high-quality and guarantees the similarity to the original scans $\Scale[0.9]{S_H}$.

\subsection{User Study Details}
We invited $70$ volunteers with computer science or modeling background to conduct the user study. 
In every question, the user is asked to select the most similar reconstructed mesh(es) compared to the given ground-truth scan.
Specifically, for each test sample, we design the following two questions:
\begin{enumerate}[label=(Q\arabic*),leftmargin=*]
    \item choose the most similar \emph{two} (compared to the ground-truth) from nine candidate meshes reconstructed using different methods (as shown in Fig.5 in the main paper).
    \item choose the most similar \emph{one} from up to three candidate meshes, which are the best reconstructions according to three different evaluation protocols (i.e., {\gicp} from two directions, and ours; highlighted in blue, purple, and orange boxes in the main paper).
\end{enumerate}
We report the results of (Q1) in Fig.~5 in the main paper, where the best (second best) is highlighted via ``$\star$'' (``$\dagger$''). We also show the statistics of the user study in Tab.~\ref{Tab.user_study}. For (Q2), on average, $76.1\%$ users agreed that our {\bicp} selected the best reconstructed mesh compared to {\gicp} evaluation protocols.

This user study shows that our {\bicp} indeed better aligns with human perception in measuring the similarity between the ground-truth and the reconstructed mesh. The additional results in Fig.~\ref{fig:apdx_main_result2} are also marked via ``$\star$'' and ``$\dagger$'' according to the above user study.

\begin{table}[t!]
\setlength{\tabcolsep}{8pt}
\caption{Detailed user study results of Fig.~5 in the main paper.}
\resizebox{\linewidth}{!}{%
\begin{tabular}{c|cc|c}
\toprule[1pt]
   Sample      & Best recon. $\star$          & 2nd best recon. $\dagger$   & Best selected by {\bicp}  \\ \bottomrule[0.5pt]
Fig.~5 (1) & 72.2\% (MGCNet) & 53.1\% (GANFit)  & 85.4\% (GANFit) \\
Fig.~5 (2) & 43.9\% (Deep3D) & 39.0\% (MGCNet)  & 70.7\% (Deep3D) \\
Fig.~5 (3) & 51.3\% (Deep3D) & 35.9\% (MGCNet)  & 76.9\% (Deep3D) \\ 
\bottomrule[1pt]
\end{tabular}%
}
\label{Tab.user_study}
\end{table}

\subsection{Comparing Different Reconstruction Methods}

\begin{figure*}[!h]
\centering
\begin{overpic}[trim=8cm 88cm 5cm 4cm,clip,width=1\linewidth,grid=false]{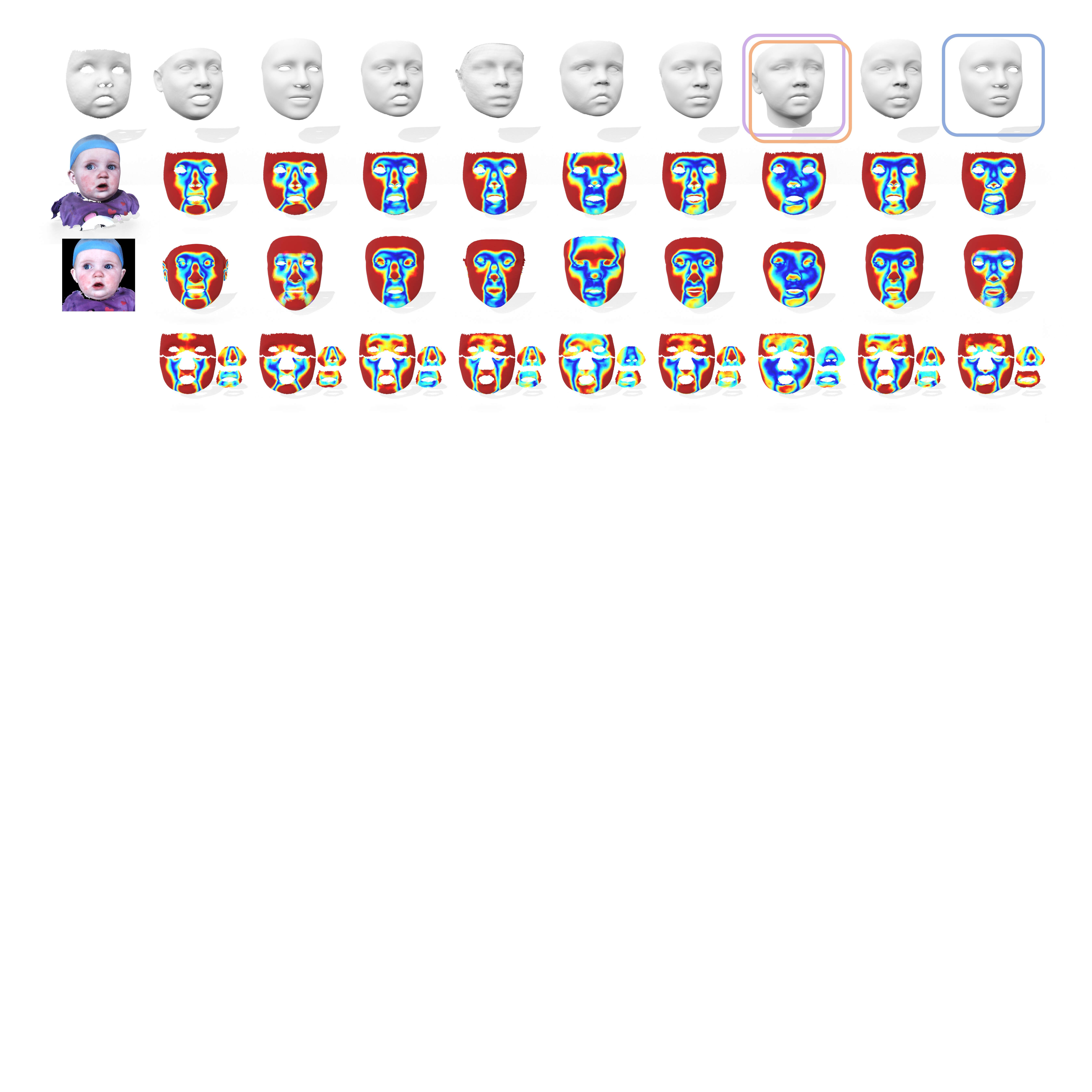}
    \put(2,39){\tiny\bfseries G.T.}
    \put(1.5,9){\tiny\bfseries Input}
    \put(10,39){\tiny\bfseries ExpNet}
    \put(20,39){\tiny\bfseries RingNet}
    \put(30,39){\tiny\bfseries MGCNet}
    \put(40,39){\tiny\bfseries PRNet}
    \put(50,39){\tiny\bfseries Deep3D}
    \put(59,39){\tiny\bfseries 3DDFA-v2}
    \put(70,39){\tiny\bfseries GANFit}
    \put(79,39){\tiny\bfseries N-3DMM}
    \put(91,39){\tiny\bfseries DECA}
    
    \put(36,37){\scriptsize\bfseries $\star$}
    \put(77,37){\scriptsize\bfseries $\dagger$}
  
\end{overpic}
\begin{overpic}[trim=8cm 88cm 5cm 4cm,clip,width=1\linewidth,grid=false]{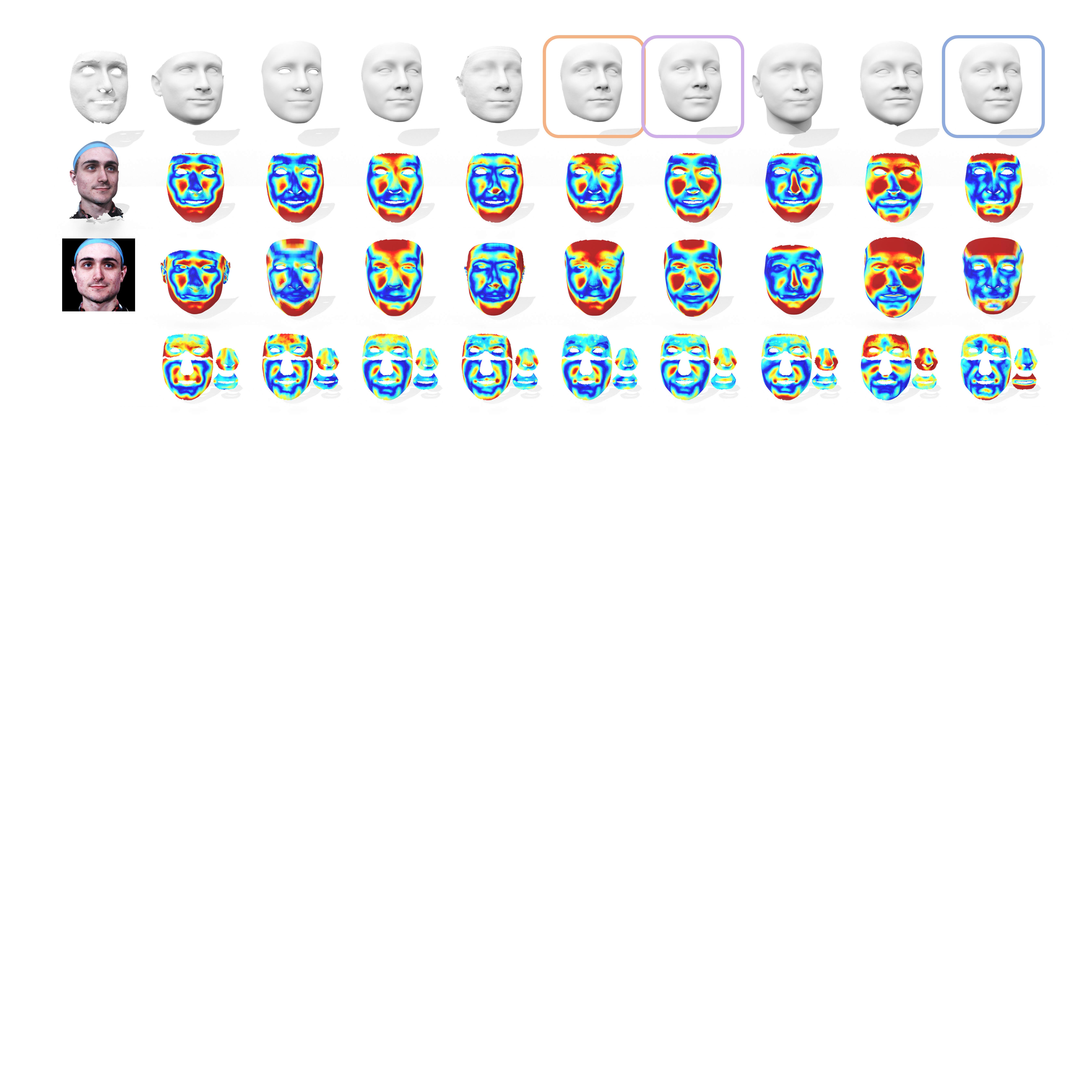}
    \put(2,39){\tiny\bfseries G.T.}
    \put(1.5,9){\tiny\bfseries Input}
    \put(56,37){\scriptsize\bfseries $\star$}
    \put(36,37){\scriptsize\bfseries $\dagger$}
\end{overpic}
\begin{overpic}[trim=8cm 88cm 5cm 4cm,clip,width=1\linewidth,grid=false]{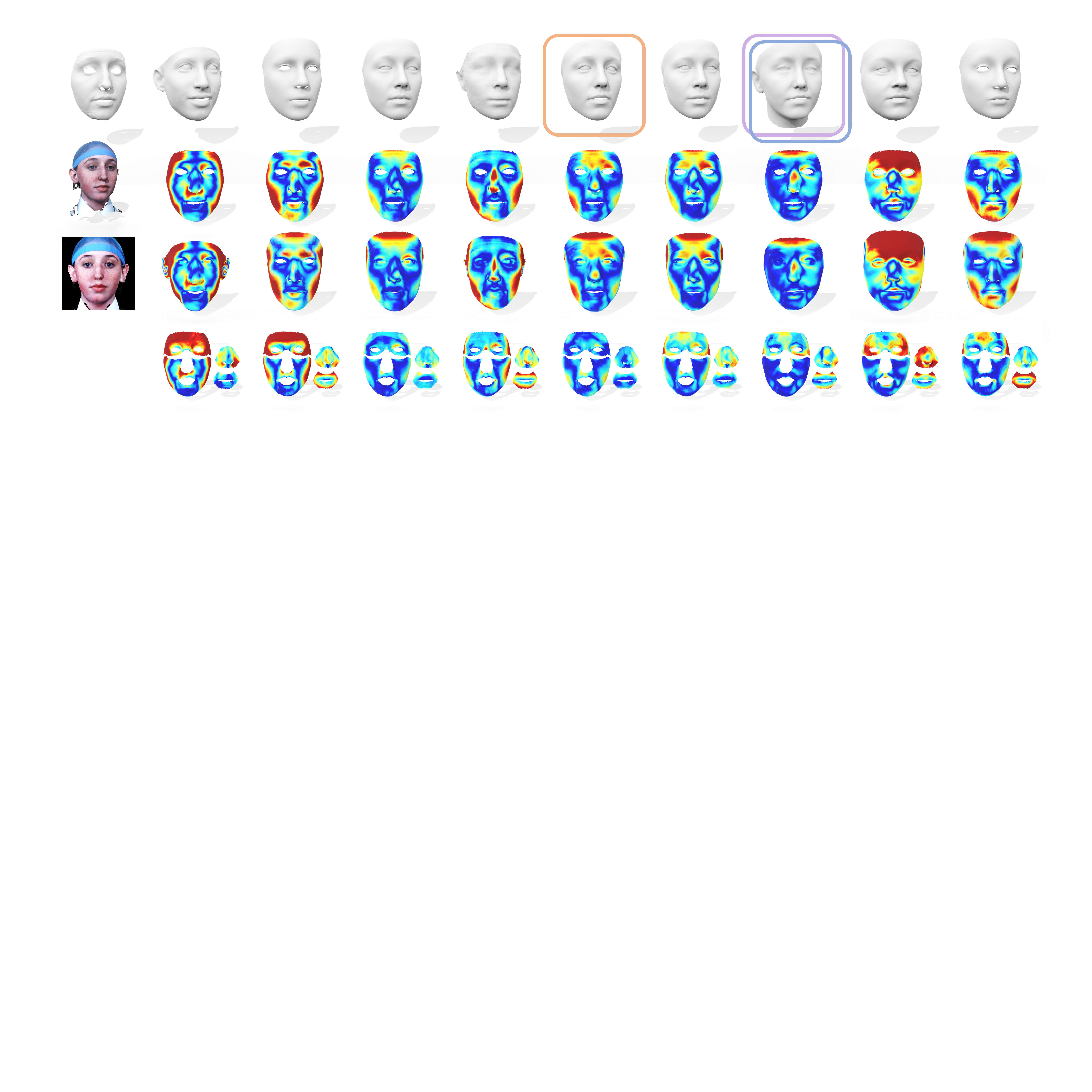}
    \put(2,39){\tiny\bfseries G.T.}
    \put(1.5,9){\tiny\bfseries Input}
    \put(56,37){\scriptsize\bfseries $\star$}
    \put(36,37){\scriptsize\bfseries $\dagger$}
\end{overpic}
\caption{\textbf{Comparing different face reconstruction methods (part 1).} We visualize the reconstruction error of each reconstructed face using the standard evaluation pipeline ({\gicp}) and our novel evaluation pipeline ({\bicp}, shown in four regions), where large (small) errors are colored in red (blue). The best reconstructed face selected using our measurement (in orange boxes) are visually closer to the ground-truth meshes than the ones selected using the standard measurements
(blue boxes for $\Scale[0.9]{e\big(T_{p\rightarrow h}^{\text{pts}}\big)}$ \& purple boxes for $\Scale[0.9]{e\big(T_{h\rightarrow p}^{\text{pts}}\big)}$). We also mark the best (second best)  reconstructed face voted in our user study by $\star$ ($\dagger$). 
The first row of each sample is the reconstructed shape, the second/third/fourth row of each sample is the error map of $\Scale[0.9]{e\big(T_{h\rightarrow p}^{\text{pts}}\big)}$)/$\Scale[0.9]{e\big(T_{p\rightarrow h}^{\text{pts}}\big)}$/ours.
These three samples are illustrated in main paper and we show bigger versions for easier comparisons.}
\label{fig:apdx_main_result1}
\end{figure*}

\begin{figure*}[!h]
\centering
    \begin{overpic}[trim=8cm 88cm 5cm 4cm,clip,width=1\linewidth,grid=false]{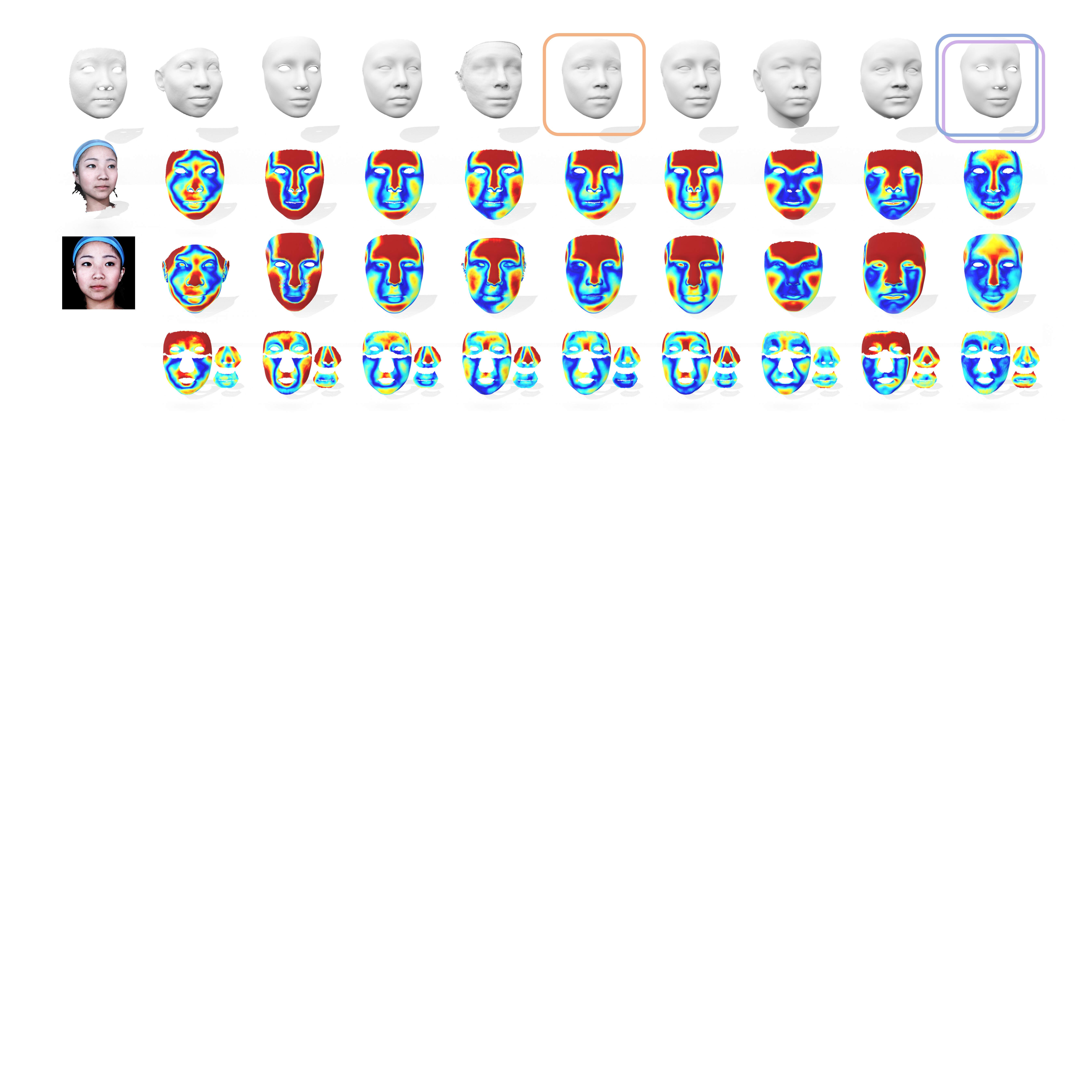}
        \put(2,39){\tiny\bfseries G.T.}
        \put(1.5,9){\tiny\bfseries Input}
        \put(10,39){\tiny\bfseries ExpNet}
        \put(20,39){\tiny\bfseries RingNet}
        \put(30,39){\tiny\bfseries MGCNet}
        \put(40,39){\tiny\bfseries PRNet}
        \put(50,39){\tiny\bfseries Deep3D}
        \put(59,39){\tiny\bfseries 3DDFA-v2}
        \put(70,39){\tiny\bfseries GANFit}
        \put(79,39){\tiny\bfseries N-3DMM}
        \put(91,39){\tiny\bfseries DECA}
        \put(36,37){\scriptsize\bfseries $\star$}
        \put(86,37){\scriptsize\bfseries $\dagger$}        

    \end{overpic}
    \begin{overpic}[trim=8cm 88cm 5cm 4cm,clip,width=1\linewidth,grid=false]{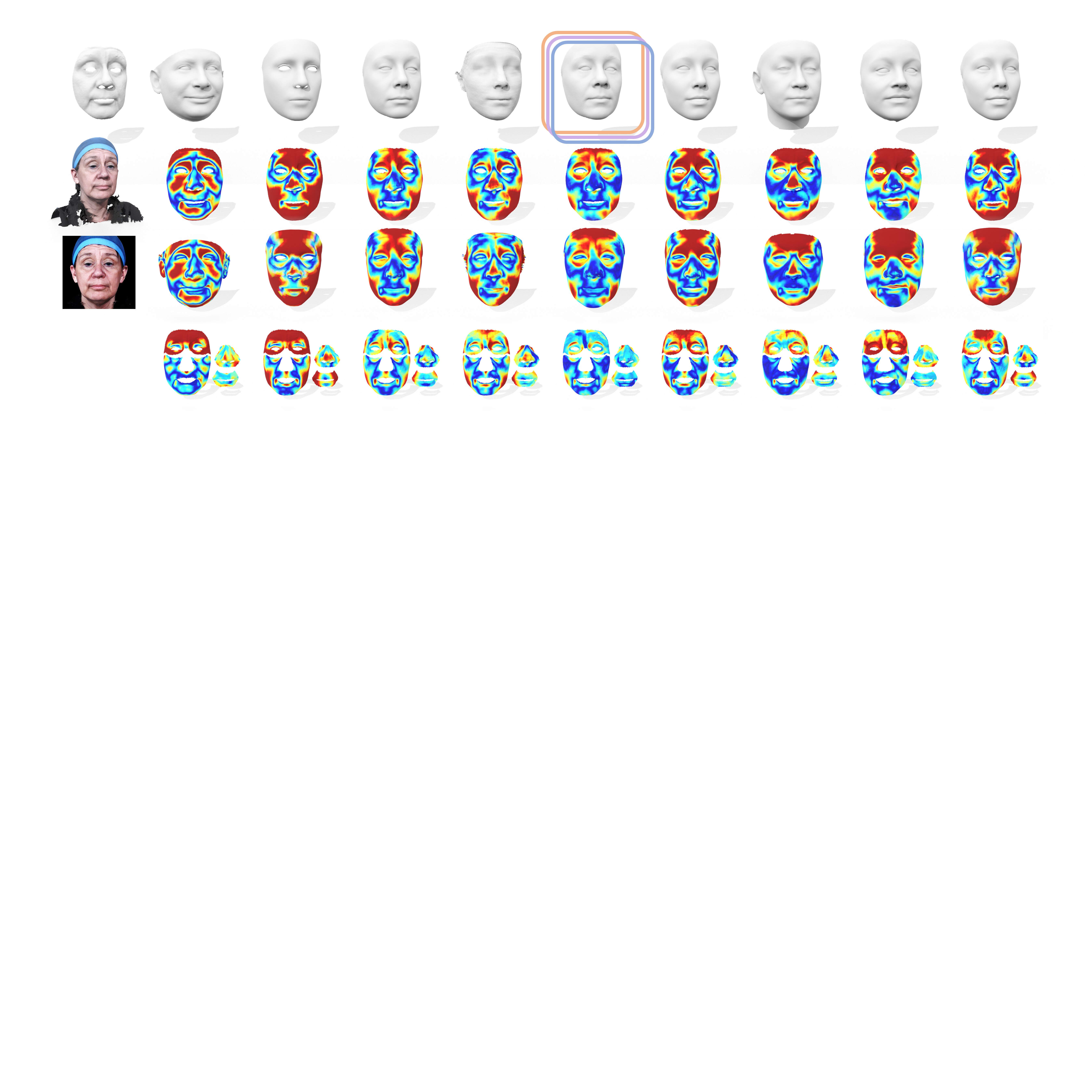}
    \put(2,39){\tiny\bfseries G.T.}
    \put(1.5,9){\tiny\bfseries Input}
    \put(56,37){\scriptsize\bfseries $\star$}
    \put(36,37){\scriptsize\bfseries $\dagger$}
    \end{overpic}
    \begin{overpic}[trim=8cm 88cm 5cm 4cm,clip,width=1\linewidth,grid=false]{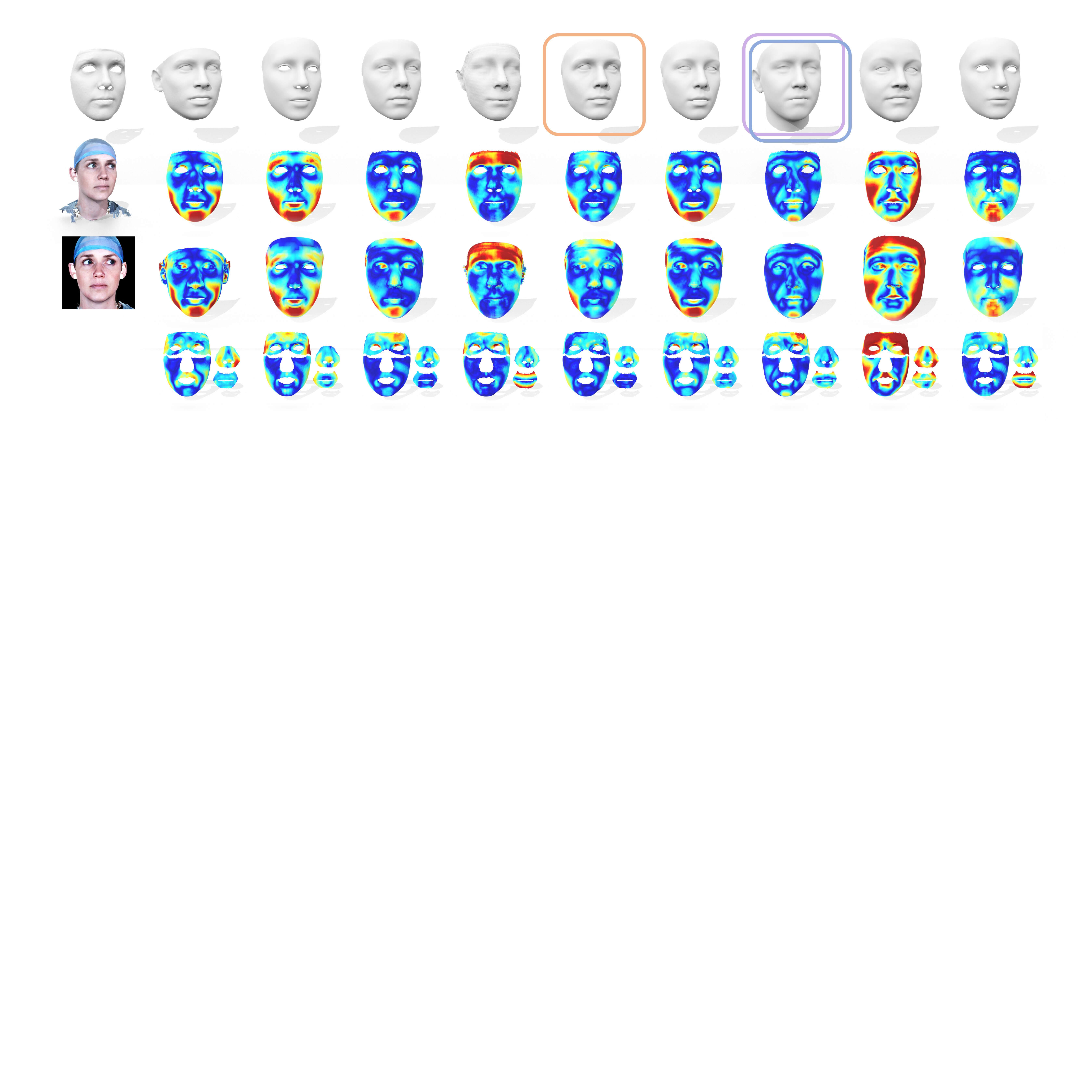}
        \put(2,39){\tiny\bfseries G.T.}
        \put(1.5,9){\tiny\bfseries Input}
        \put(56,37){\scriptsize\bfseries $\star$}
        \put(36,37){\scriptsize\bfseries $\dagger$}
    \end{overpic}
    \caption{\textbf{Comparing different face reconstruction methods (part 2).} We visualize the reconstruction error of each reconstructed face using the standard evaluation pipeline ({\gicp}) and our novel evaluation pipeline ({\bicp}, shown in four regions), where large (small) errors are colored in red (blue). The best reconstructed face selected using our measurement (in orange boxes) are visually closer to the ground-truth meshes than the ones selected using the standard measurement (blue boxes for $\Scale[0.9]{e\big(T_{p\rightarrow h}^{\text{pts}}\big)}$ \& purple boxes for $\Scale[0.9]{e\big(T_{h\rightarrow p}^{\text{pts}}\big)}$). We also mark the best (second best)  reconstructed face voted in our user study by $\star$ ($\dagger$). 
    The first row of each sample is the reconstructed shape, the second/third/fourth row of each sample is the error map of $\Scale[0.9]{e\big(T_{h\rightarrow p}^{\text{pts}}\big)}$)/$\Scale[0.9]{e\big(T_{p\rightarrow h}^{\text{pts}}\big)}$/ours.}
    \label{fig:apdx_main_result2}
\end{figure*}

We visualize the error map of different methods using the standard {\gicp} based evaluation pipeline (evaluated on both directions between the constructed face and the original scan) or our proposed {\bicp} based evaluation pipeline in Fig.~\ref{fig:apdx_main_result1} and Fig.~\ref{fig:apdx_main_result2}, where the errors are globally normalized across different methods, and blue (red) represents smaller (larger) error. 
Note that, the {\gicp} based errors of $\Scale[0.9]{e\big(\mymap{p}{h}{pts}\big)}$ are computed on the reconstructed faces $\Scale[0.9]{S_P}$ (see the \emph{third} row of each sample in Fig.~\ref{fig:apdx_main_result1} and Fig.~\ref{fig:apdx_main_result2}), while our {\bicp} based errors are computed on the four fine-grained regions on the high-resolution scans $\Scale[0.9]{S_H}$ (see the \emph{fourth} row of each sample in Fig.~\ref{fig:apdx_main_result1} and Fig.~\ref{fig:apdx_main_result2}).

In some cases, the global-wise error map may exhibit misleading results mainly due to inaccurate alignment between $\Scale[0.9]{S_P}$ and $\Scale[0.9]{S_H}$, which makes it hard to identify the best predicted face from different methods. 
As a comparison, our region-aware pipeline is more fair by making a comparison based on the errors defined on the same mesh $\Scale[0.9]{S_H}$ among different methods. 
Indeed, the best predicted face selected by {\bicp} is visually more similar to the input scan compared to the face selected by {\gicp} when there is a disagreement. 
At the same time, our {\bicp} can suggest which method performs the best in a particular region (see Fig.~5 and Tab.~4 in the main paper).

\clearpage

\begin{figure*}[!t]
\centering
    \begin{overpic}[trim=2cm 123cm 5cm 1cm,clip,width=1\linewidth,grid=false]{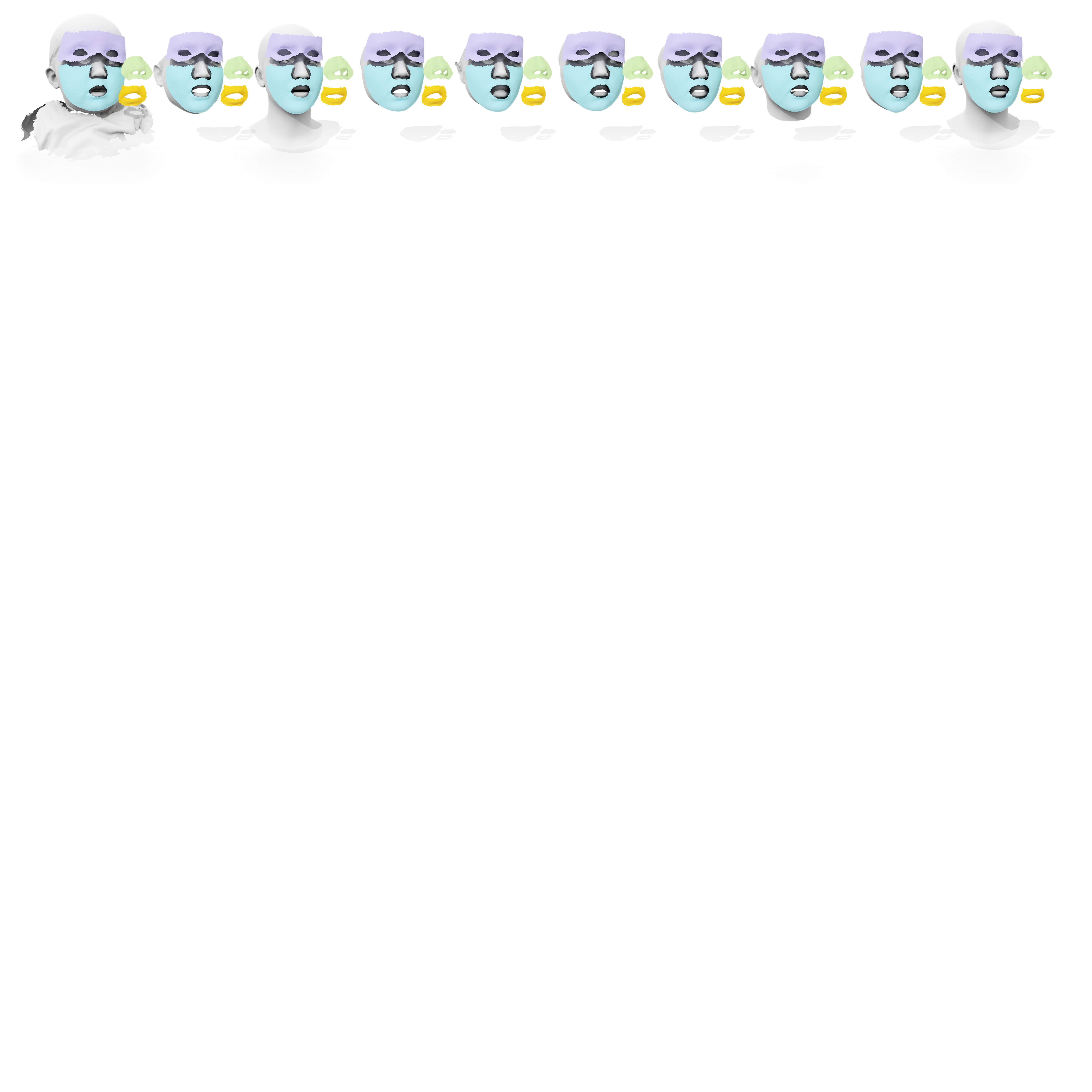}
        \put(5,13){\tiny\bfseries G.T.}
        \put(13,13){\tiny\bfseries ExpNet}
        \put(23,13){\tiny\bfseries RingNet}
        \put(33,13){\tiny\bfseries MGCNet}
        \put(43,13){\tiny\bfseries PRNet}
        \put(52,13){\tiny\bfseries Deep3D}
        \put(60,13){\tiny\bfseries 3DDFA-v2}
        \put(71,13){\tiny\bfseries GANFit}
        \put(81,13){\tiny\bfseries N-3DMM}
        \put(92,13){\tiny\bfseries DECA}
    \end{overpic}
    \begin{overpic}[trim=2cm 123cm 5cm 1cm,clip,width=1\linewidth,grid=false]{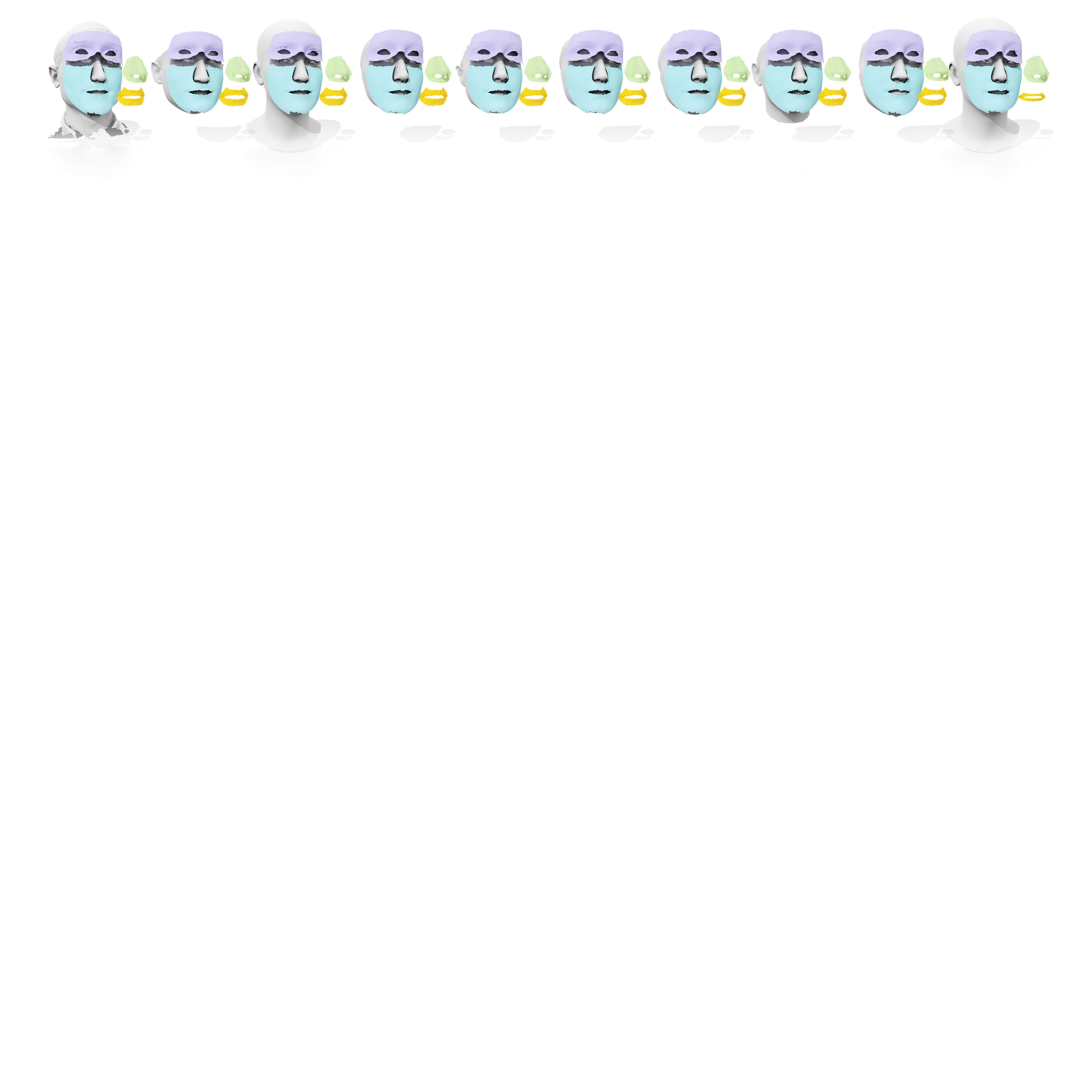}
    \end{overpic}
    \begin{overpic}[trim=2cm 123cm 5cm 1cm,clip,width=1\linewidth,grid=false]{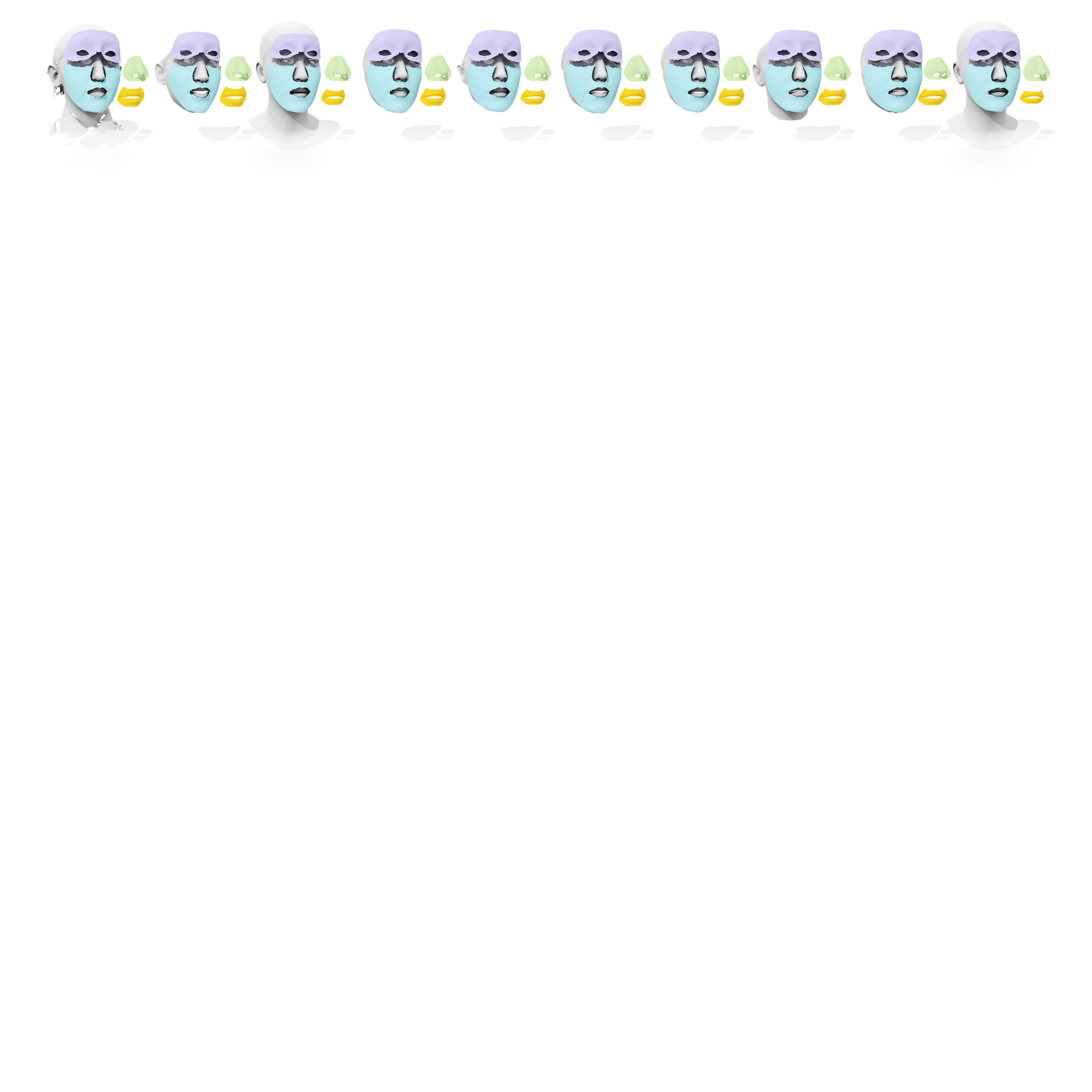}
    \end{overpic}
    \begin{overpic}[trim=2cm 123cm 5cm 1cm,clip,width=1\linewidth,grid=false]{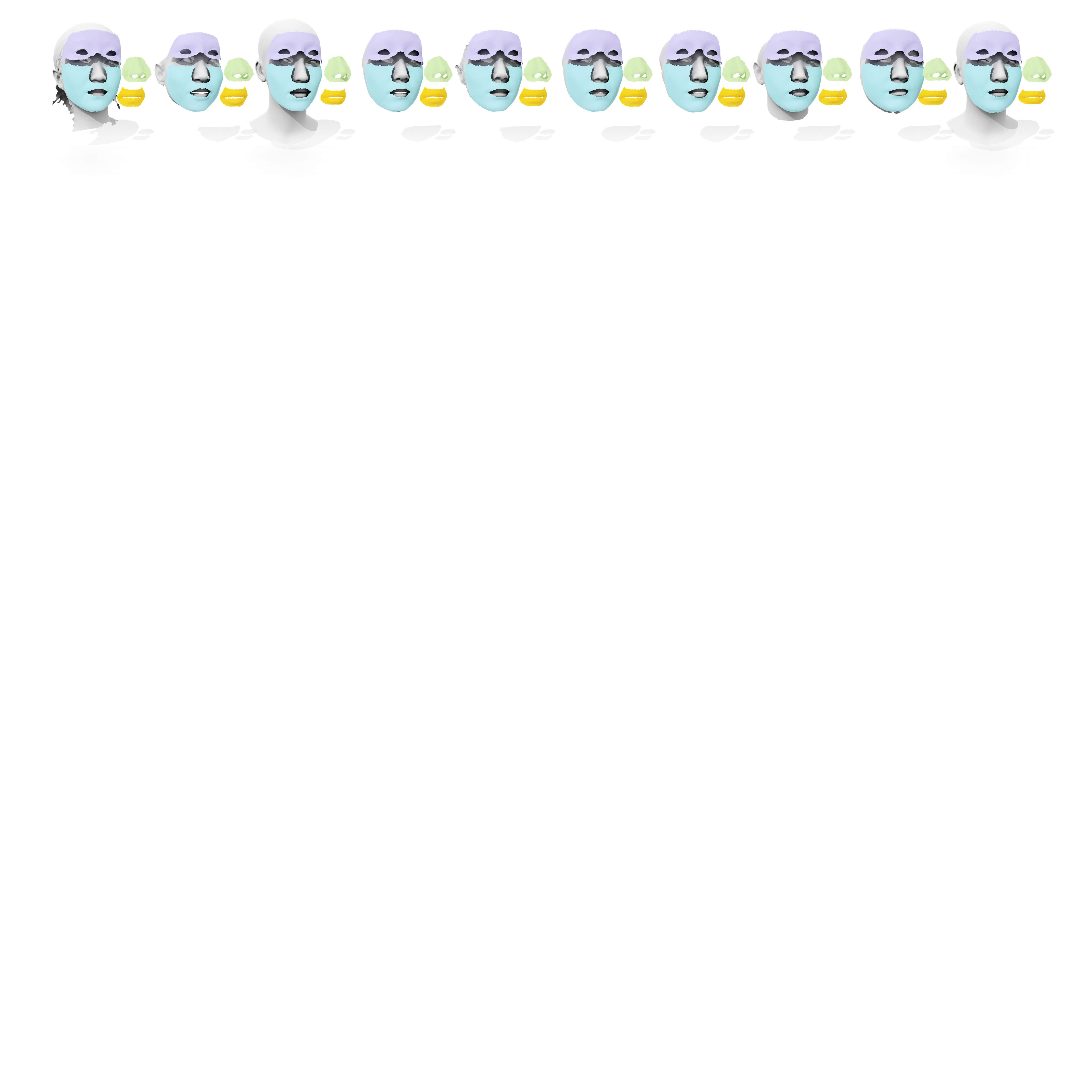}
    \end{overpic}
    \begin{overpic}[trim=2cm 123cm 5cm 1cm,clip,width=1\linewidth,grid=false]{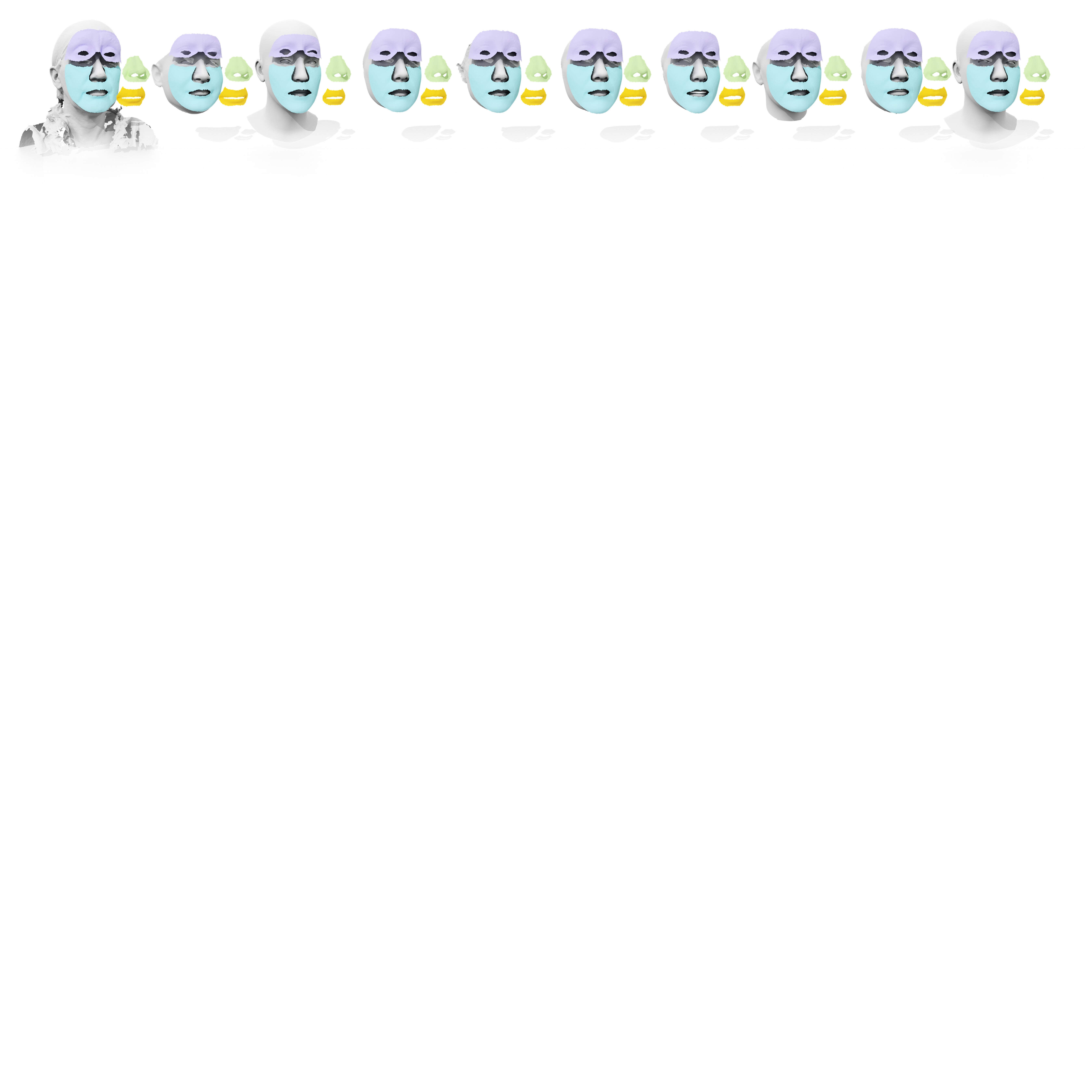}
    \end{overpic}
    \begin{overpic}[trim=2cm 123cm 5cm 1cm,clip,width=1\linewidth,grid=false]{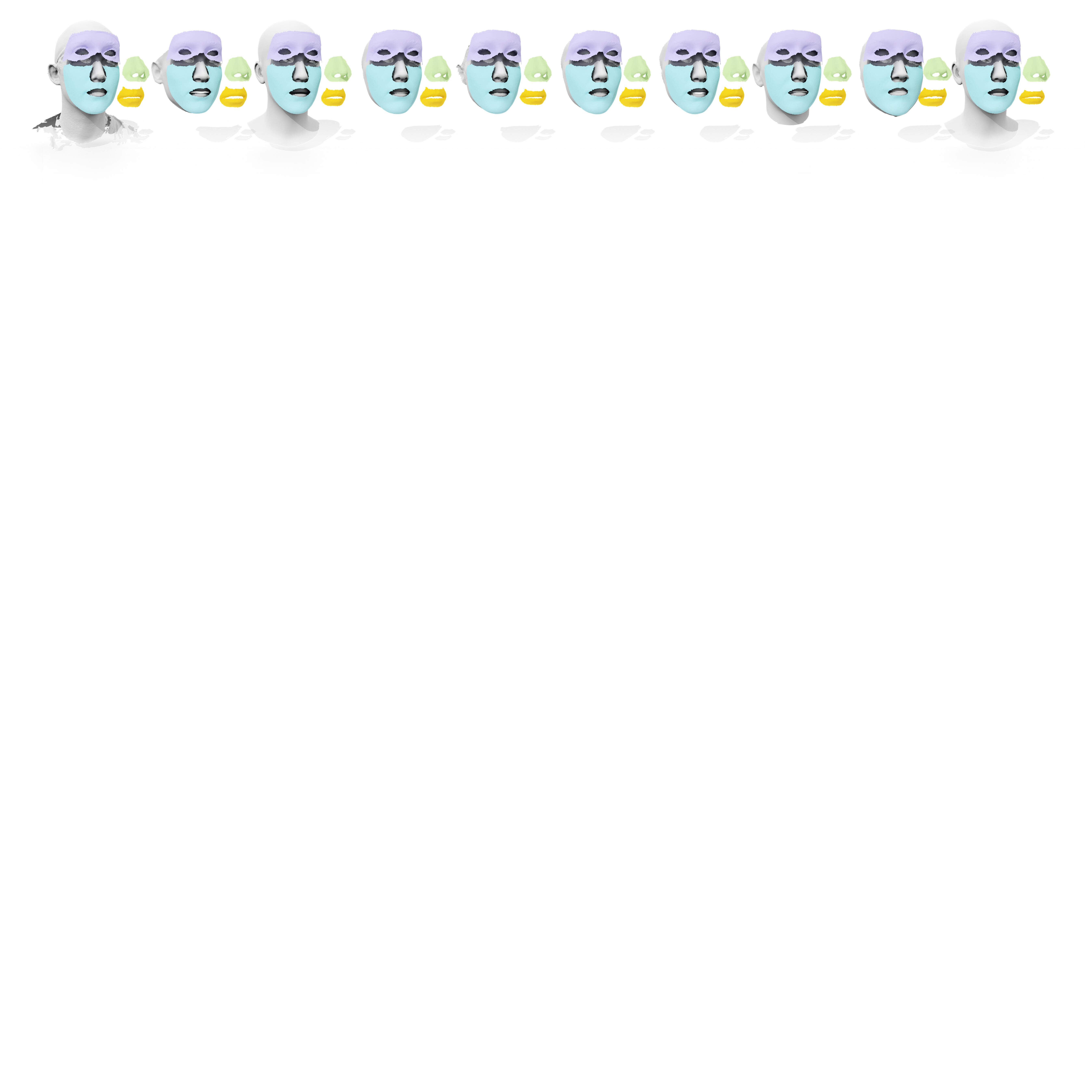}
    \end{overpic}
    \caption{\textbf{Examples of the deformed regions $\Scale[0.9]{\mathcal{R}_H^*}$ of each method.}
    We illustrate the deformed regions $\Scale[0.9]{\mathcal{R}_H^*}$, i.e., the intermediate results obtained after applying {\nicp} to deform the G.T. region $\Scale[0.9]{\mathcal{R}_H}$ (the \emph{first} column) to fit $\Scale[0.9]{S_P^*}$ in our evaluation pipeline.}
    \label{fig:apdx_NICP_result}
\end{figure*}

As explained in Sec.~6 in the main paper, 
our fine-grained region-wise alignment and the two-step coarse-to-fine registration
effectively helps {\nicp} to converge to a reasonably deformed shape $\Scale[0.9]{\mathcal{R}_H^*}$.
See Fig.~\ref{fig:apdx_NICP_result} for such examples, where we visualize the deformed regions on top of the reconstructed faces.

\subsection{Comparing Different 3DMMs} \label{append:3DMMcmpdiscuss}
\begin{figure*}[!t]
    \centering
    \begin{overpic}[trim=0cm 30cm 39cm 0cm,clip,width=0.3\linewidth,grid=false]{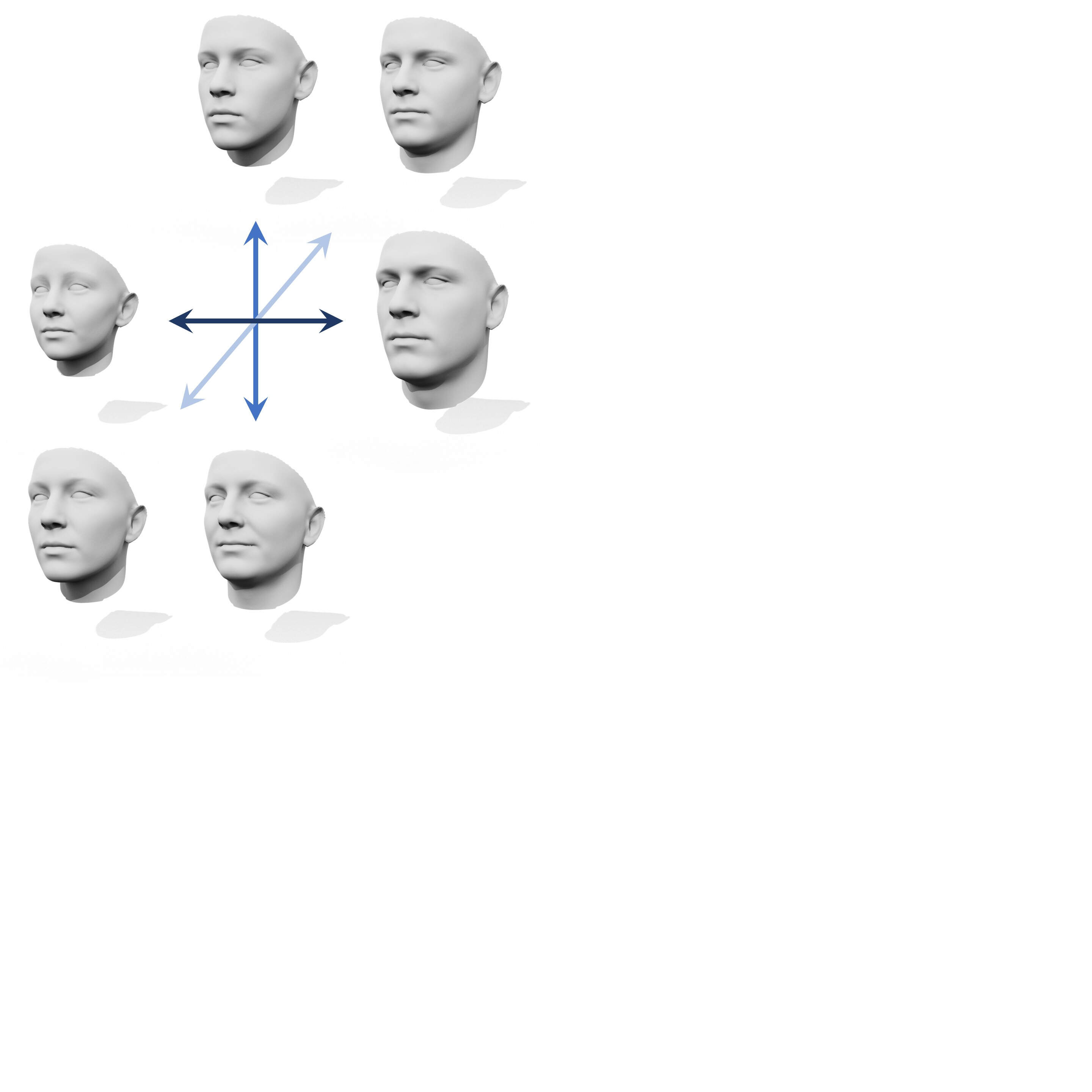}
    \put(0,80){\scriptsize\bfseries BFM}
    \end{overpic}\hspace{10pt}
    \begin{overpic}[trim=0cm 30cm 39cm 0cm,clip,width=0.3\linewidth,grid=false]{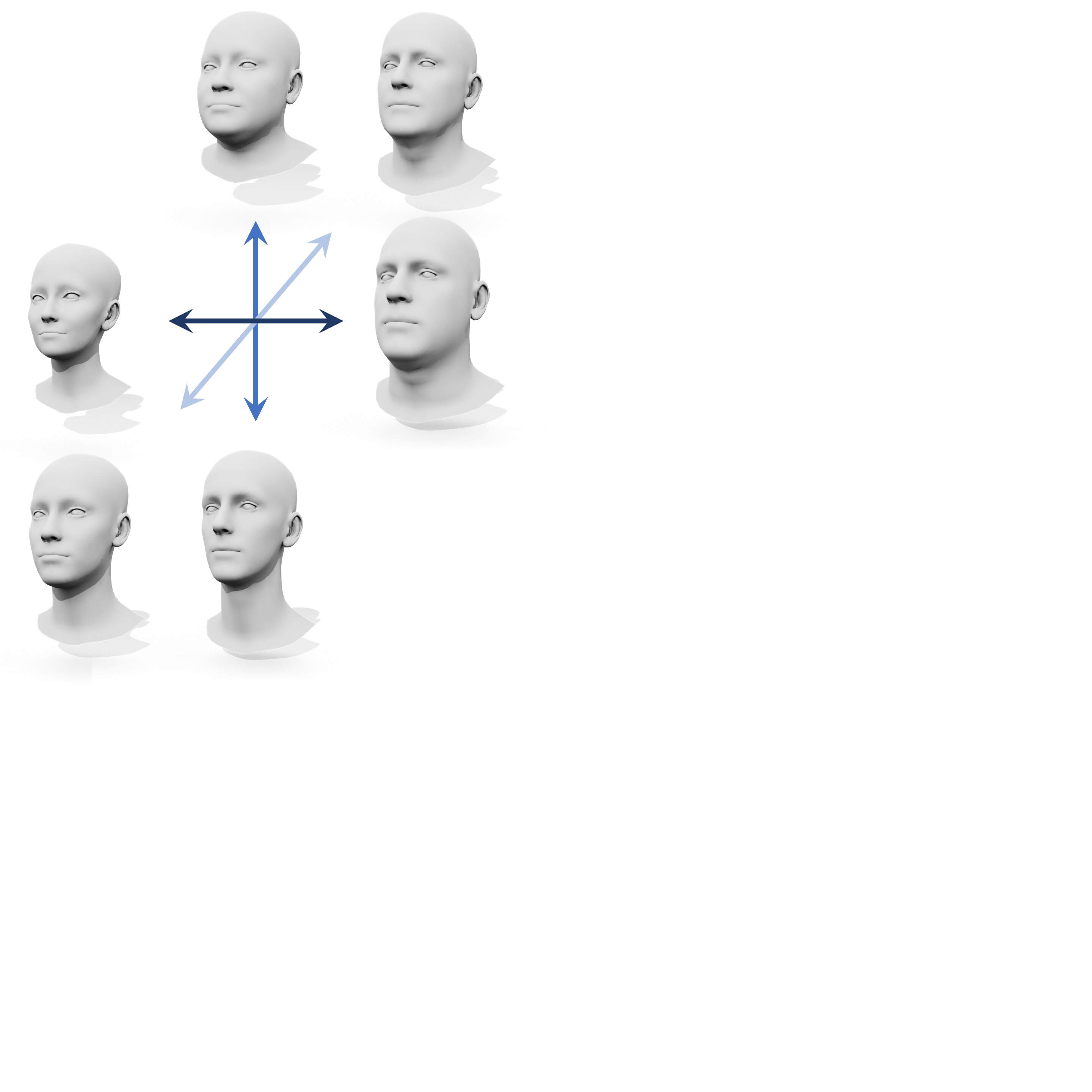}
    \put(0,80){\scriptsize\bfseries FLAME}
    \end{overpic}\hspace{10pt}
    \begin{overpic}[trim=0cm 30cm 39cm 0cm,clip,width=0.3\linewidth,grid=false]{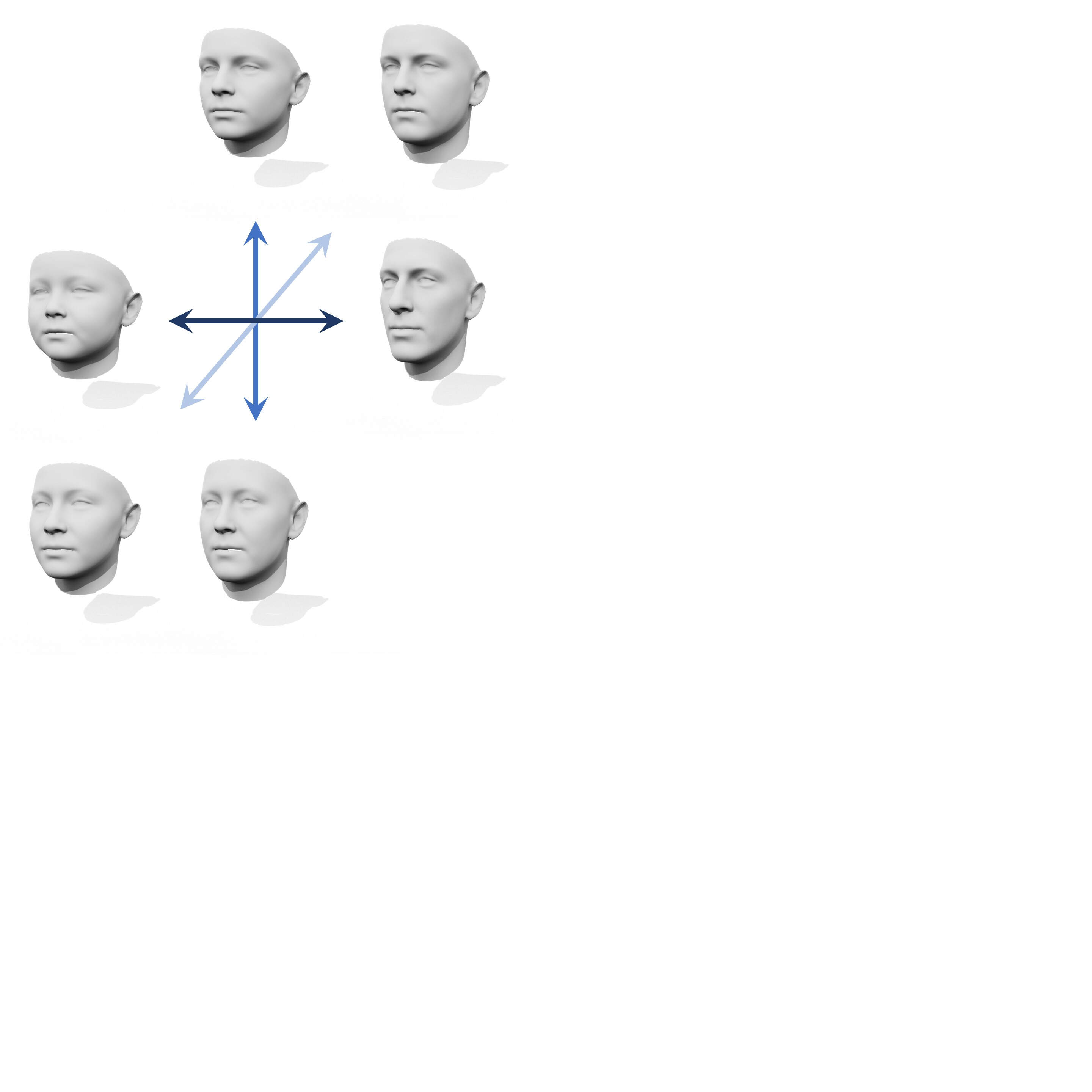}
    \put(0,80){\scriptsize\bfseries LSFM}
    \end{overpic}
    \begin{overpic}[trim=-1cm 30cm 39cm -1cm,clip,width=0.3\linewidth,grid=false]{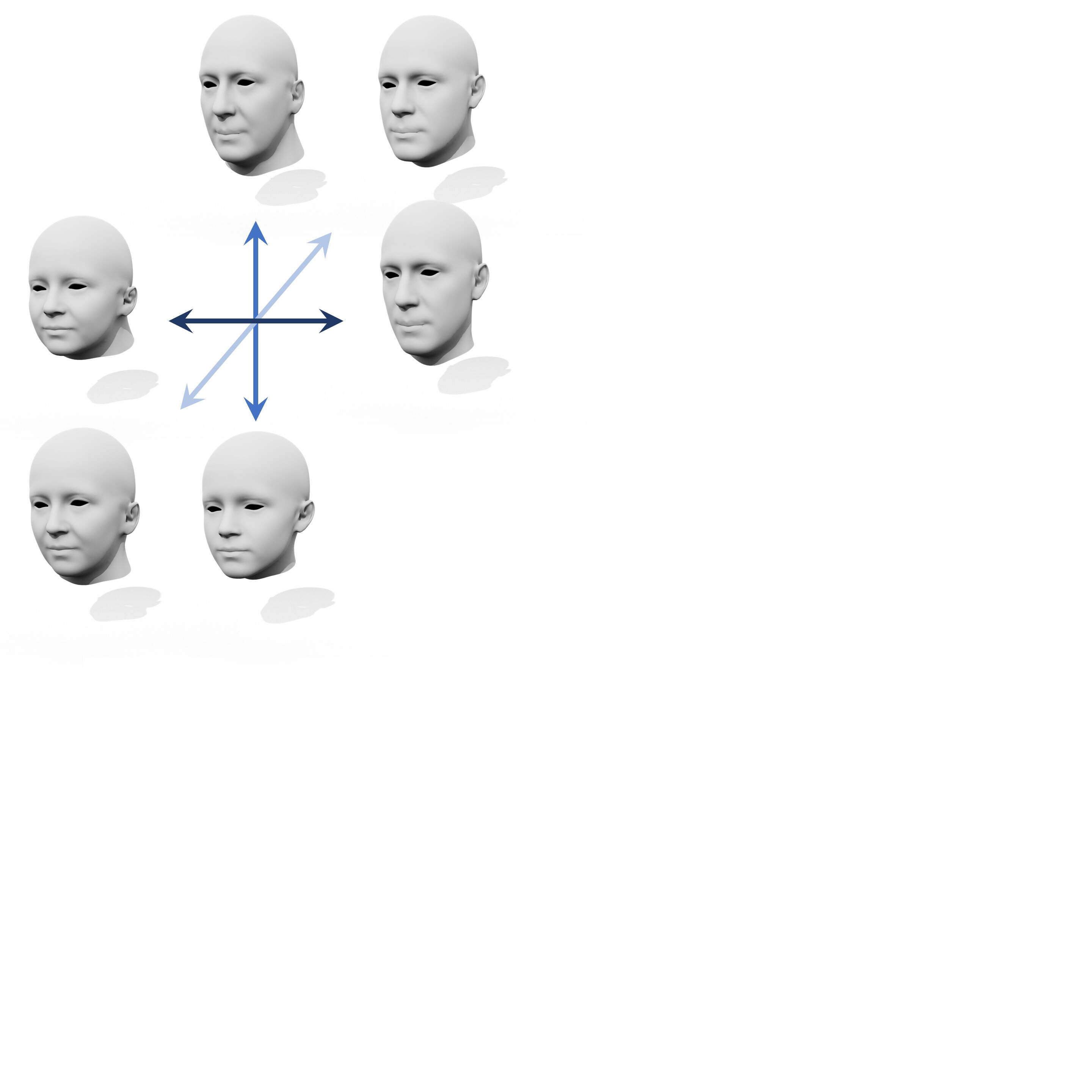}
    \put(0,80){\scriptsize\bfseries LYHM}
    \end{overpic}\hspace{10pt}
    \begin{overpic}[trim=-1cm 30cm 39cm -1cm,clip,width=0.3\linewidth,grid=false]{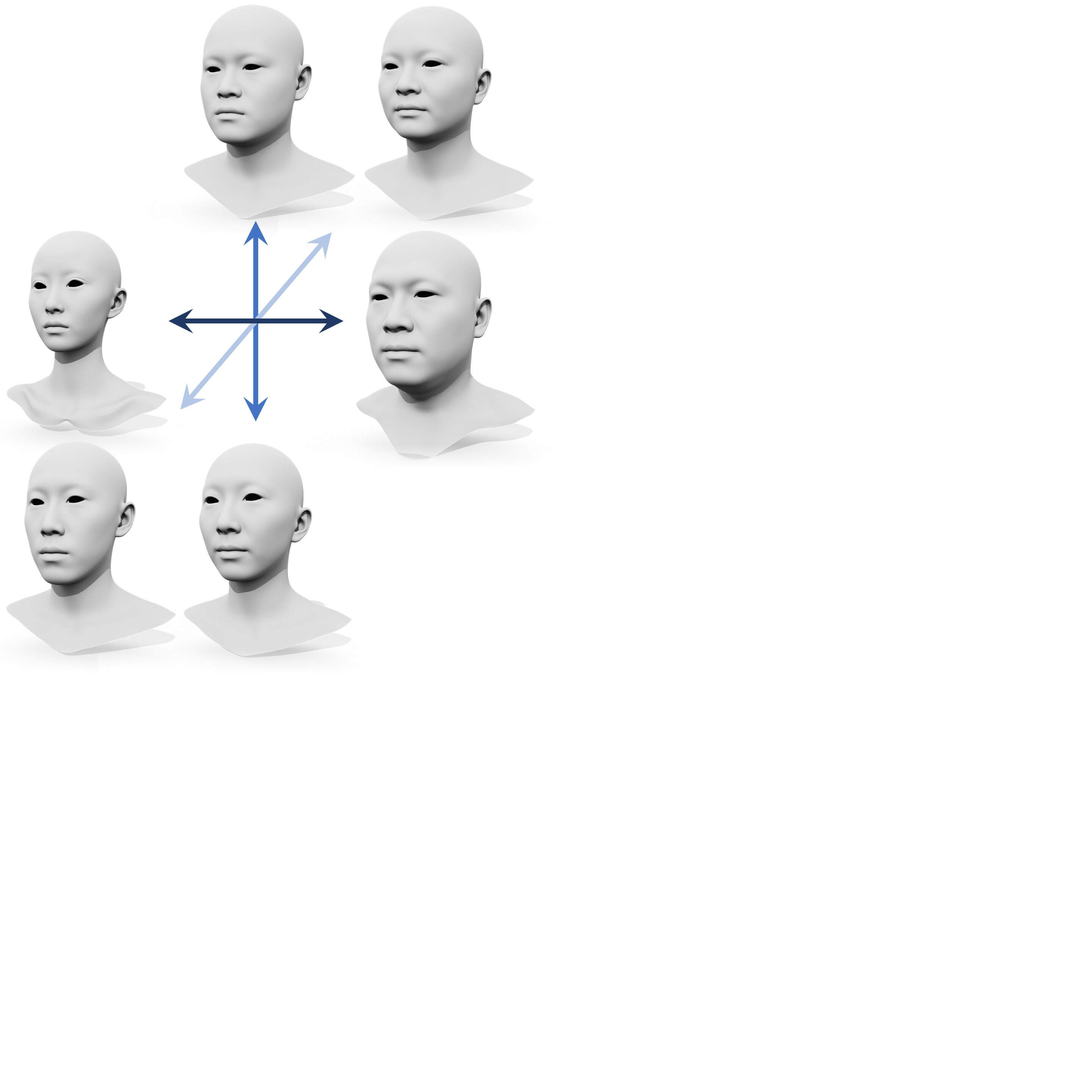}
    \put(0,80){\scriptsize\bfseries {\nextone}}
    \end{overpic}\hspace{10pt}
    \begin{overpic}[trim=-1cm 30cm 39cm -1cm,clip,width=0.3\linewidth,grid=false]{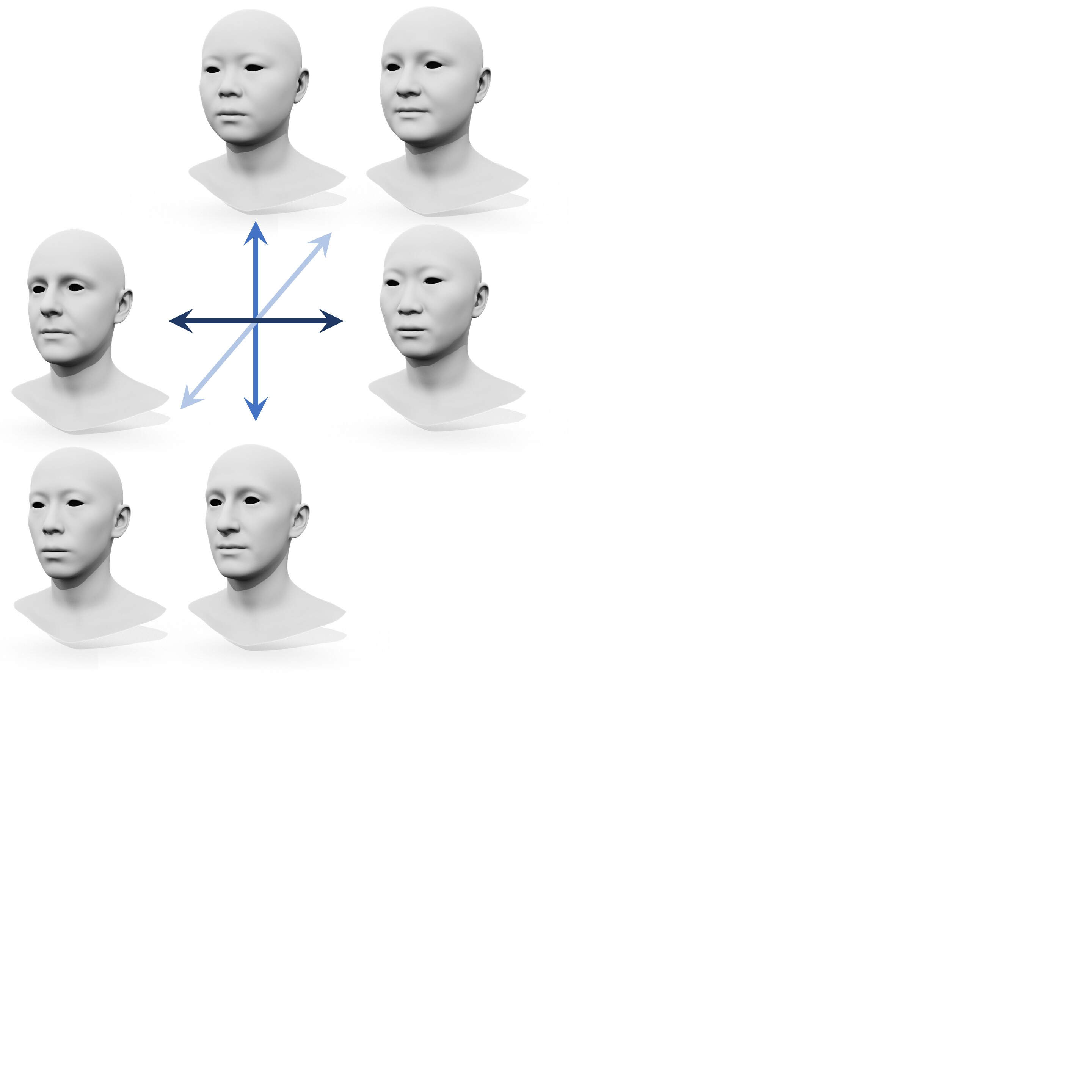}
    \put(0,80){\scriptsize\bfseries {\nextpp}}
    \put(7,70){\scriptsize\bfseries {(Ours)}}
    \end{overpic}
    \caption{\textbf{Model variations of different 3DMMs}. We show the shape (geometry) variation of BFM~\cite{bfm09} (\emph{Top left}), FLAME~\cite{flametopo} (\emph{Top middle}),  LSFM~\cite{3dmm10000} (\emph{Top right}), 
    LYHM~\cite{LYHM} (\emph{Bottom left}), {\nextone}~\cite{hifi3dface2021tencentailab} (\emph{Bottom middle}), the proposed {\nextpp} (\emph{Bottom right}). For shape variations, the first three principal components are visualized at $\pm 3$ standard deviations.}
    \label{fig:3dmm_var}
\end{figure*}

\myparagraph{Model Variations of Different 3DMMs}
Fig.~\ref{fig:3dmm_var} shows the shape variations of different 3DMMs. 
As we discussed in the main paper, previous 3DMMs have limited shape variations because of the imbalanced ethnic scans. 
In contrast, {\nextpp} is capable of expressing individuals in different ethnic, gender, and age groups with better generalization for downstream face reconstruction tasks.

\begin{figure}[!htp]
    \centering
    \begin{overpic}[trim=0cm 26cm 27cm 0cm,clip,width=1\linewidth,grid=false]{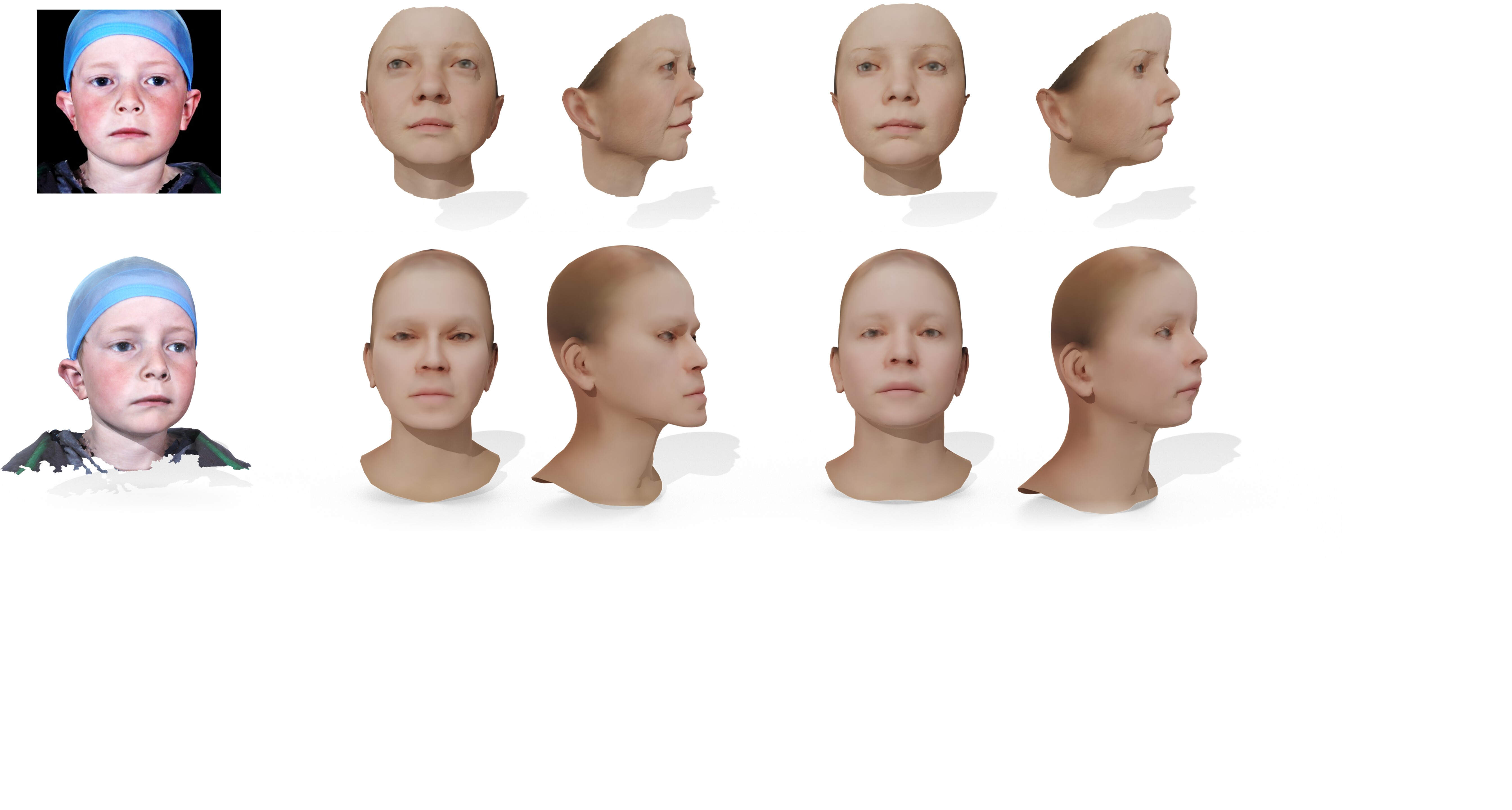}
    \put(7,44){\footnotesize Input}
    \put(1,23){\footnotesize Ground-truth}
    \put(34,44){\footnotesize Without Depth}
    \put(75,44){\footnotesize With Depth}
    \put(25,32){\footnotesize\bfseries \rotatebox{90}{BFM}}
    \put(25,8){\footnotesize\bfseries \rotatebox{90}{FLAME}}
    \end{overpic}
    \caption{Failure examples of RBG fitting and RGB-D fitting using FLAME~\cite{flametopo} and BFM~\cite{bfm09} 3DMMs. The reconstructed shapes may look reasonable in the frontal view but reveal their poor quality from other views. Providing depth information for fitting can significantly improve the reconstruction quality.}\label{fig:append:eg_failure}
\end{figure}

\myparagraph{BFM \& FLAME on RGB Fitting}
As shown in Fig.~\ref{fig:append:eg_failure}, the reconstructed faces from some 3DMMs (especially BFM~\cite{bfm09} and FLAME~\cite{flametopo}) on RGB Fitting from a single image can be misshapen for the following reasons:
(1) It has been acknowledged that 3D reconstruction from a 2D image is a severely ill-posed. In an under-constrained setting (such as without depth information), the reconstruction quality can be poor due to the limited expressiveness of the 3DMMs (such as the FLAME results shown in Fig.~\ref{fig:append:eg_failure}).
(2) On the other hand, the quality of the scans that are used for constructing 3DMMs can also affect the reconstruction quality. 
Take BFM as an example, the reconstructed example shown in Fig.~\ref{fig:append:eg_failure} has unnatural noise. 
As a comparison, LSFM that adopts the same topology as BFM achieves less noisy results with higher quality when we assign small weights to regularization terms for fitting using these two 3DMMs, since LSFM is constructed from larger number scans with higher quality.

\begin{figure*}[!t]
    \centering
    \begin{overpic}[trim=0cm 100cm 0cm 0cm,clip,width=1\linewidth,grid=false]{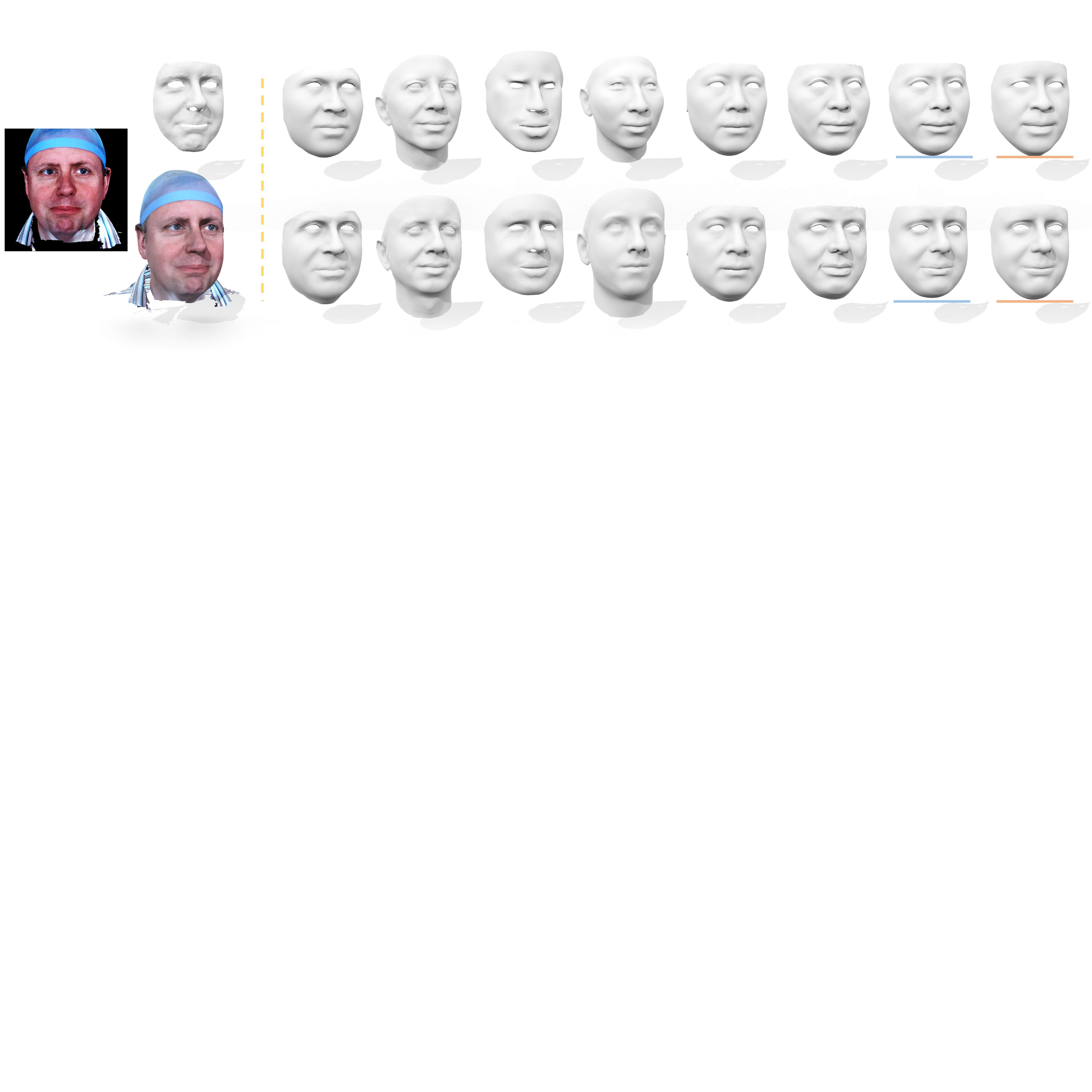}
        \put(3,20){\scriptsize\bfseries Input}
        \put(15,26){\scriptsize\bfseries G.T.}
        \put(26,26){\scriptsize\bfseries LYHM}
        \put(36,26){\scriptsize\bfseries BFM}
        \put(43,26){\scriptsize\bfseries FLAME}
        \put(54,26){\scriptsize\bfseries LSFM}
        \put(65,26){\scriptsize\bfseries FS}
        \put(71,26){\scriptsize\bfseries {\nextone}}
        \put(81,26){\scriptsize\bfseries {\nexttwo}}
        \put(93,26){\scriptsize\bfseries\itshape \underline{Ours}}
    \end{overpic}
    \begin{overpic}[trim=0cm 100cm 0cm 0cm,clip,width=1\linewidth,grid=false]{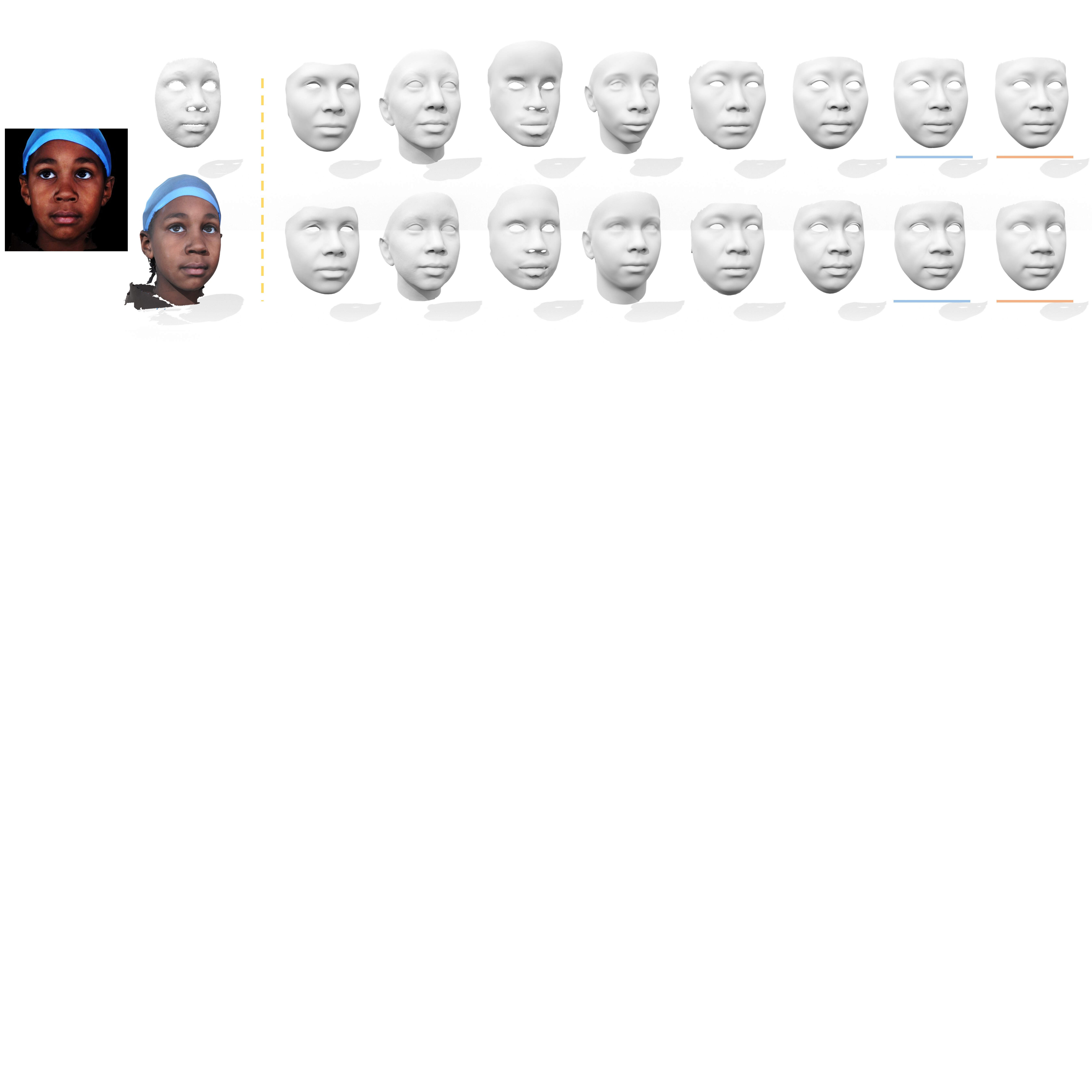}
    \end{overpic}
    \begin{overpic}[trim=0cm 100cm 0cm 0cm,clip,width=1\linewidth,grid=false]{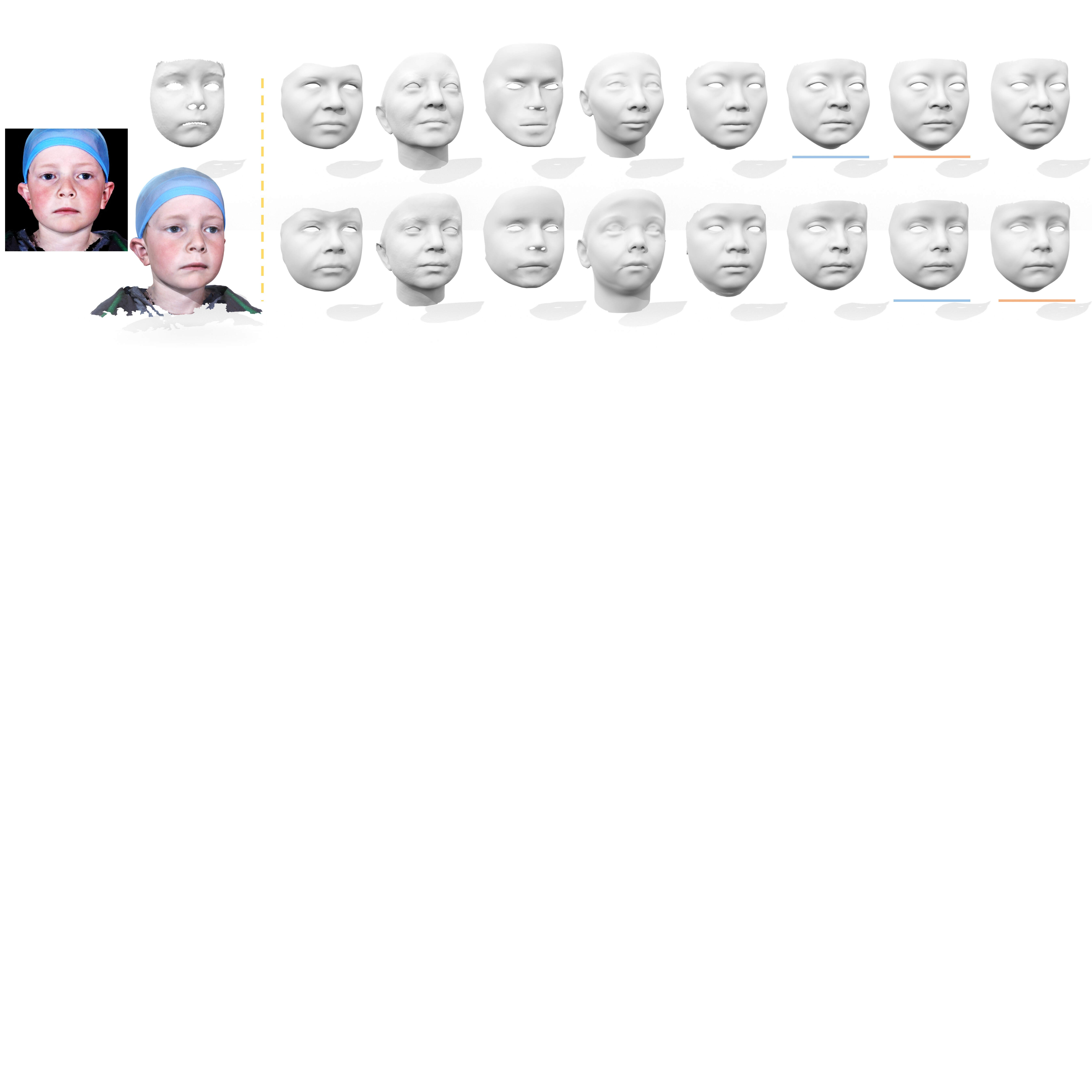}
    \end{overpic}
    \caption{\textbf{Comparing different 3DMMs with RGB(-D) fitting (part 1).} 
    We highlight the best (second best) reconstructed face via red (blue) underline chosen by the proposed evaluation pipeline, and {\nextpp} shows generally the most realistic face than others, quantitatively and perceptually.
    From left to right, LYHM~\cite{LYHM}, BFM~\cite{bfm09}, FLAME~\cite{flametopo}, LSFM~\cite{3dmm10000}, FaceScape~\cite{facescape}, {\nextone}~\cite{hifi3dface2021tencentailab}, {\nexttwo}~\cite{hifi3dface2021tencentailab}, and the proposed {\nextpp} are compared. The \emph{first} (\emph{second}) row of each sample shows the results of RGB (RGB-D) fitting.
    }
    \label{fig:apdx_3dmm_result1}
\end{figure*}

\begin{figure*}[!ht]
    \centering
    \begin{overpic}[trim=0cm 100cm 0cm 0cm,clip,width=1\linewidth,grid=false]{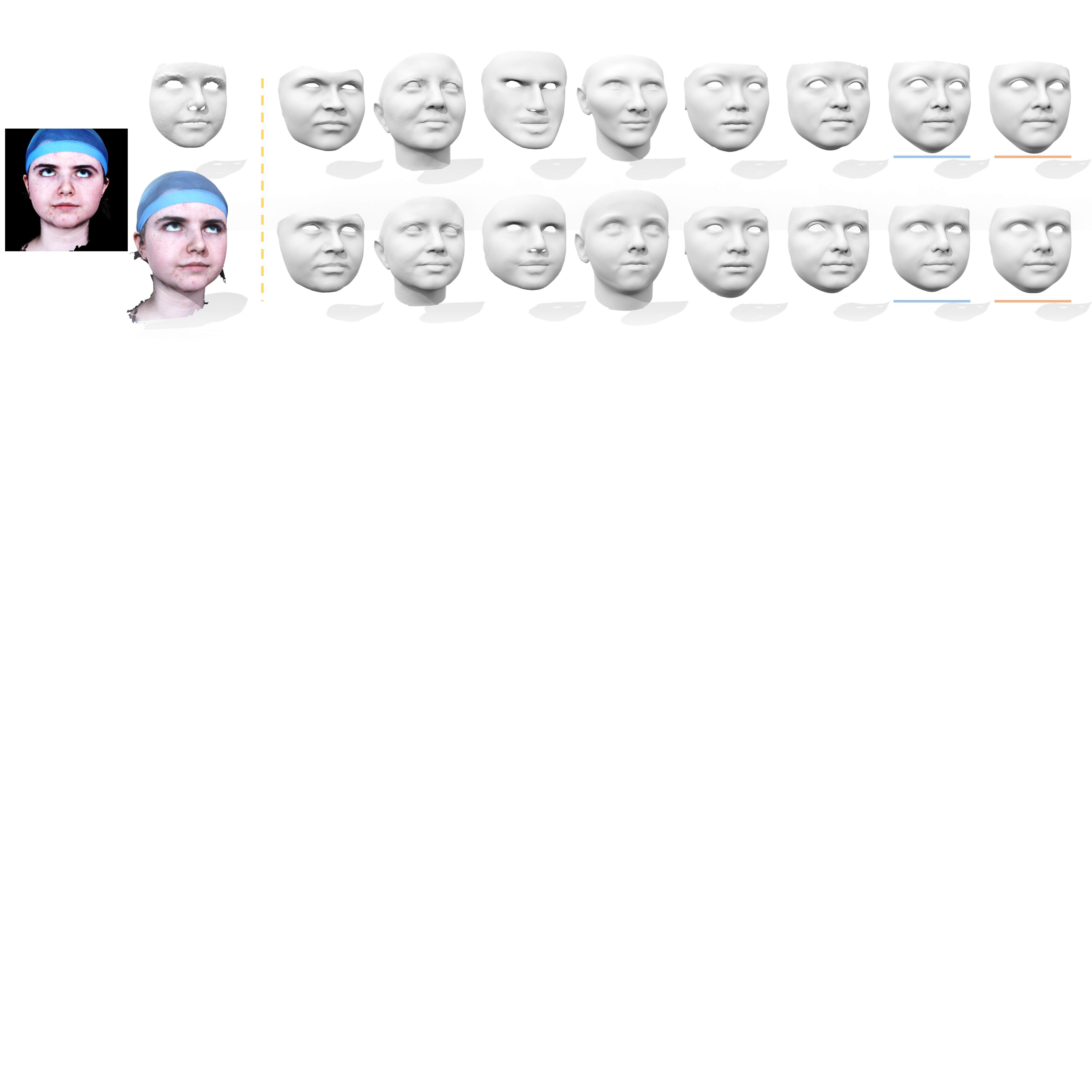}
        \put(3,20){\scriptsize\bfseries Input}
        \put(15,26){\scriptsize\bfseries G.T.}
        \put(26,26){\scriptsize\bfseries LYHM}
        \put(36,26){\scriptsize\bfseries BFM}
        \put(43,26){\scriptsize\bfseries FLAME}
        \put(54,26){\scriptsize\bfseries LSFM}
        \put(65,26){\scriptsize\bfseries FS}
        \put(71,26){\scriptsize\bfseries {\nextone}}
        \put(81,26){\scriptsize\bfseries {\nexttwo}}
        \put(93,26){\scriptsize\bfseries\itshape \underline{Ours}}
    \end{overpic}
    \begin{overpic}[trim=0cm 100cm 0cm 0cm,clip,width=1\linewidth,grid=false]{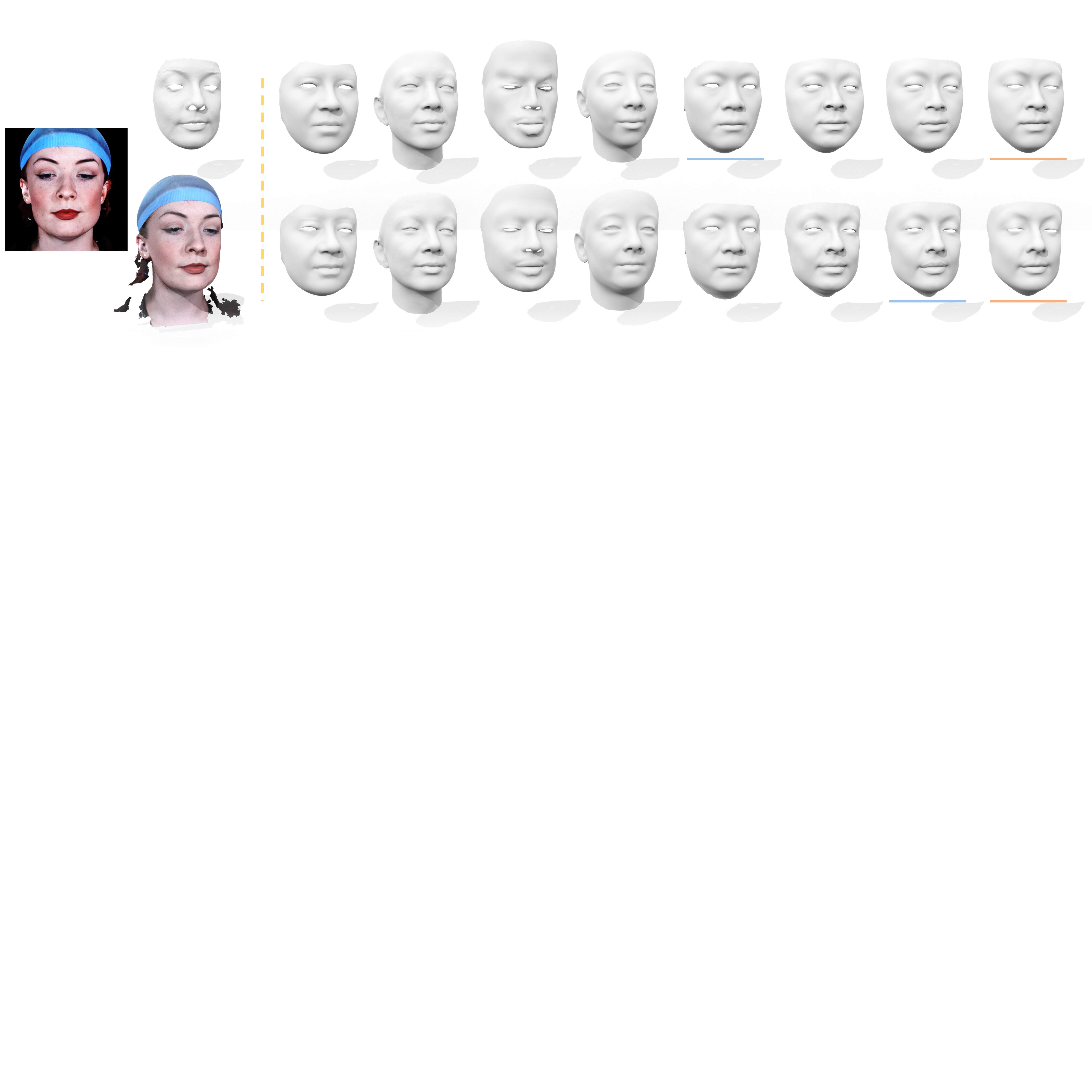}
    \end{overpic}
    \begin{overpic}[trim=0cm 100cm 0cm 0cm,clip,width=1\linewidth,grid=false]{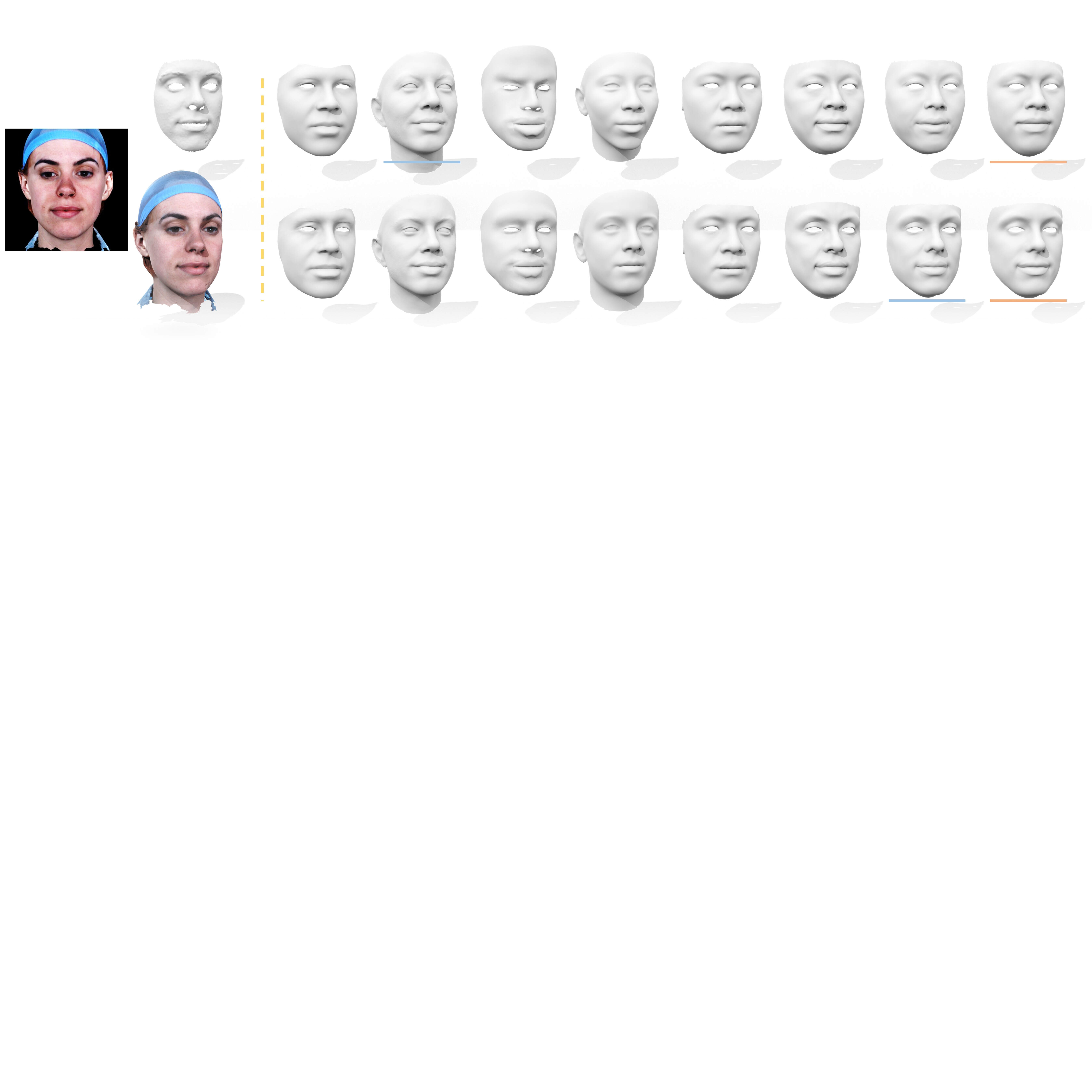}
    \end{overpic}
    \caption{\textbf{Comparing different 3DMMs with RGB(-D) fitting (part 2).} 
    We highlight the best (second best) reconstructed face via red (blue) underline chosen by the proposed evaluation pipeline, and {\nextpp} shows generally the most realistic face than others, quantitatively and perceptually.
    From left to right, LYHM~\cite{LYHM}, BFM~\cite{bfm09}, FLAME~\cite{flametopo}, LSFM~\cite{3dmm10000}, FaceScape~\cite{facescape}, {\nextone}~\cite{hifi3dface2021tencentailab}, {\nexttwo}~\cite{hifi3dface2021tencentailab}, and the proposed {\nextpp} are compared. The \emph{first} (\emph{second}) row of each sample shows the results of RGB (RGB-D) fitting.
    }
    \label{fig:apdx_3dmm_result2}
\end{figure*}

\begin{figure*}[!h]
    \centering
    \begin{overpic}[trim=0cm 2cm 0cm 0cm,clip,width=1\linewidth,grid=false]{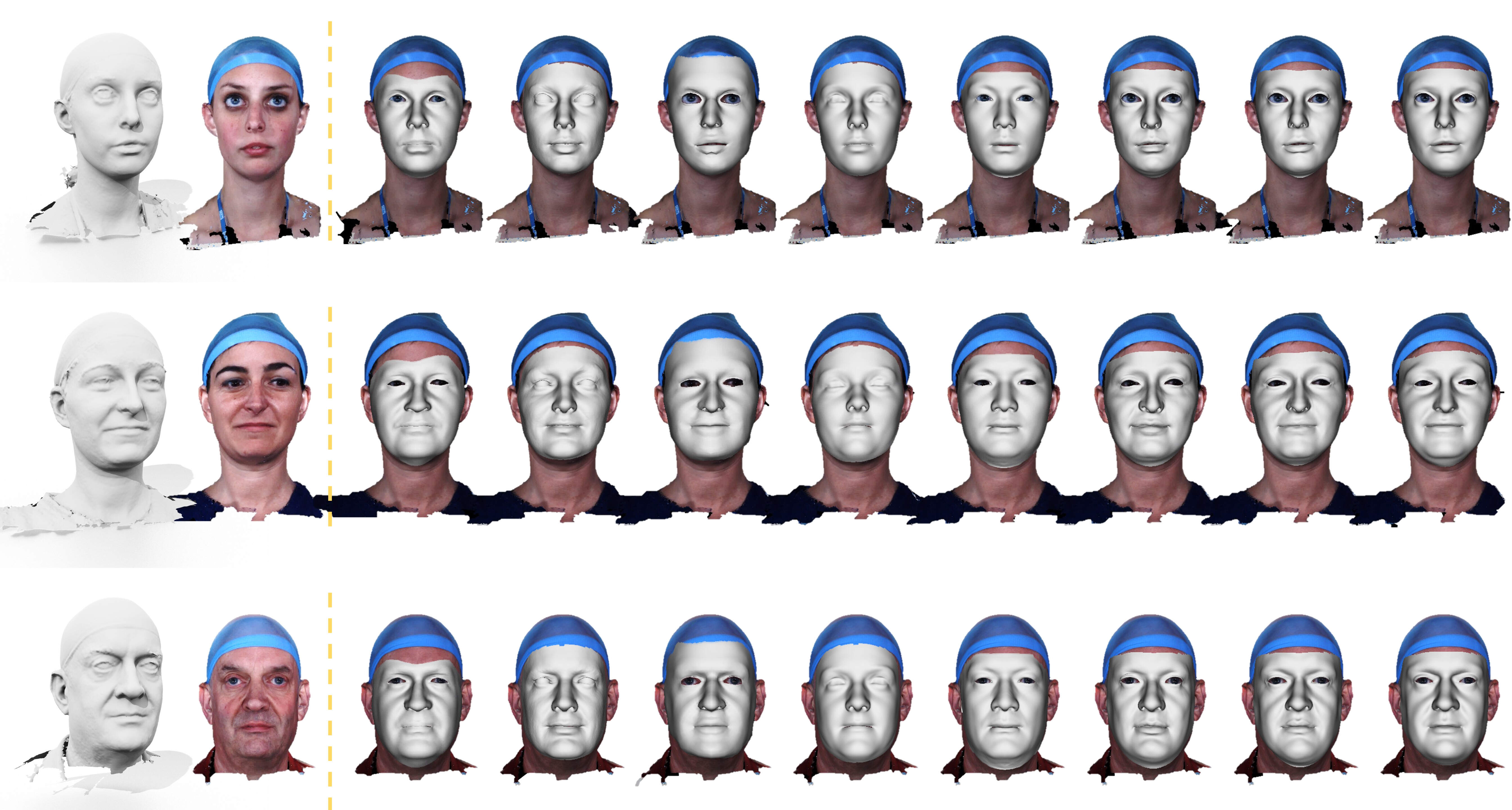}
        \put(4,52){\scriptsize\bfseries G.T.}
        \put(14,52){\scriptsize\bfseries Input}
        \put(24,52){\scriptsize\bfseries LYHM}
        \put(34,52){\scriptsize\bfseries BFM}
        \put(43,52){\scriptsize\bfseries FLAME}
        \put(54,52){\scriptsize\bfseries LSFM}
        \put(65,52){\scriptsize\bfseries FS}
        \put(71,52){\scriptsize\bfseries {\nextone}}
        \put(81,52){\scriptsize\bfseries {\nexttwo}}
        \put(93,52){\scriptsize\bfseries\itshape \underline{Ours}}
    \end{overpic}
    \begin{overpic}[trim=0cm 2cm 0cm 0cm,clip,width=1\linewidth,grid=false]{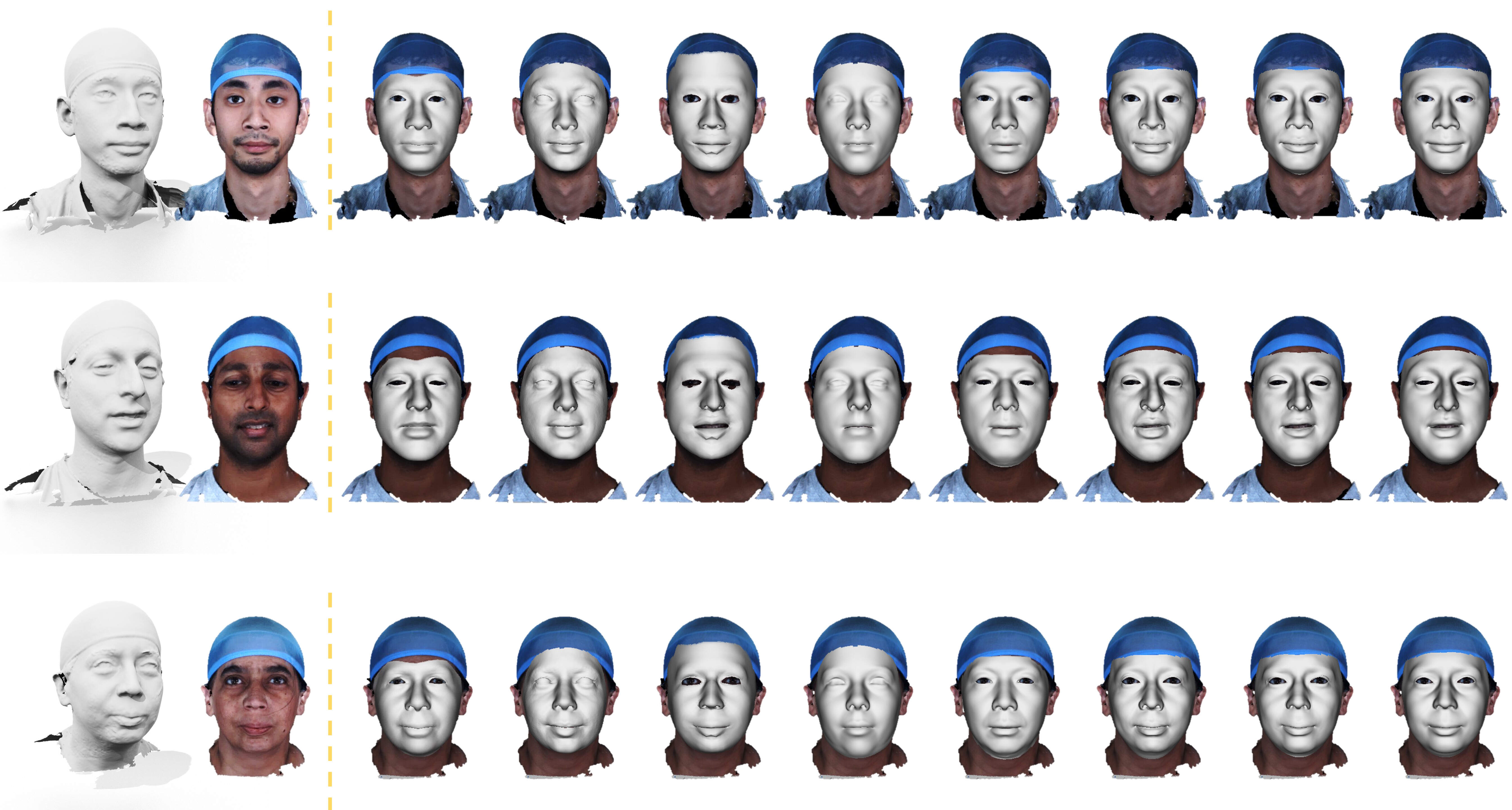}
    \end{overpic}
    \caption{\textbf{Comparing different 3DMMs with RGB-D fitting (part 3).}
    We visualize rendered shapes on input image via predicted camera parmeters. 
    {\nextpp} shows best visualized results with high fidelity features among other 3DMMs.
    From left to right, LYHM~\cite{LYHM}, BFM~\cite{bfm09}, FLAME~\cite{flametopo}, LSFM~\cite{3dmm10000}, FaceScape~\cite{facescape}, {\nextone}~\cite{hifi3dface2021tencentailab}, {\nexttwo}~\cite{hifi3dface2021tencentailab}, and the proposed {\nextpp} are compared.
    }
    \label{fig:append:3dmm_image}
\end{figure*}

\myparagraph{{\nextpp} on RGB-D Fitting}
In order to avoid reconstructing misshapen faces, RGB fitting relies more on face prior (i.e., impose stronger regularization on the 3DMM parameter $\alpha$) due to its access to only limited geometry supervision.
In contrast, RGB-D fitting
relies more on the denser and more informative depth supervision,
which is more suitable to evaluate the expressiveness different 3DMMs. 
In Fig.~\ref{fig:apdx_3dmm_result1}, Fig.~\ref{fig:apdx_3dmm_result2}, and Fig.~\ref{fig:append:3dmm_image}, we show face reconstruction results using different 3DMMs on RGB and RGB-D fitting.
And indeed, {\nextpp} outperforms the other 3DMMs on RGB-D fitting.

\myparagraph{Conclusion}
Thanks to our region-based evaluation pipeline, we find that a 3DMM can have different levels of expressiveness in different regions. 
For example, FLAME~\cite{flametopo} can reconstruct the nose region well, but fail to express the overall shape and the curvature in the forehead region. 
We believe it is promising to investigate how to construct a region-aware 3DMM in the future work, which integrates the advantages of different 3DMMs and introduces more fine-grained prior for 3D face reconstruction.

\section{Implementation Details}\label{append:implementation}

\subsection{Details of RGB(-D) Regression}

\begin{table*}[!t]
\begin{minipage}[t]{0.49\linewidth}
\caption{Loss parameters in Eq.~\eqref{eq:totalloss} for RGB fitting.}\label{tab:param_rgb}
\resizebox{\textwidth}{!}{%
\input{latex/tables/table_apdx_params_rgb}
}
\end{minipage}
\hfill
\begin{minipage}[t]{0.51\linewidth}
\caption{Loss parameters in Eq.~\eqref{eq:totalloss} for RGB-D fitting.}\label{tab:param}
\resizebox{\textwidth}{!}{%
\input{latex/tables/table_apdx_params}
}
\end{minipage}
\begin{tablenotes}
\item \footnotesize{*FaceScape does not converge with small $\omega_{\text{shp}}$ weight.}
\end{tablenotes}
\end{table*}

For the baselines, we use the officially released 3DMMs including the shape basis and the texture basis. 
For our newly introduced {\nextpp}, we use the same texture basis as {\nextone} and {\nexttwo}.
Following~\cite{hifi3dface2021tencentailab}, we render images with the estimated parameters (i.e., 3DMM shape \& texture parameters, second-order spherical harmonics lighting parameters, and pose parameters) via a differentiable renderer~\cite{unsupcvpr2018}, and adopt RGB photo loss, depth loss, identity perceptual loss, landmark loss and regularization terms to optimize these parameters, which are defined as follows.

\myparagraph{RGB Photo Loss.} The pixel-wise RGB photometric loss is defined as:
\begin{equation}
    \begin{aligned}
        L_{\text{rgb}}=\big\Vert I_{\text{rgb}}-\hat{I}_{\text{render}} \big\Vert_2
    \end{aligned}
    \label{eq:rgbloss}
\end{equation}
where $\Scale[0.9]{I_{\text{rgb}}}$ is the input RGB image, $\Scale[0.9]{\hat{I}_{\text{render}}}$ is the rendered RGB image from the differentiable renderer using the predicted parameters. 
We adopt $\Scale[0.9]{l_{2,1}}$-norm for its demonstrated robustness against outliers than $\Scale[0.9]{l_2}$-norm~\cite{hifi3dface2021tencentailab}.

\myparagraph{Depth Loss.} The depth loss is defined as:
\begin{equation}
    \begin{aligned}
        L_{\text{dep}}=\rho \Big(\big\Vert I_{\text{dep}}-\hat{I}_{z} \big\Vert_2^2\Big)
    \end{aligned}
    \label{eq:depthloss}
\end{equation}
where $\rho(\cdot)$ defines a truncated $\Scale[0.9]{l_2}$-norm that clips the per-pixel MSE, $\Scale[0.9]{I_{\text{dep}}}$ is the input depth image, $\Scale[0.9]{\hat{I}_{z}}$ is the rendered depth image from the differentiable renderer. The truncated function is proved to be more robust to depth outliers~\cite{hifi3dface2021tencentailab}.

\myparagraph{Identity Perceptual Loss.} To capture high-level identity information, we apply the following identity perceptual loss:
\begin{equation}
    \begin{aligned}
        L_{\text{id}}=\big\Vert \psi\big(I_{\text{rgb}}\big)-\psi\big(\hat{I}_{\text{render}}\big) \big\Vert_2^2
    \end{aligned}
    \label{eq:idloss}
\end{equation}
where $\psi(\cdot)$ is the high-level identity features exacted from a pretrained face recognition model. In our experiments, we use features from the $fc7$ layer of VGGFace model~\cite{FaceVGG}.

\myparagraph{Landmark loss.} To achieve better fitting quality, we ask professional artist to extend the $68$ keypoints defined on each 3DMM into $86$\footnote{Corresponding to the third set of $86$ keypoints we discussed in Sec.~\ref{append:keypointdefine}} keypoints. 
Landmark loss is defined as the weighted average distances between the detected 2D landmarks and the projected landmarks from the predicted 3D model.
\begin{equation}
    \begin{aligned}
        L_{\text{lmk}}=\frac{1}{\mathcal{F}}\sum_{f_j\in \mathcal{F}} \omega_j \big\Vert f_j-\Pi (\Phi(v_j))\big\Vert_2^2
    \end{aligned}
    \label{eq:lmkloss}
\end{equation}
where $\Scale[0.9]{f_j \in \mathcal{F}}$ are the detected landmarks, 
$\Pi(\cdot)$ represents world-to-image plane projection with given camera parameters, $\Phi(\cdot)$ represents 6DoF head pose that rigidly rotates and translates the mesh, and $\Scale[0.9]{v_j}$ represent keypoints on the mesh.
The weight $\Scale[0.9]{\omega_j}$ controls the importance of each keypoint, where we set $50$ for those located in eye, nose and mouth region, and $1$ otherwise.

\myparagraph{Regularization.} To ensure the plausibility of the reconstructed faces,
we apply regularization to the shape and texture parameters:
\begin{equation}
    \begin{aligned}
        L_{\text{reg}}=\omega_{\text{shp}} \big\Vert \alpha_{\text{shp}} - \alpha_{\text{shp}}^\mu \big\Vert_2^2 + \omega_{\text{tex}} \big\Vert \alpha_{\text{tex}} - \alpha_{\text{tex}}^{\mu} \big\Vert_2^2 
    \end{aligned}
    \label{eq:regloss}
\end{equation}
where $\Scale[0.9]{\alpha_{\text{shp}}}$ / $\Scale[0.9]{\alpha_{\text{tex}}}$ and $\Scale[0.9]{\alpha_{\text{shp}}^{\mu}}$/$\Scale[0.9]{\alpha_{\text{tex}}^{\mu}}$ represent the predict shape/texture parameters and mean face shape/texture parameters respectively.

The final total loss function to be minimized is defined as the weighted sum of each part:
\begin{equation}
        L_{\text{total}}=\omega_{\text{rgb}} L_{\text{rgb}} + \omega_{\text{dep}} L_{\text{dep}} + \omega_{\text{id}}L_{\text{id}} +\omega_{\text{lmk}}L_{\text{lmk}}+L_{\text{reg}}
    \label{eq:totalloss}
\end{equation}

\myparagraph{Parameters}
In our experiments, we use Adam optimizer~\cite{adam} in TensorFlow for $1000$ iterations with a learning rate $0.05$ decaying exponentially in every $50$ iterations.
We fine-tune the parameters $\Scale[0.9]{\omega_{\text{rgb}},\omega_{\text{dep}},\omega_{\text{id}},\omega_{\text{lmk}},\omega_{\text{shp}},\omega_{tex}}$ for each 3DMM and use the best combinations as reported in Tab.~\ref{tab:param_rgb} and \ref{tab:param}.

\subsection{Parameters for the Evaluation Pipeline}

We implement our evaluation pipeline in Python. It takes about $3$ hours to evaluate $100$ individuals in {\name} on a single Intel i7-9700 CPU for each method, including both the global wise error and $4$ region-aware error. Taking BFM with $35709$ vertex (Deep3D) as an example, the baseline {\gicp} takes $0.81$ minute while ours takes $1.73$ minute per sample for evaluation. Specifically, for our evaluation pipeline, it takes $0.48$ second for keypoint alignment, $1.28$ minute for region alignment ({\ricp}), and $0.45$ minute for error computation ({\bicp}).

In {\ricp}, we set the maximum number of iterations to $100$ and stop early if the change of matching error is less than $10^{-6}$. 
The weight $\Scale[0.9]{w_\mathcal{K}}$ (in Algo.~1 in the main text) is set to the ratio between the number of vertex in $\Scale[0.9]{\mathcal{R}_H}$ and in $\Scale[0.9]{\mathcal{K}_P}$. 
We use a two-stage {\nicp} approach (Algo.~2, step 3 in the main text) to avoid unsatisfactory deformation results in {\bicp}. 
At the first stage, we only include the landmark term and the stiffness term for initial deformation with weight $50$ and $150$ respectively. 
At the second stage we include all the terms including distance term, landmark term, and stiffness term with weight $1$, $5$, $50$ respectively. 
In both stages, the weight of the stiffness term gradually decays to allow for more localized deformations.

\subsection{Experiments Setup}
\myparagraph{Preprocess Input Image.} For learning-based face reconstruction, we apply MTCNN~\cite{MTCNN} to detect and crop the face region of the input frontal images provided by our {\name}, and resize them into a resolution of $300 \times 300$. As for RGB(-D) fitting, we resize input images to $512\times 512$ without cropping.

\myparagraph{Baselines for Face Reconstruction}
We use the officially pretrained model released by the previous work~\cite{ExpNet,RingNet,MGCNet,PRNet,deep3d,3DDFA_v2,ganfit,Nonlinear3DMM,DECA} for the face reconstruction experiments. 
We make sure that none of these methods have seen or fine-tuned on the tested images in {\name}.
We compare different methods by evaluating the similarity between the reconstructed face and the ground-truth from {\name} using our evaluation pipeline. 
Note that only the facial region of each method is considered during the evaluation phase. 
Moreover, we also reorder the detected $68$ keypoints of each method such that they are in correspondences with the $68$ keypoints on the ground-truth for proper evaluation. 
Recall that our evaluation pipeline is based on predefined region masks where the eyeballs, nostrils, and mouth cavity are not considered for more fair comparisons
since some reconstructed faces (e.g., FLAME) do not have eyeballs or nostrils or mouth cavity. 

\myparagraph{Error in Metric Units}
During the evaluating, we rescale the $\Scale[0.9]{S_H}$ in {\name} and the predicted shapes $\Scale[0.9]{S_P}$ back to its original size in LYHM~\cite{LYHM} such that the shape difference between $\Scale[0.9]{S_H}$ and $\Scale[0.9]{S_P}$ is measured in proper metric units and reflects real-world difference.

\section{Limitation \& Future Work}\label{append:limitaion}
Our work still has some limitations. 
Although using in-the-lab images produced with controlled configurations
can faithfully reflect the reconstruction ability of existing methods, the robustness of different methods is not investigated.
We leave it for future work by generating more challenging images with different variations (e.g., expressions, backgrounds and occlusions) and extending our {\name} benchmark to in-the-wild environment.
Besides, our evaluation pipeline is computationally expensive since it requires alignment and deformation of each of the $4$ regions.
In the future, we would like to investigate more powerful and more efficient evaluation approaches.

%
%
\bibliographystyle{splncs04}
\bibliography{egbib}
\end{document}

%% file: latex/header.tex
\usepackage{graphicx}
\usepackage{amsmath}
\usepackage{amssymb}
\usepackage{booktabs}
\usepackage{enumitem}
\usepackage{dsfont}
\usepackage{algorithm, algorithmic}
\algsetup{linenosize=\small}
\usepackage{tablefootnote}
\usepackage[flushleft]{threeparttable} 

\usepackage{xcolor}
\usepackage{color,colortbl}
\usepackage{overpic}
\usepackage{wrapfig}
\usepackage[normalem]{ulem}

\input{insbox.tex}
\newcommand*{\Scale}[2][4]{\scalebox{#1}{$#2$}}%

\newcommand{\name}{REALY}

\newcommand{\nextone}{HIFI3D}
\newcommand{\nexttwo}{HIFI3D$^A$}
\newcommand{\nextpp}{HIFI3D$^{\pmb{+}\pmb{+}}$}

\newcommand{\tempshape}{$S_{\textnormal{temp}}$}
\newcommand{\tempkpt}{$\mathcal{K}_{\textnormal{temp}}$}
\newcommand{\tempregion}{$\mathcal{R}_{\textnormal{temp}}$}

\newcommand{\gicp}{\small\bfseries{\textit{g}ICP}}
\newcommand{\ricp}{\small\bfseries{\textit{r}ICP}}
\newcommand{\nicp}{\small\bfseries{\textit{n}ICP}}
\newcommand{\bicp}{\small\bfseries{\textit{b}ICP}}

\usepackage{multirow}
\usepackage{multicol}
\usepackage{soul}
\setul{0.5ex}{0.3ex}

\newcommand{\myparagraph}[1]{\vspace{2mm}\noindent\textbf{#1}}

\definecolor{col_nose}{RGB}{181,228,140}
\definecolor{col_mouth}{RGB}{255,183,3}
\definecolor{col_forehead}{RGB}{189,178,255}
\definecolor{col_cheek}{RGB}{144,224,239}
\definecolor{col_table}{RGB}{175,227,246} 

\definecolor{mycolor1}{rgb}{0.85000,0.32500,0.09800}%
\definecolor{mycolor2}{rgb}{0.92900,0.69400,0.12500}%
\definecolor{mycolor3}{rgb}{0.49400,0.18400,0.55600}%
\definecolor{mycolor4}{rgb}{0.87843,0.76471,0.98824}%
\definecolor{mycolor5}{rgb}{0.46600,0.67400,0.18800}%
\definecolor{mycolor6}{rgb}{0.30100,0.74500,0.93300}%
\definecolor{mycolor7}{rgb}{0.00000,0.44700,0.74100}%

\definecolor{best_two}{RGB}{72,149,239} 
\definecolor{best}{RGB}{179,11,0} 

\newcommand{\mymap}[3]{T_{{#1 \rightarrow #2}}^{ \ifthenelse{\equal{#3}{}}{}{ \,\text{#3} }}}
\DeclareMathOperator*{\argmin}{arg\,min}

\newcommand{\rot}{\mathbf{R}}
\newcommand{\tran}{\mathbf{t}}

\usepackage{orcidlink}

\usepackage{hyperref}

\usepackage[capitalize]{cleveref}
\crefname{section}{Sec.}{Secs.}
\Crefname{section}{Section}{Sections}
\Crefname{table}{Table}{Tables}
\crefname{table}{Tab.}{Tabs.}

\usepackage{pgf,tikz}

\usetikzlibrary{shapes,arrows,chains}
\usetikzlibrary{calc}

%% file: latex/tables/table_3DMM_cmp.tex
\resizebox{1\linewidth}{!}{%
\begin{tabular}{c|cccccccc|c}
\toprule[1pt]
        & BFM~\cite{bfm09}  & FWH~\cite{FaceWarehouse} & FLAME~\cite{flametopo} & LSFM~\cite{3dmm10000} & LYHM~\cite{LYHM} & FS~\cite{facescape} &{\nextone}~\cite{hifi3dface2021tencentailab} & {\nexttwo}~\cite{hifi3dface2021tencentailab}  & \textit{\textbf{Ours}} \\ \midrule[1pt]
\# scans & 200    & 140           & 3800  & 8402 &{1212}  & {938}           & 200   & 200   & 1957    \\
\rowcolor{col_table!20} 
$n_v$    & 53215  & 11510         & 5023  & 53215 & {11510}   & {26317}        & 20481  & 20481 & 20481  \\
$n_f$    & 105840 & 22800         & 9976  & 105840 &  {22800}   & {52261}      & 40832 & 40832 & 40832  \\
\rowcolor{col_table!20}
\# basis & 199    & 50            & 300   & 158 &   {100}  & {300}         & 200  &  500  & 526  \\
\bottomrule[1pt]
\end{tabular}%
}
\begin{tablenotes}
\scriptsize
\item {\nexttwo} stands for the ``augmented'' version of {\nextone}~\cite{hifi3dface2021tencentailab}, which employs data augmentation techniques to construct 3DMM from $200$ scans.
\end{tablenotes}


%% file: latex/algos/alg_ricp.tex
\begin{algorithm}[!t]
    \caption{$\Scale[0.9]{S_{P}^{*} =}$  {\ricp}$\Scale[0.9]{\big( S_P \rightarrow S_H}$@$\Scale[0.9]{\mathcal{R}_H  \big)}$}
    \label{Alg:myicp}
	\renewcommand{\algorithmicrequire}{\textbf{Input:}}
	\renewcommand{\algorithmicensure}{\textbf{Goal}}
	\algsetup{linenosize=\scriptsize}
	\begin{algorithmic}[1]
	\footnotesize
		\ENSURE Rigidly align $S_P$ to the \emph{region} $R_H$ of $S_H$.
		\REQUIRE High-res mesh $S_H$ with region $R_H$ and keypoints $\mathcal{K}_H$; a predicted mesh $S_P$ with keypoints $\mathcal{K}_P$; weights $w_\mathcal{K}$ for keypoints alignment; maximum iteration $\mathbb{K}$.
		\STATE $\Scale[0.9]{S_{P}^{(0)} =}$  \textbf{\textit{g}ICP} $\Scale[0.9]{\big( S_P \rightarrow S_H\big)}\quad$ %
		\FOR{$0\le k \le \mathbb{K}$}
		\STATE Find nn-map $\Scale[0.9]{T}$ from region $\Scale[0.9]{\mathcal{R}_H}$ to $\Scale[0.9]{S_P^{(k)}}$.
		\STATE Solve $\Scale[0.9]{[\rot, \tran] = \argmin \sum\limits_{v\in\mathcal{R}_H}\big\Vert \rot T(v) + \tran - v\big\Vert_F^2 + w_{\mathcal{K}} \big\Vert \rot\mathcal{K}_P + \tran - \mathcal{K}_H \big\Vert_F^2}$.
		\STATE Obtain $\Scale[0.9]{S_p^{(k+1)}}$ by transforming $\Scale[0.9]{S_p^{(k)}}$ via $\Scale[0.9]{(\rot, \tran)}$.
		\ENDFOR
		\STATE Set $\Scale[0.9]{S_{P}^{*} \leftarrow S_{P}^{(\mathbb{K})}}$. %
	\end{algorithmic} 
\end{algorithm}

%% file: latex/algos/alg_bicp.tex
\begin{algorithm}[!t]
    \caption{$\Scale[0.9]{\big[S_{P}^{*}, \mathcal{R}_{H}^{*}\big] =}$  {\bicp}$\Scale[0.9]{\big( S_P \leftrightarrow S_H}$@$\Scale[0.9]{\mathcal{R}_H  \big)}$}
    \label{Alg:bicp}
	\renewcommand{\algorithmicrequire}{\textbf{Input:}}
	\renewcommand{\algorithmicensure}{\textbf{Goal}}
	\algsetup{linenosize=\scriptsize}
	\begin{algorithmic}[1]
	\footnotesize
		\ENSURE Rigidly align $S_P$ and non-rigidly deform $\mathcal{R}_H$ for better alignment in region $\mathcal{R}_H$
		\STATE $\Scale[0.9]{S_{P}^{*} =}$  \textbf{\textit{r}ICP}$\Scale[0.9]{\big( S_P \rightarrow S_H}$@$\Scale[0.9]{\mathcal{R}_H  \big)}$ {\textcolor{gray}{\itshape\footnotesize{\% call Algo.~\ref{Alg:myicp}}}}
		\STATE Find nn-map $\Scale[0.9]{T}$ from region $\Scale[0.9]{\mathcal{R}_H}$ to $\Scale[0.9]{S_{P}^{*}}$
		\STATE $\Scale[0.9]{\mathcal{R}_H^* = }$ \textbf{\textit{n}ICP}$\Scale[0.9]{\Big( \mathcal{R}_H \,\big\vert\, V_{\mathcal{R}_H} \rightarrow T\big(V_{\mathcal{R}_H}\big) \Big)} + \Scale[0.9]{w_{\mathcal{K}}}$ \textbf{\textit{n}ICP}$\Scale[0.9]{\Big( \mathcal{R}_H \,\big\vert\, \mathcal{K}_H \rightarrow \mathcal{K}_P \Big)} $
		{\textcolor{gray}{\itshape\footnotesize{\% where \textbf{\textit{n}ICP}$\Scale[0.8]{\big( S \,\vert\, X \rightarrow  Y\big)}$ applies non-rigid ICP to deform $\Scale[0.8]{S}$ such that the points $\Scale[0.8]{X}$ on $\Scale[0.8]{S}$ are expected to be mapped to new positions $\Scale[0.9]{Y}$}}}
	\end{algorithmic} 
\end{algorithm}

%% file: latex/tables/table_toy_exp.tex
\begin{minipage}[!t]{1\columnwidth}
\begin{minipage}[!t]{0.68\columnwidth}
\caption{The distance $e(\cdot)$ between $S_P$ (aligned by {\ricp}/{\gicp}, or with G.T. alignment) and $S_L$ of the four examples in Fig.~\ref{fig:replace_sample}.} 
\label{tab:replace_exp}
\resizebox{\linewidth}{!}{%
\begin{tabular}{l|ccccc|c|c}
\toprule[1pt]
\multicolumn{1}{c|}{$\mathcal{R}^{\text{rm}}$ / } & \multicolumn{5}{c|}{\footnotesize\bfseries{\textit{r}ICP} (ours)} & \multicolumn{1}{c|}{\footnotesize\bfseries{\textit{g}ICP}} & \footnotesize \bfseries G.T.\\ 
\multicolumn{1}{c|}{$e (\text{mm}/10)$}  & \setulcolor{col_nose} \textbf{@\ul{nose}}     & \setulcolor{col_mouth}\textbf{@\ul{mouth}}  & \setulcolor{col_forehead} \textbf{@\ul{forehead}}   &   \multicolumn{1}{c}{\setulcolor{col_cheek} \textbf{@\ul{cheek}}}  &    \multicolumn{1}{c|}{\setulcolor{gray}  \textbf{\ul{all}}} & $\Scale[0.85]{e\big(\mymap{p}{\textcolor{red}{l}}{\textcolor{blue}{vtx}}\big)}$  &
$\Scale[0.85]{e\big(\mymap{p}{\textcolor{red}{l}}{\textcolor{blue}{vtx}}\big)}$
\\ \midrule[1pt]
\setulcolor{col_nose}\textbf{\ul{nose}} & \cellcolor{col_nose!30}\textbf{8.376} & 0.331          & 1.200          & 1.484          & 11.392                 & 40.490        & 11.972                  \\
\setulcolor{col_mouth}\textbf{\ul{mouth}}                      & 0.550          &\cellcolor{col_mouth!30} \textbf{4.372} & 1.164         & 3.053          & 9.139                 & 20.889         & 5.195                      \\
\setulcolor{col_forehead}\textbf{\ul{forehead}}                    & 0.636          & 0.165          & \cellcolor{col_forehead!30}\textbf{7.107} & 0.630          & 8.537                 & 12.695           & 6.463                      \\
\setulcolor{col_cheek}\textbf{\ul{cheek}}                        & 1.417         & 0.397          & 0.547          & \cellcolor{col_cheek!30}\textbf{3.631} & 5.992                 & 18.754            & 3.943            \\ \bottomrule[1pt]
\end{tabular}
}
\end{minipage}
\hfill
\begin{minipage}[!t]{0.31\columnwidth}
\caption{Error of the nn-map obtained via different ICPs (Fig.~\ref{fig:replace_sample}).}\label{tab:icp_cmp_exp}
\resizebox{\linewidth}{!}{%
\begin{tabular}{l|ccc}
\toprule[1pt]
\multicolumn{1}{c|}{$\mathcal{R}^{\text{rm}}$ /} & \multirow{2}{*}{\footnotesize\bfseries{\textit{g}ICP}} & \multirow{2}{*}{\footnotesize\bfseries{\textit{r}ICP}} & \multirow{2}{*}{\footnotesize\bfseries{\textit{b}ICP}}\\
\multicolumn{1}{c|}{$e (\text{mm})$}  &                    &       &           \\ \midrule[1pt]
\setulcolor{col_nose}\textbf{\ul{nose}}                  & 2.882 & 0.706 & \textbf{0.670} \\
\setulcolor{col_mouth}\textbf{\ul{mouth}}                & 1.699 & 0.459 & \textbf{0.407} \\
\setulcolor{col_forehead}\textbf{\ul{forehead}}             & 0.707 & 0.581 & \textbf{0.520} \\
\setulcolor{col_cheek}\textbf{\ul{cheek}}                & 1.219 & 0.107 & \textbf{0.105} \\ \bottomrule[1pt]
\end{tabular}
}
\end{minipage}
\end{minipage}

%% file: latex/img_camera_ready_lowres/figtex_region_replace.tex
\centering
\begin{overpic}[trim=2cm 19cm 23cm -3cm,clip,width=1\linewidth,grid=false]{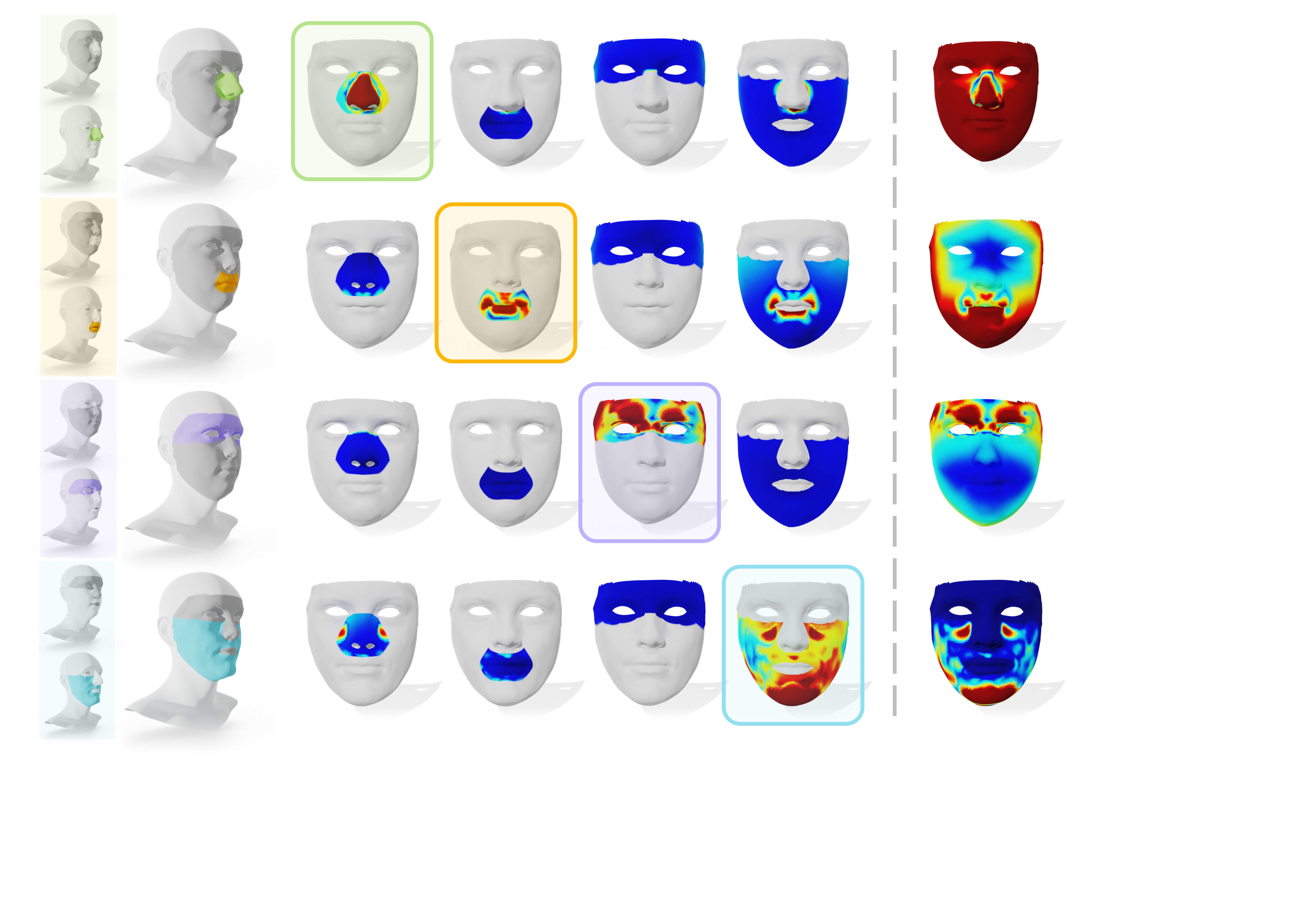}
\put(0,67){\tiny\itshape\bfseries $S_L$}
\put(9,67){\tiny\itshape\bfseries $S_P$}
\put(0,58){\tiny\itshape\bfseries $S_1$}
\put(0,49){\tiny\itshape\bfseries $S_L$}
\put(0,41){\tiny\itshape\bfseries $S_2$}
\put(0,31.5){\tiny\itshape\bfseries $S_L$}
\put(0,23){\tiny\itshape\bfseries $S_3$}
\put(0,13){\tiny\itshape\bfseries $S_L$}
\put(0,5){\tiny\itshape\bfseries $S_4$}

\put(3,72.5){\scriptsize Modified}
\put(5,69.5){\scriptsize Shapes}
\put(22,72.5){\scriptsize\bfseries Our Region-based Alignment}
\put(86,72.5){\scriptsize Global}
\put(82,69.5){\scriptsize Alignment}

\put(27.5,69.3){\tiny\bfseries\itshape @nose}
\put(40,69.3){\tiny\bfseries\itshape @mouth}
\put(52,69.3){\tiny\bfseries\itshape @forehead}
\put(67.2,69.3){\tiny\bfseries\itshape @cheek}
\end{overpic}

%% file: latex/img_camera_ready_lowres/figtex_res_mtd.tex
\centering
\begin{overpic}[trim=0cm 99cm 0cm 3cm,clip,width=1\linewidth,grid=false]{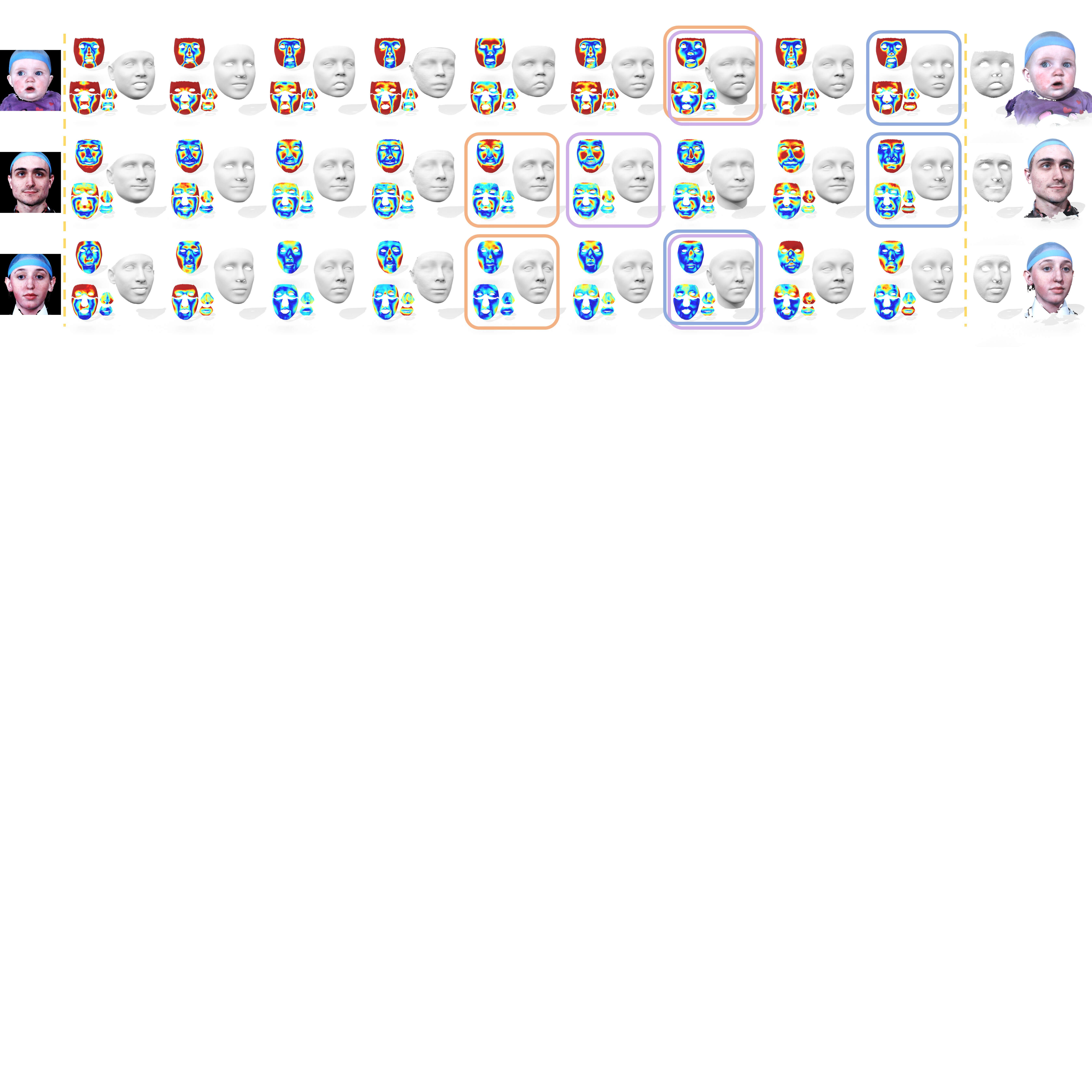}
\put(0,28.5){\scriptsize\bfseries Input}
\put(8,28.5){\scriptsize\bfseries \cite{ExpNet}}
\put(17.5,28.5){\scriptsize\bfseries \cite{RingNet}}
\put(27,28.5){\scriptsize\bfseries \cite{MGCNet}}
\put(35.5,28.5){\scriptsize\bfseries \cite{PRNet}}
\put(45,28.5){\scriptsize\bfseries \cite{deep3d}}
\put(54,28.5){\scriptsize\bfseries \cite{3DDFA_v2}}
\put(63.5,28.5){\scriptsize\bfseries \cite{ganfit}}
\put(72.5,28.5){\scriptsize\bfseries \cite{Nonlinear3DMM}}
\put(82,28.5){\scriptsize\bfseries \cite{DECA}}
\put(91.5,28.5){\scriptsize\bfseries G.T.}

\put(31,26.5){\scriptsize\bfseries $\star$}
\put(49,16.8){\scriptsize\bfseries $\star$}
\put(49,7.5){\scriptsize\bfseries $\star$}

\put(68,26.5){\scriptsize\bfseries $\dagger$}
\put(31,16.8){\scriptsize\bfseries $\dagger$}
\put(31,7.5){\scriptsize\bfseries $\dagger$}

\end{overpic}

%% file: latex/tables/table_main_result.tex
\scriptsize
\resizebox{1\linewidth}{!}{%
\begin{tabular}{r|ccccccccc|cccc}
\toprule[1pt]
\multirow{3}{*}{\begin{tabular}[c]{@{}l@{}}methods /\\ $e$ (mm)\end{tabular}} & \multicolumn{9}{c|}{{\bicp} (ours)} & \multicolumn{4}{c}{{\gicp}}     \\
        &\multicolumn{2}{c}{\cellcolor{col_nose!30}@$\mathcal{R}_N$(nose)}  &\multicolumn{2}{c}{\cellcolor{col_mouth!30}@$\mathcal{R}_M$(mouth)}   & \multicolumn{2}{c}{\cellcolor{col_forehead!30}@$\mathcal{R}_{F}$(forehead)}   &                \multicolumn{2}{c}{\cellcolor{col_cheek!30} @$\mathcal{R}_{C}$(cheek)}                                &  \cellcolor{gray!30} all    & \multicolumn{2}{c}{$\Scale[0.9]{e\big(\mymap{p}{h}{pts}\big)}$}  & \multicolumn{2}{c}{$\Scale[0.9]{e\big(\mymap{h}{p}{pts}\big)}$}       \\ 
                                                                          & avg.                   & std.                  & avg.                   & std.                  & avg.                    & std.                  & avg.                   & \multicolumn{1}{c|}{std.}                  & avg.    & avg.               & std.   & avg.               & std.  \\  \midrule[1pt]
ExpNet~\cite{ExpNet}                                                                    & 2.509  & 0.486 & 1.912                & 0.450 & 3.084             & 1.005                & 1.717  & \multicolumn{1}{c|}{0.590} & 2.306          & 2.650                & 0.549 & 2.297 & 0.616\\
RingNet~\cite{RingNet}                                                                   & 1.934                 & 0.458                & 2.074                 & 0.616                & 2.995              & 0.908                & 2.028               & \multicolumn{1}{c|}{0.720}                & 2.258          & 2.016                 & 0.489 &2.762 & 1.016  \\
MGCNet~\cite{MGCNet}                                                                   & 1.771               & 0.380                & \cellcolor{best_two!20}{\textbf{1.417}}                & 0.409                & \cellcolor{best_two!20}{\textbf{2.268}}                & 0.503               & 1.639               & \multicolumn{1}{c|}{0.650}                & \cellcolor{best_two!20}{\textbf{1.774}}         & 2.388                & 0.865  & 2.094 & 0.750  \\
PRNet~\cite{PRNet}                                                                    & 1.923               & 0.518                & 1.838              & 0.637                & 2.429            & 0.588                & 1.863               & \multicolumn{1}{c|}{0.698}                & 2.013          & 3.036                & 0.933 &2.302 & 0.747 \\
Deep3D~\cite{deep3d}                                                                     & \cellcolor{best_two!20}{\textbf{1.719}}             & 0.354                & \cellcolor{best!20}{\textbf{1.368}}  & 0.439 & \cellcolor{best!20}{\textbf{2.015}} & 0.449 & 1.528 & \multicolumn{1}{c|}{0.501}                & \cellcolor{best!20}{\textbf{1.657}} & 2.142              & 0.651  & \cellcolor{best!20}{\textbf{1.908}} & 0.553\\
3DDFA-v2~\cite{3DDFA_v2}                                                                  & 1.903            & 0.517                & 1.597              & 0.478                & 2.477          & 0.647                & 1.757              & \multicolumn{1}{c|}{0.642}                & 1.926          & 2.788               & 0.951  & 2.279& 0.765\\
GANFit~\cite{ganfit}                                                                  & 1.928           & 0.490                & 1.812            & 0.544                & 2.402               & 0.545                & \cellcolor{best!20}{\textbf{1.329}}               & \multicolumn{1}{c|}{0.504}                & 1.868          & \cellcolor{best_two!20}{\textbf{1.899}}             & 0.730  & \cellcolor{best_two!20}{\textbf{1.999}} & 0.748\\
N-3DMM~\cite{Nonlinear3DMM}                                                                   & 2.936              & 0.810                & 2.375             & 0.599                & 4.582           & 1.448                & 1.918               & \multicolumn{1}{c|}{0.801}                & 2.953          & 3.681           & 1.566 & 3.252& 1.198 \\
DECA-c~\cite{DECA}                                                             & \cellcolor{best!20}{\textbf{1.697}}  & {0.355} & 2.516             & 0.839               & 2.349  & 0.576 & \cellcolor{best_two!20}{\textbf{1.479}}           & \multicolumn{1}{c|}{0.535}                & 2.010          & \cellcolor{best!20}{\textbf{1.698}} & 0.397  & 2.183 & 0.798 \\ \bottomrule[1pt]
\end{tabular}%
}

%% file: latex/img_camera_ready_lowres/figtex_eg_merge.tex
\centering

\begin{overpic}[trim=7cm 41cm 86cm 5cm,clip,width=1\textwidth,grid=false]{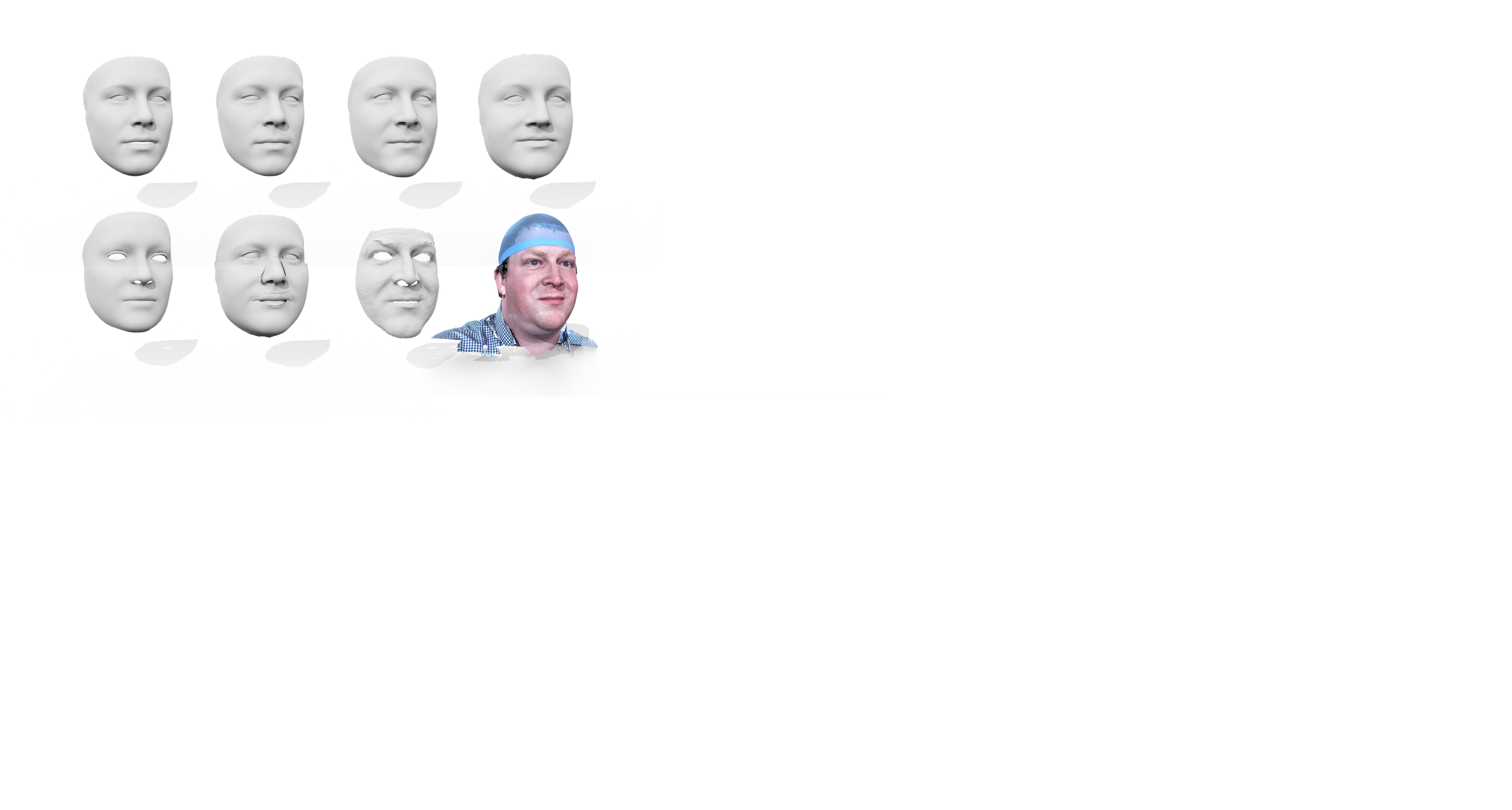}

\put(0,64){\tiny\bfseries $\mathcal{R}_N$\cite{3DDFA}}
\put(25,64){\tiny\bfseries $\mathcal{R}_M$\cite{MGCNet}}
\put(50,64){\tiny\bfseries $\mathcal{R}_F$\cite{deep3d}}
\put(75,64){\tiny\bfseries $\mathcal{R}_C$\cite{Nonlinear3DMM}}
\put(0,33){\tiny\bfseries \textit{g}ICP\cite{DECA}}
\put(28,33){\tiny\bfseries\itshape \underline{Merged}}
\put(54,33){\tiny\bfseries Ground-truth}

\end{overpic}

%% file: latex/img_camera_ready_lowres/figtex_eg_convergence.tex
\centering
\begin{overpic}[trim=0cm 42cm 39cm 0cm,clip,width=0.98\textwidth,grid=false]{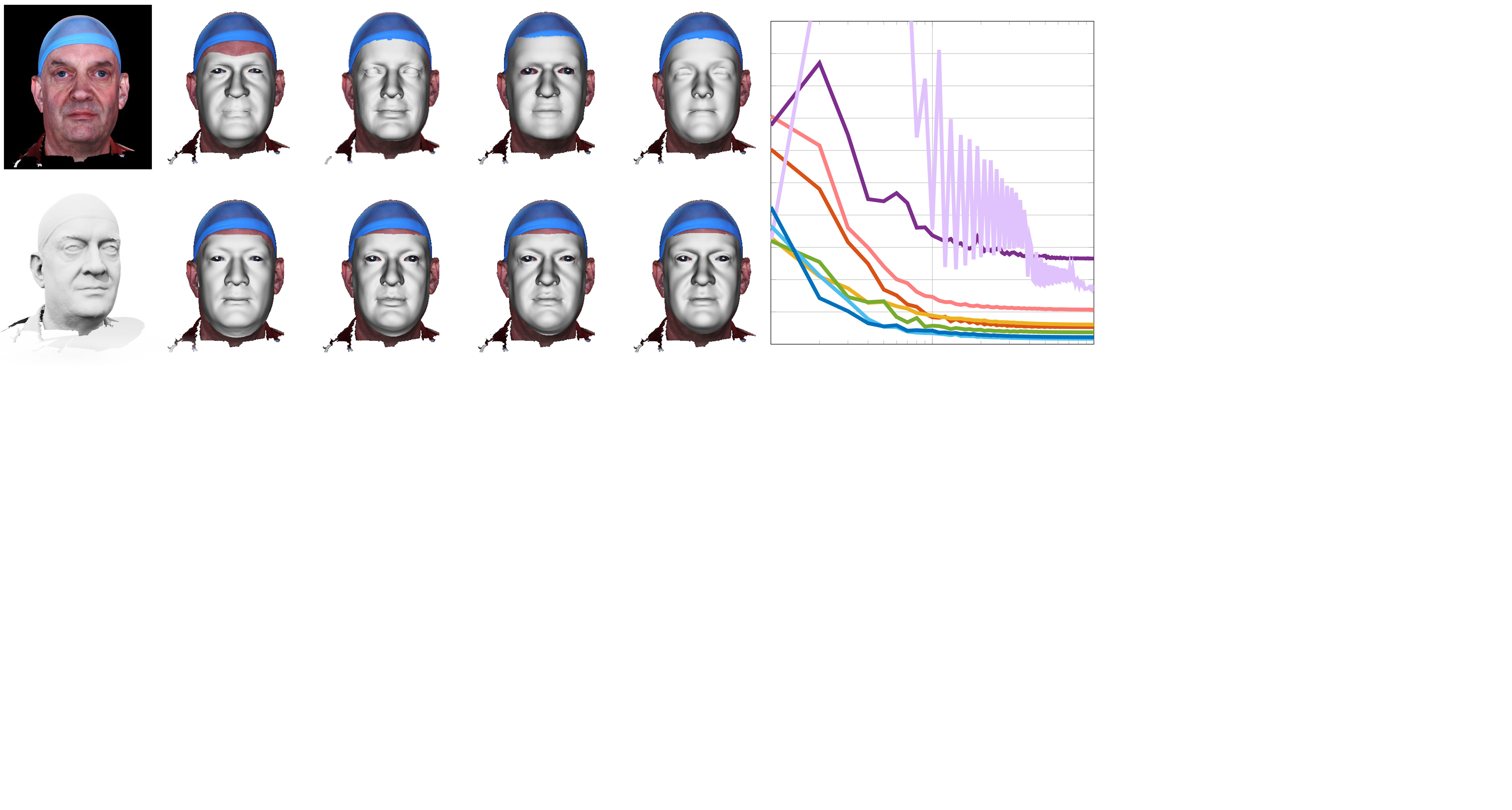}
\put(17,34){\tiny \textcolor{red!50}{LYHM}}
\put(32,34){\tiny \textcolor{mycolor1}{BFM}}
\put(44,34){\tiny \textcolor{mycolor2}{FLAME}}
\put(59,34){\tiny \textcolor{mycolor3}{LSFM}}
\put(19,16.75){\tiny \textcolor{mycolor4}{FS}}
\put(30,16.75){\tiny \textcolor{mycolor5}{\nextone}}
\put(44,16.75){\tiny \textcolor{mycolor6}{\nexttwo}}
\put(60,16.75){\tiny \textcolor{mycolor7}{\underline{Ours}}}
\put(2.5,34){\scriptsize\bfseries Input}
\put(4,16.75){\tiny\bfseries G.T.}
\put(71,34){\scriptsize\bfseries Loss w.r.t. iter}
\end{overpic}

%% file: latex/tables/table_rgb_fitting_result.tex
\resizebox{1\linewidth}{!}{%
\begin{tabular}{r | cccccccc| c | cccccccc| c}
\toprule[1pt]
\multirow{3}{*}{\begin{tabular}[c]{@{}l@{}}3DMMs /\\ $e$ (mm)\end{tabular}} 
 &\multicolumn{9}{c|}{\textbf{\cellcolor{pink!20}  \textit{ RGB Fitting}}}                                                                                 & \multicolumn{9}{c}{\cellcolor{gray!10} \textbf{\textit{RGB-D Fitting}}}                                                                                 \\
&\multicolumn{2}{c}{\cellcolor{col_nose!30}@$\mathcal{R}_N$}  &\multicolumn{2}{c}{\cellcolor{col_mouth!30}@$\mathcal{R}_M$}   & \multicolumn{2}{c}{\cellcolor{col_forehead!30}@$\mathcal{R}_{F}$}   &                \multicolumn{2}{c}{\cellcolor{col_cheek!30} @$\mathcal{R}_{C}$}          &  \cellcolor{gray!30} all        &\multicolumn{2}{c}{\cellcolor{col_nose!30}@$\mathcal{R}_N$}  &\multicolumn{2}{c}{\cellcolor{col_mouth!30}@$\mathcal{R}_M$}   & \multicolumn{2}{c}{\cellcolor{col_forehead!30}@$\mathcal{R}_{F}$}   &                \multicolumn{2}{c}{\cellcolor{col_cheek!30} @$\mathcal{R}_{C}$}          &  \cellcolor{gray!30} all                         \\
 & avg.                   & std.                  & avg.                   & std.                  & avg.                    & std.                  & avg.                   & \multicolumn{1}{c|}{std.}                  & avg.    & avg.                   & std.                  & avg.                   & std.                  & avg.                    & std.                  & avg.                   & \multicolumn{1}{c|}{std.}                  & avg. \\
\midrule[1pt]
BFM~\cite{bfm09}   & 2.925                & 0.704               & 2.175               & 0.550                & 3.359                & 0.660                & 1.742            & 0.410
                 & 2.550                & 1.700  & 0.277 & 1.170  &  0.355 & 2.308  & 0.501 & 0.587 &  0.100 & 1.441 \\
FLAME~\cite{flametopo}  & \cellcolor{best_two!10}{\textbf{2.700}}                & 0.543               & 2.616               & 0.476               & 3.891                & 0.786              & 2.737               & 0.687              & 2.986               & 1.687     & 0.232 & 1.397  & 0.354  &  2.178 & 0.609 & 0.495 & 0.125 & 1.439             \\
LSFM~\cite{3dmm10000}   & \cellcolor{best!10}{\textbf{2.455}}                & 0.666               & 2.446               & 0.768             & 4.062               & 0.807               & 3.756               &1.292                 & 3.180               & 1.727  & 0.320 & 1.906  & 0.638  & 2.370  & 0.612 & 0.869 & 0.218 & 1.718                 \\
FS~\cite{facescape}   & 2.852                & 0.776                & 2.524                & 0.827                & \cellcolor{best_two!10}{\textbf{2.430}}             & 0.613               & 1.739                & 0.450                & 2.386               & 2.181    & 0.494 & 2.468  & 0.866  & 2.057  & 0.597 & 1.003 & 0.208 & 1.927              \\
{\nextone}~\cite{hifi3dface2021tencentailab}   & 2.974                & 0.752                & \cellcolor{best_two!10}{\textbf{1.285}}              & 0.364               & 2.519                & 0.490                & 2.070                & 0.533                & 2.212               & \cellcolor{best_two!10}{\textbf{1.653}}         & 0.258 &  0.909 &  0.332 & 1.343  & 0.366 & 0.468 &  0.121 & 1.093         \\
{\nexttwo}~\cite{hifi3dface2021tencentailab}  & 3.076                & 0.709 & \cellcolor{best!10}{\textbf{1.201}} & 0.399                & 2.527 & 0.561                & 1.866 & 0.566                & \cellcolor{best_two!10}{\textbf{2.167}}               & 1.746                & 0.271 & \cellcolor{best!10}{\textbf{0.607}}                & 0.338                & \cellcolor{best_two!10}{\textbf{1.235}}   & 0.363 & \cellcolor{best_two!10}{\textbf{0.302}}  & 0.084  &  \cellcolor{best_two!10}{\textbf{0.972}}                 \\\midrule[0.8pt]
LYHM*~\cite{LYHM}   & 2.723                & 0.578 & 1.988                & 0.556                & 3.752                & 0.716              & \cellcolor{best!10}{\textbf{1.475}}             & 0.439                & 2.485              & 2.144       & 0.331 & 1.654  &  0.520 &  3.174 & 0.676 & 0.673 & 0.155 & 1.911          \\
\multicolumn{1}{c|}{{\textbf{Ours}*}}   & 2.898 & 0.732               & 1.288 & 0.408 & \cellcolor{best!10}{\textbf{2.216}} & 0.612 & \cellcolor{best_two!10}{\textbf{1.599}} & 0.537 & \cellcolor{best!10}{\textbf{2.000}} & \cellcolor{best!10}{\textbf{1.542}} & 0.258 &  \cellcolor{best_two!10}{\textbf{0.621}} &  0.341 & \cellcolor{best!10}{\textbf{1.085}}  & 0.359 & \cellcolor{best!10}{\textbf{0.265}} & 0.080 & \cellcolor{best!10}{\textbf{0.878}}   \\ \bottomrule[1pt]
\end{tabular}%
}
\begin{tablenotes}
\scriptsize
\item *Some of the test shapes in {\name} are used to construct LYHM and our 3DMM.
\end{tablenotes}

%% file: latex/img_camera_ready_lowres/figtex_res_basis.tex
\centering
\begin{overpic}[trim=0cm 104cm 0cm 0cm,clip,width=1\linewidth,grid=false]{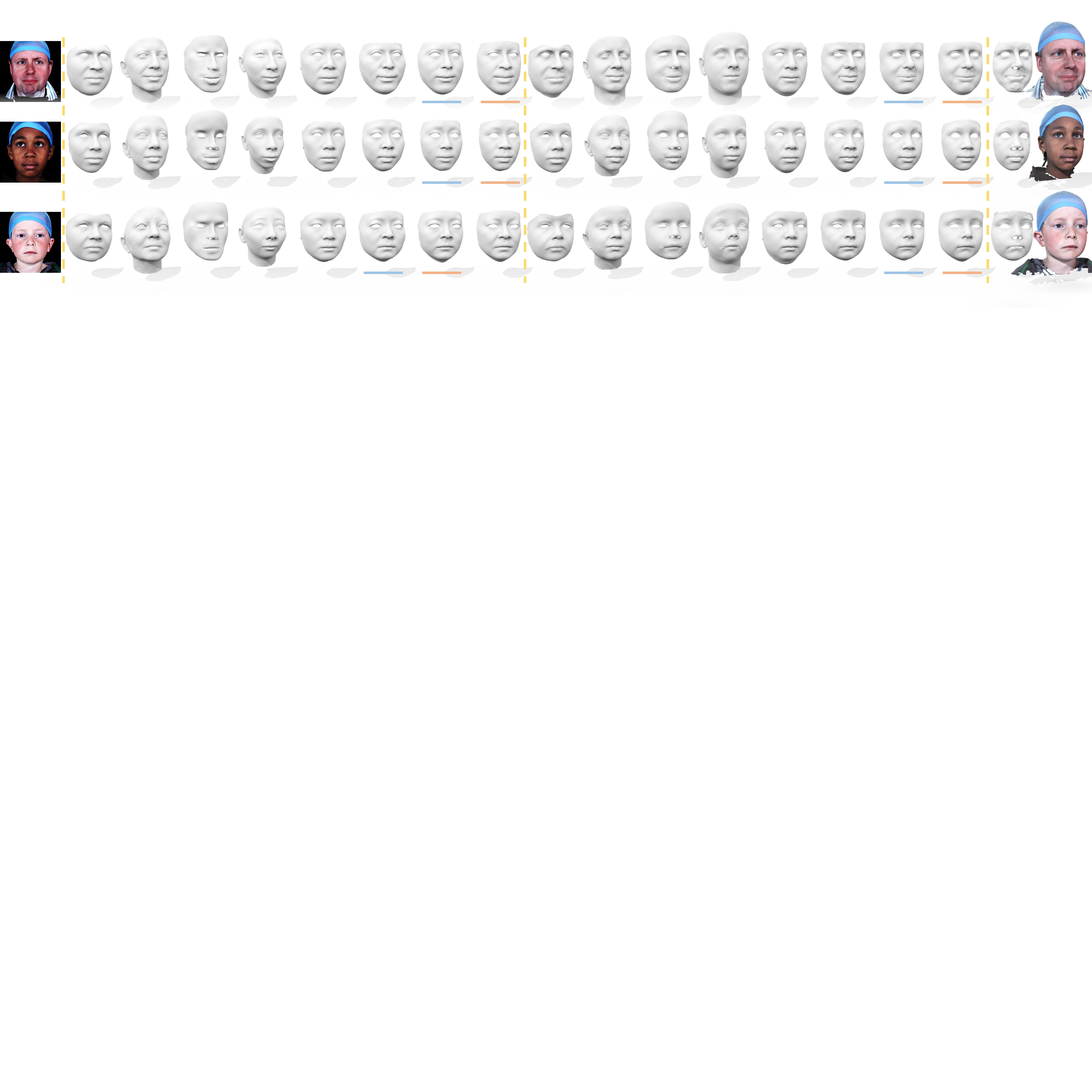}
\put(0,25.6){\scriptsize\bfseries Input}
\put(15,26){\scriptsize\bfseries Without Depth (RGB)}
\put(58,26){\scriptsize\bfseries With Depth (RGB-D)}

\put(6.5,24){\tiny\bfseries \cite{LYHM}}
\put(12,24){\tiny\bfseries \cite{bfm09}}
\put(17.5,24){\tiny\bfseries \cite{flametopo}}
\put(23,24){\tiny\bfseries \cite{3dmm10000}}
\put(28,24){\tiny\bfseries \cite{facescape}}
\put(34,24){\tiny\bfseries \cite{hifi3dface2021tencentailab}}
\put(39.5,24){\tiny\bfseries \cite{hifi3dface2021tencentailab}$^A$}
\put(44.2,24){\tiny\bfseries\itshape \underline{Ours}}

\put(49,24){\tiny\bfseries \cite{LYHM}}
\put(54.5,24){\tiny\bfseries \cite{bfm09}}
\put(60,24){\tiny\bfseries \cite{flametopo}}
\put(65.5,24){\tiny\bfseries \cite{3dmm10000}}
\put(70.5,24){\tiny\bfseries \cite{facescape}}
\put(76.5,24){\tiny\bfseries \cite{hifi3dface2021tencentailab}}
\put(81.5,24){\tiny\bfseries \cite{hifi3dface2021tencentailab}$^A$}
\put(86.5,24){\tiny\bfseries\itshape \underline{Ours}}

\put(92,25.6){\scriptsize\bfseries G.T.}
\end{overpic}

%% file: latex/img_camera_ready_lowres/figtex_three_next.tex
\centering
\begin{overpic}[trim=0cm 22.5cm 0cm 0cm,clip,width=1\linewidth,grid=false]{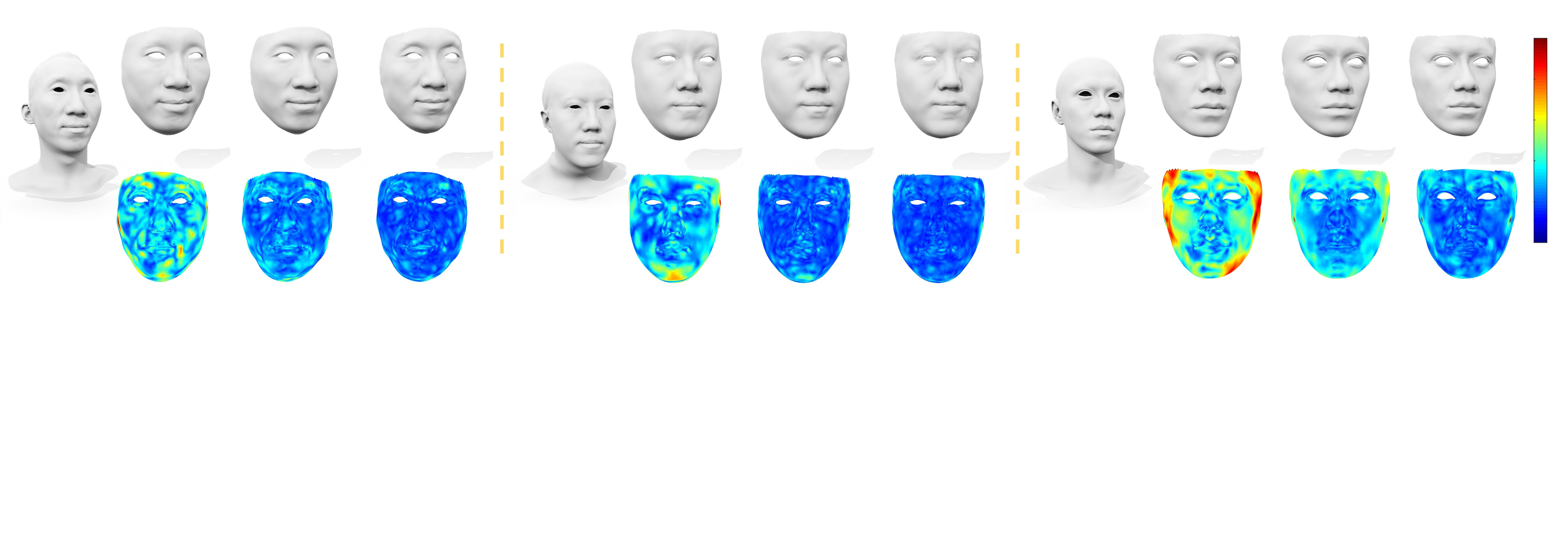}
\put(1,19){\tiny\bfseries G.T. }
\put(3,17){\tiny\bfseries $S_L$}

\put(33.5,19){\tiny\bfseries G.T. }
\put(35.5,17){\tiny\bfseries $S_L$}

\put(66,19){\tiny\bfseries G.T. }
\put(68,17){\tiny\bfseries $S_L$}

\put(7,19){\tiny\bfseries {\nextone}}
\put(15,19){\tiny\bfseries {\nexttwo}}
\put(25,19){\tiny\bfseries \underline{Ours}}

\put(40,19){\tiny\bfseries {\nextone}}
\put(48,19){\tiny\bfseries {\nexttwo}}
\put(58,19){\tiny\bfseries \underline{Ours}}

\put(73,19){\tiny\bfseries {\nextone}}
\put(81,19){\tiny\bfseries {\nexttwo}}
\put(90.5,19){\tiny\bfseries \underline{Ours}}

\put(5.5,0.5){\tiny $0.86$}\put(9,0.5){\tiny $\pm$}\put(10.5,0.5){\tiny $0.42$}
\put(14.5,0.5){\tiny $0.55$}\put(18,0.5){\tiny $\pm$}\put(19.5,0.5){\tiny $0.25$}
\put(23.5,0.5){\tiny $0.41$}\put(27,0.5){\tiny $\pm$}\put(28.5,0.5){\tiny $0.21$}

\put(39,0.5){\tiny $0.78$}\put(42.5,0.5){\tiny $\pm$}\put(44,0.5){\tiny $0.40$}
\put(48,0.5){\tiny $0.43$}\put(51.5,0.5){\tiny $\pm$}\put(53,0.5){\tiny $0.21$}
\put(57,0.5){\tiny $0.37$}\put(60.5,0.5){\tiny $\pm$}\put(62,0.5){\tiny $0.19$}

\put(72.5,0.5){\tiny $1.11$}\put(76,0.5){\tiny $\pm$}\put(77.5,0.5){\tiny $0.63$}
\put(81.5,0.5){\tiny $0.72$}\put(85,0.5){\tiny $\pm$}\put(86.5,0.5){\tiny $0.38$}
\put(90.5,0.5){\tiny $0.50$}\put(94,0.5){\tiny $\pm$}\put(95.5,0.5){\tiny $0.25$}

\put(97.8,3){\tiny $0$}
\put(95,17.6){\tiny $10$mm}
\end{overpic}

%% file: latex/img_camera_ready_supp_lowres/figtex_3dmm_topo.tex
    \centering
    \begin{overpic}[trim=0cm 0cm 0cm 0cm,clip,width=1\linewidth,grid=false]{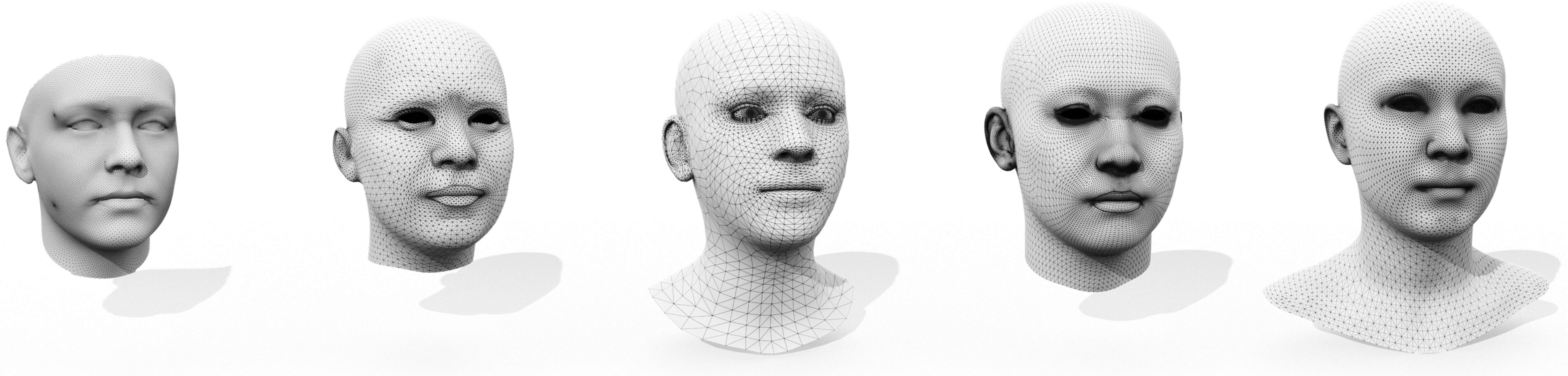}
    \put(4,26){\scriptsize\bfseries BFM}
    \put(2.5,23){\scriptsize\bfseries \& LSFM}
    \put(18,26){\scriptsize\bfseries FaceWareHouse}
    \put(21,23){\scriptsize\bfseries \& LYHM}
    
    \put(43,25){\scriptsize\bfseries FLAME}
    \put(63,25){\scriptsize\bfseries FaceScape}
    \put(85,25){\scriptsize\bfseries {\nextone}}
    
    \put(2.5,3){\scriptsize\bfseries $n_v = 53215$}
    \put(2,0){\scriptsize\bfseries $n_f = 105840$}
    
    \put(22.5,3){\scriptsize\bfseries $n_v = 11510$}
    \put(22.5,0){\scriptsize\bfseries $n_f = 22800$}
    
    \put(46,3){\scriptsize\bfseries $n_v = 5023$}
    \put(46,0){\scriptsize\bfseries $n_f = 9976$}
    
    \put(66.5,3){\scriptsize\bfseries $n_v = 26317$}
    \put(66.5,0){\scriptsize\bfseries $n_f = 52261$}
    
    \put(87,3){\scriptsize\bfseries $n_v = 20481$}
    \put(87,0){\scriptsize\bfseries $n_f = 40832$}
    \end{overpic}

 

%% file: latex/tables/table_apdx_params_rgb.tex
\centering
\begin{tabular}{l|cccccc}
\toprule[1pt]
3DMMs  & $\omega_{\text{rgb}}$    & $\omega_{\text{dep}}$  & $\omega_{\text{id}}$   & $\omega_{\text{lmk}}$  & $\omega_{\text{shp}}$     & $\omega_{\text{tex}}$ \\ \midrule[1pt]
BFM             & 1000.0 & 0.0 & 4.00 & 5.0  & 5.0     & 1.0 \\
FLAME           & 1000.0 & 0.0 & 4.00 & 5.0  & 2.0     & 10.0 \\
LSFM            & 1000.0 & 0.0 & 4.00 & 5.0  & 2.0     & 1.0 \\
FS              & 1000.0 & 0.0 & 0.10 & 5.0  & 50000.0* & 0.0 \\
{\nextone}      & 1000.0 & 0.0 & 4.00 & 5.0  & 2.0     & 1.0 \\
{\nexttwo}      & 1000.0 & 0.0 & 4.00 & 5.0  & 2.0     & 1.0 \\
LYHM            & 1000.0 & 0.0 & 4.00 & 10.0 & 100.0    & 0.01 \\
\textbf{Ours}   & 1000.0 & 0.0 & 4.00 & 5.0  & 2.0     & 1.0 \\ \bottomrule[1pt]
\end{tabular}%


%% file: latex/tables/table_apdx_params.tex
\centering
\begin{tabular}{l|cccccc}
\toprule[1pt]
3DMMs  & $\omega_{\text{rgb}}$    & $\omega_{\text{dep}}$  & $\omega_{\text{id}}$   & $\omega_{\text{lmk}}$  & $\omega_{\text{shp}}$     & $\omega_{\text{tex}}$ \\ \midrule[1pt]
BFM    & 1000.0 & 1000.0 & 1.00 & 5.0  & 5.0     & 1.0 \\
FLAME  & 1000.0 & 1000.0 & 1.00 & 5.0  & 2.0     & 2.0 \\
LSFM   & 1000.0 & 1000.0 & 1.00 & 5.0  & 2.0     & 1.0 \\
FS     & 1000.0 & 1000.0 & 0.10 & 5.0  & 50000.0* & 0.0 \\
{\nextone}   & 1000.0 & 1000.0 & 1.00 & 5.0  & 2.0     & 1.0 \\
{\nexttwo} & 1000.0 & 1000.0 & 1.00 & 5.0  & 2.0     & 1.0 \\
LYHM   & 1000.0 & 1000.0 & 1.00 & 10.0 & 20.0    & 0.2 \\
\textbf{Ours}   & 1000.0 & 1000.0 & 1.00 & 5.0  & 2.0     & 1.0 \\ \bottomrule[1pt]
\end{tabular}%